\documentclass[lettersize,journal]{IEEEtran}
\pdfoutput=1
\usepackage{amsmath,amsfonts}
\usepackage{algorithmic}
\usepackage{algorithm}
\usepackage{array}
\usepackage[caption=false,font=normalsize,labelfont=sf,textfont=sf]{subfig}
\usepackage{textcomp}
\usepackage{stfloats}
\usepackage{url}
\usepackage{verbatim}
\usepackage{graphicx}
\usepackage{cite}
\usepackage{booktabs}
\usepackage{threeparttable}
\usepackage{multirow}
\usepackage{pifont}
\usepackage{bbding}
\begin{document}

\title{Efficient Portrait Matte Creation With Layer Diffusion and Connectivity Priors}

\author{Zhiyuan Lu, Hao Lu,~\IEEEmembership{Member,~IEEE} and Hua Huang,~\IEEEmembership{Senior Member,~IEEE}
\thanks{Corresponding author: Hua Huang.}% <-this % stops a space
\thanks{Zhiyuan Lu and Hua Huang are with the School of Artificial Intelligence, Beijing Normal University, Beijing 100875, China (e-mail: zylu@mail.bnu.edu.cn; huahuang@bnu.edu.cn).}
\thanks{ Hao Lu is with the School of Artificial Intelligence and Automation, Huazhong University of Science and Technology, Wuhan 430074, China
 (e-mail: hlu@hust.edu.cn).}}

\markboth{Journal of \LaTeX\ Class Files,~Vol.~14, No.~8, August~2021}%
{Shell \MakeLowercase{\textit{et al.}}: A Sample Article Using IEEEtran.cls for IEEE Journals}


\maketitle

\begin{abstract}
Learning effective deep portrait matting models requires training data of both high quality and large quantity. Neither quality nor quantity can be easily met for portrait matting, however. Since the most accurate ground-truth portrait mattes are acquired in front of the green screen, it is almost impossible to harvest a large-scale portrait matting dataset in reality. This work shows that one can leverage text prompts and the recent Layer Diffusion model to generate high-quality portrait foregrounds and extract latent portrait mattes. However, the portrait mattes cannot be readily in use due to significant generation artifacts. Inspired by the connectivity priors observed in portrait images, that is, the border of portrait foregrounds always appears connected, a connectivity-aware approach is introduced to refine portrait mattes. Building on this, a large-scale portrait matting dataset is created, termed LD-Portrait-20K, with $20,051$ portrait foregrounds and high-quality alpha mattes. Extensive experiments demonstrated the value of the LD-Portrait-20K dataset, with models trained on it significantly outperforming those trained on other datasets. In addition, comparisons with the chroma keying algorithm and an ablation study on dataset capacity further confirmed the effectiveness of the proposed matte creation approach. Further, the dataset also contributes to state-of-the-art video portrait matting, implemented by simple video segmentation and a trimap-based image matting model trained on this dataset.
\end{abstract}


\section{Introduction}
\IEEEPARstart{P}{ortrait} matting is a critical task in computer vision, with widespread applications in video editing, virtual try-on, and online meetings~\cite{boda2018survey,li2013image,beyer1965traveling,fang2022user,xu2022self}. The goal is to precisely separate the foreground portrait from an image and predict the transparency of each pixel (alpha matte) for seamless foreground-background % integration
separation. However, portrait matting is inherently an underdetermined problem, originated from the ill-posed nature of matting equation
\begin{equation}
\label{eq1}
    I=\alpha F+(1-\alpha)B\,,
\end{equation}
where $I$ is an RGB image, $\alpha$ is the blending coefficient, a.k.a. the alpha matte, and $F$ and $B$ denote the foreground and background, respectively.
It can be observed that, each pixel only has three known values but requires estimating seven unknowns~\cite{wang2008image,aksoy2017designing,levin2007closed,karacan2017alpha,johnson2016sparse}.

Classic matting methods typically rely on assumptions about color distributions or pixel differences, making them effective in simple scenarios. However, they struggle in complex environments, especially when addressing fine edge details and semi-transparent regions, often resulting in blurred edges and loss of detail. These methods also require user interaction, which increases complexity and cost~\cite{li2019survey,wang2007optimized,sun2004poisson,li2017patch,zheng2008fuzzymatte,lee2017parallel,varnousfaderani2013weighted}. Deep learning has significantly advanced portrait matting by leveraging powerful feature extraction and modeling capabilities, achieving notable performance improvements in handling intricate edges and backgrounds~\cite{cho2018deep,li2023deep, ye2024unifying,lin2023omnimatterf, ding2022deep}. Nevertheless, their reliance on large-scale, high-quality datasets remains a major limitation, as creating such datasets is costly and lacks sufficient diversity and scale. This highlights the urgent need for efficient and scalable methods to construct diverse, high-quality portrait matting datasets.

\begin{figure*}
    \centering
    \includegraphics[scale=.14]{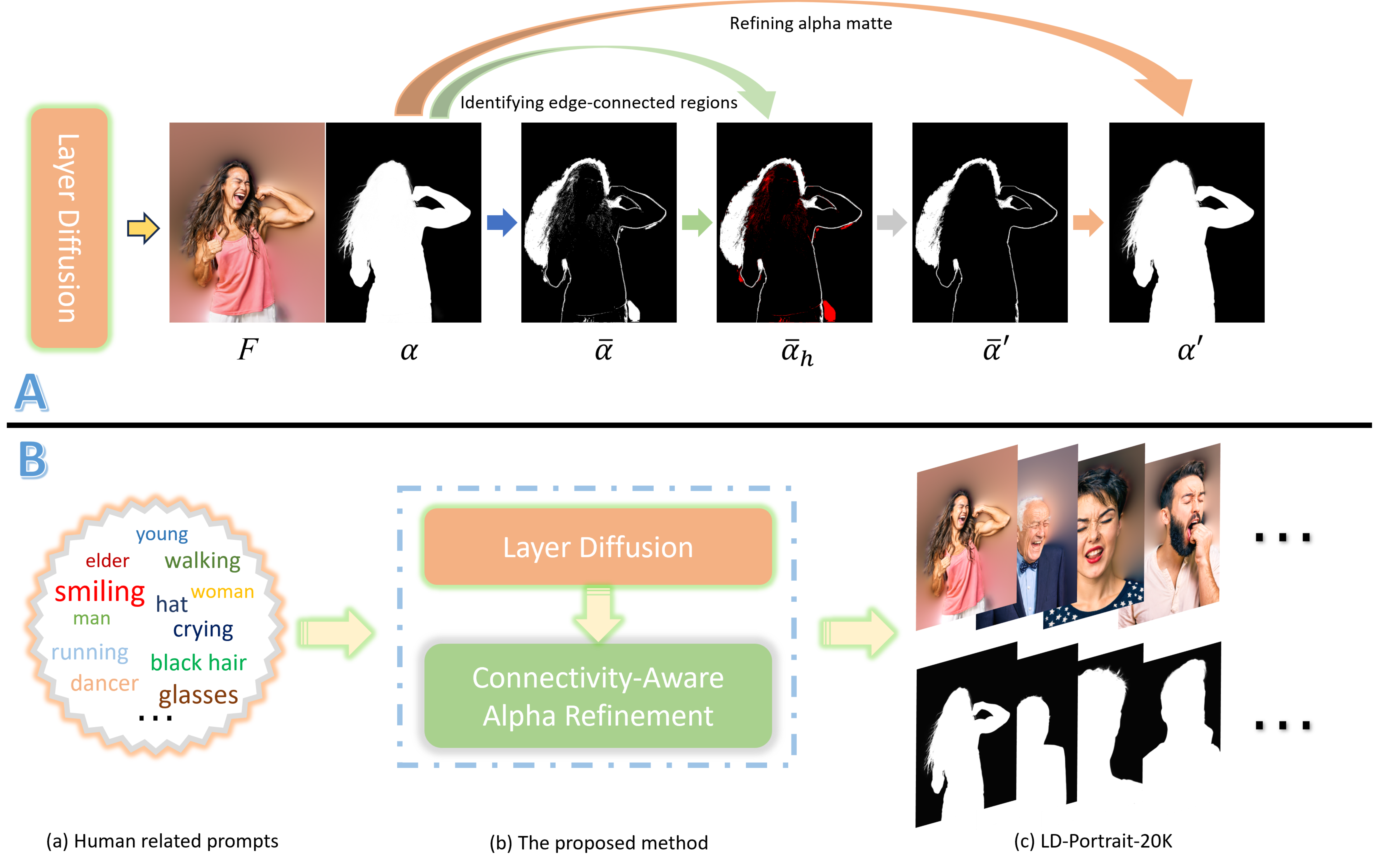}
    \caption{\textbf{Overview of efficient portrait matte creation}. A represents the process of Connectivity-Aware Alpha Refinement, where the background regions of $F$ (where the alpha value is strictly zero) are padded by Layer Diffusion using an iterative Gaussian filter to avoid aliasing and unnecessary edge patterns. $\bar{\alpha}$ represents regions in the alpha matte where the pixel values are neither 0 nor 255, while the red areas in $\bar{\alpha}_h$ indicate pixels with erroneous alpha values. B illustrates the dataset creation process, where a diverse dataset is generated through a wide range of carefully designed prompts. }
    \label{figure-intro-1}
\end{figure*}

Matting datasets are primarily created through automated or manual annotation. Automated methods, such as chroma keying~\cite{shen2016deep,zhang2021attention,lin2021real}, extract foreground alpha mattes from solid-color backgrounds. Albeit efficient, these methods rely on uniform backgrounds, limiting data diversity and introducing artifacts like color bleeding, reducing model generalization. In contrast, manual annotation, using tools like Photoshop, delivers high-quality results~\cite{xu2017deep,qiao2020attention,li2021deep,li2022bridging}. However, it is labor-intensive, costly, and difficult to scale, especially for high-resolution images or videos with intricate edges and semi-transparent regions. Both approaches face challenges in addressing complex lighting and interactions between foreground and background, underscoring the need for innovative dataset construction methods.

The emergence of diffusion models has brought revolutionary breakthroughs in the creation of matting datasets~\cite{burgert2024magick}. As a powerful generative technology, diffusion models synthesize highly realistic image content through an iterative denoising process starting from random noise. Notably, the Layer Diffusion model~\cite{zhang2024transparent} simultaneously generates transparent images with foregrounds and corresponding alpha mattes, % offering crucial support 
pointing out a valuable direction for the rapid construction of large-scale, high-quality matting datasets. However, the alpha mattes generated by Layer Diffusion still exhibit some inconsistencies, rendering them unsuitable for direct use in training matting models. As shown in Figure~\ref{figure-intro-1}, while Layer Diffusion is capable of producing fine alpha values along human edges, it often generates significant errors in regions of absolute foreground or background. If correcting these errors while preserving fine alpha values, the rapid generation of large volumes of high-quality matting data could be achieved. % 

Through initial studies, a prior of human alpha mattes was observed: \textit{semi-transparent regions are predominantly concentrated along the edges of the body, these edges comprise multiple connected regions, with each connected region corresponding to an independent semi-transparent area.} This finding reveals the inherent connectivity of semi-transparent regions in alpha mattes. Based on this observation, an improved dataset construction method is proposed. Specifically, for the alpha mattes generated by Layer Diffusion, only the connected semi-transparent regions along the edges are retained, and erroneous values in other regions are corrected based on their foreground or background attributes. This method not only preserves edge details but also effectively reduces interference from erroneous alpha values, thereby significantly enhancing the quality of the generated data and overcoming the limitations of traditional approaches.

Using the proposed method, the LD-Portrait-20K dataset was efficiently constructed, comprising 20,051 high-quality human foreground images and their corresponding precise alpha mattes. To ensure the diversity and representativeness of the dataset, various attributes such as gender, age, pose, expression, and clothing styles were incorporated. Extensive comparative experiments validated the exceptional quality of the LD-Portrait-20K dataset. The experiments demonstrated that models trained on this dataset significantly outperformed those trained on other datasets. Furthermore, comparisons with the chroma keying algorithm highlighted the robustness of the method, while ablation studies on dataset capacity confirmed its scalability and effectiveness. Experimental results also revealed that LD-Portrait-20K holds significant value for video portrait matting tasks. By integrating segmentation models with trimap-based matting approaches, the dataset played a pivotal role in constructing a state-of-the-art video matting framework, demonstrating its broad applicability in advancing portrait and video matting tasks.

The main contributions of this paper are as follows.
\begin{itemize}

\item A prior was observed that the semi-transparent regions in a human alpha matte are connected. Based on this prior, a cost-effective method was developed for constructing portrait matting datasets, enabling the rapid generation of large-scale, high-quality matting data.

\item The LD-Portrait-20K dataset was constructed, and its superior quality and effectiveness in portrait matting were validated through comparative experiments with publicly available datasets.

\item By integrating the training results of LD-Portrait-20K on ViTMatte-B~\cite{yao2024vitmatte} with SAM2~\cite{ravi2024sam}, a simple video matting baseline was developed that achieves state-of-the-art performance.

\end{itemize}

\section{Related Work}
This section provides a review of previous work relevant to the study. It begins by examining existing methods for constructing matting datasets, highlighting the challenges they encounter. Subsequently, it explores current portrait matting datasets, analyzing their strengths, weaknesses, and limitations. Through this review, the study identifies the shortcomings in existing datasets, laying a solid foundation for understanding the background and establishing a comparative context for the proposed dataset construction method.

\subsection{Methods for Creating Matting Datasets}

Methods for creating matting datasets can generally be divided into two categories: manual annotation and automated generation.

\textbf{Manual annotation} is a common approach for obtaining high-quality matting datasets. This process involves manually delineating the boundaries between the foreground and background to generate the alpha matte of an image. Annotators typically use tools such as Adobe Photoshop or GIMP to carefully refine the boundaries, especially for complex edges like hair or fur, ensuring high precision. The primary advantage of manual annotation is its ability to produce highly accurate datasets, making it particularly suitable for tasks requiring detailed processing.

For example, Xu et al.~\cite{xu2017deep} manually annotated $543$ images with simple or solid-color backgrounds using Photoshop, generating alpha channels and pure foreground colors. Background images were randomly selected from MS COCO~\cite{lin2014microsoft} and Pascal VOC~\cite{everingham2010pascal} to create composites, resulting in the Adobe Image Matting (AIM) dataset, which contains $49,300$ training images and $1,000$ test images. This dataset includes complex matting scenarios such as hair, fur, and translucent objects, but the number of unique foregrounds is limited to about $250$, resulting in a lack of diversity. To address this issue, Yu et al.~\cite{qiao2020attention} constructed the Distinction-646 dataset, which includes $646$ different foregrounds. Using the same method as Xu, they generated $59,600$ training images and $1,000$ test images. However, despite the increased variety of foregrounds, these datasets still only include foreground annotations, and their limited scale restricts their diversity and usability.

Subsequent datasets aimed to improve diversity and realism. The AIM-500 dataset~\cite{li2021deep} was the first natural image matting test set, containing $500$ high-resolution real-world images annotated using professional software. Li et al.~\cite{li2022bridging} further introduced the AM-2k dataset, which consists of $2,000$ high-resolution animal images with $100$ images per category, showcasing different appearances and backgrounds. Each image has a short side exceeding $1,080$ pixels. Although these datasets enhance foreground diversity and realism by incorporating real-world backgrounds, the manual annotation process remains labor-intensive, costly, and difficult to scale. It requires significant expertise to maintain annotation consistency, particularly for intricate edges and semi-transparent regions. Annotating video datasets presents even greater challenges, as it must account for dynamic lighting and complex foreground-background interactions.

\textbf{Automated methods} leverage algorithms to directly compute Ground Truth data under specific conditions. Compared to manual annotation, automated methods are more cost-effective and efficient. For instance, the early alphamatting.com dataset~\cite{rhemann2009perceptually} used triangulation in a controlled studio environment to generate high-quality alpha mattes. However, the complexity of the process limited the dataset to only $27$ training images and 8 test images. DAPM~\cite{shen2016deep} employed KNN matting~\cite{chen2013knn} and closed-form matting to produce a dataset of $2,000$ low-resolution portrait images, but the dataset quality was constrained by the method's limitations.

Chroma keying matting is another common automated method, frequently used for video matting datasets. Zhang et al.~\cite{zhang2021attention} collected video footage with green-screen backgrounds and used chroma keying matting to extract $108$ video clips and their corresponding alpha mattes. Lin et al.~\cite{lin2021real} collected $484$ high-resolution green-screen videos and applied chroma keying matting to generate $240,709$ unique alpha channels and foreground images, covering a wide range of individuals, clothing, and poses. They also compiled a portrait matting dataset containing 13,655 images, which is larger than other publicly available datasets. However, the quality of the alpha mattes obtained through chroma keying is often inconsistent.

Recently, researchers have begun exploring the use of generative models to automate the creation of matting datasets. Burgert et al.~\cite{burgert2024magick} pioneered this approach by using generative models to synthesize foreground images, compositing them onto solid-color backgrounds, and applying chroma keying matting or Photoshop to compute alpha mattes. However, this approach still fundamentally relies on traditional techniques. Zhang et al.~\cite{zhang2024transparent} proposed a method for simultaneously generating images and their corresponding alpha values. However, the quality of the generated alpha mattes was insufficient for direct use in training.

In summary, existing methods for creating matting datasets struggle to balance efficiency and quality. Developing an efficient and scalable approach to produce large-scale, high-quality datasets remains an important challenge.

\subsection{Portrait Matting Datasets}

Portrait matting, as a crucial task in image matting, plays an essential role in applications such as virtual try-on, background replacement in online meetings, and intelligent image editing. Compared to general object matting, portrait matting imposes more stringent requirements on datasets, as it must handle complex edge details (e.g., hair and translucent objects) while accommodating various poses and diverse backgrounds. High-quality portrait matting datasets are fundamental for training and evaluating matting models. However, existing datasets face limitations in scale, annotation quality, and diversity.

The AIM dataset~\cite{xu2017deep} contains $211$ portrait foreground images, most of which are consecutive video frames, resulting in limited diversity. To address this limitation, Late Fusion~\cite{zhang2019late} combined $228$ portrait images from the internet with the $211$ images from AIM, creating a new dataset. Similarly, the Distinction-646 dataset~\cite{qiao2020attention} includes $364$ portrait images, but the scale remains relatively small.

The P3M-10K dataset~\cite{ma2023rethinking} introduced the first privacy-preserving portrait matting dataset by collecting internet images and applying facial blurring, resulting in $9,421$ training images and $500$ test images. Sun et al.~\cite{sun2023ultrahigh} focused on high-resolution portrait matting and developed the HHM50k dataset, which includes $52,887$ high-resolution images. Additionally, several large-scale portrait matting datasets have been developed but remain unpublished~\cite{chen2018semantic,lin2021real,sengupta2020background,liu2023end}.

In the domain of video portrait matting, VideoMatting108\cite{zhang2021attention} and VideoMatte240K\cite{lin2021real} are two essential datasets. VideoMatting108 consists of $108$ video clips, each providing multiple frames of foregrounds along with their corresponding alpha mattes, making it suitable for training video matting models. VideoMatte240k, on the other hand, is a large-scale dataset comprising $240,709$ frames from $400$ high-quality video clips, offering extensive annotations of foregrounds and transparency. It is widely used for training and evaluating efficient video matting models.

Finally, some datasets serve as benchmarks, such as PhotoMatte85~\cite{lin2021real}, which includes $85$ portrait foreground images and their corresponding alpha channels. Ke et al.~\cite{ke2022modnet} introduced PPM-100, a validation dataset containing $100$ real-world portrait images with foreground and background annotations. Subsequent experiments compare the constructed dataset with publicly available portrait matting datasets and perform quantitative evaluations on these benchmarks.

\begin{figure*}[ht]
     \centering
    \includegraphics[scale=.22]{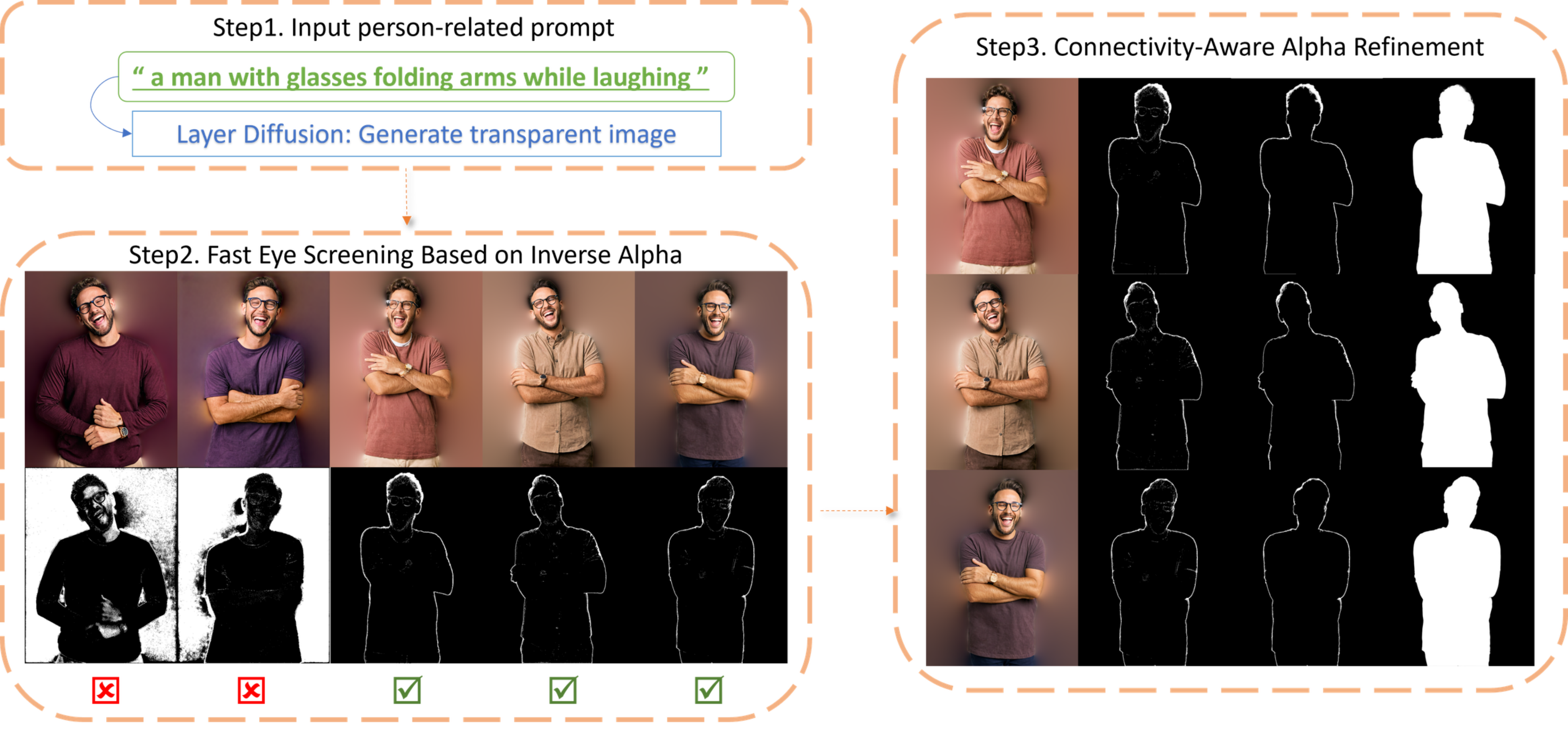}
    \caption{\textbf{The framework of the proposed method}. The first row of images in Step 2 shows the portrait images generated by Layer Diffusion, while the second row displays their corresponding inverse alpha. In Step 3, the first column shows the RGB images, the second column displays the initial inverse alpha, the third column shows the optimized inverse alpha, and the fourth column presents the final alpha matte.}
    \label{figure-method-1}   
\end{figure*}

\section{Method and Dataset}

This section introduces a general method for constructing portrait matting datasets and provides detailed information about the LD-Portrait-20K dataset. The overall framework of the proposed method is illustrated in Figure~\ref{figure-method-1} and comprises three main steps. First, a descriptive prompt about human is input into Layer Diffusion to generate the corresponding RGB images and alpha mattes. Next, the generated images undergo a preliminary screening process to remove outliers. Finally, the screened images are processed through Connectivity-Aware Alpha Refinement to optimize the alpha mattes, resulting in high-quality outputs. The details of each step are explained in the following subsections.

\textbf{Definitions.} In the grayscale images used in this paper, the range of alpha matte values is $[0, 255]$, where $0$ represents the background region, and $255$ represents the foreground region. For an image $I$ with an alpha matte $\alpha$, % we define 
its inverse alpha $\bar{\alpha}$ is defined as follows: pixels with values of $0$ or $255$ in $\alpha$ are assigned a value of $0$ in $\bar{\alpha}$, while all other pixels are assigned a value of $255$ in $\bar{\alpha}$. As a result, $\bar{\alpha}$ corresponds to all the semi-transparent regions in $\alpha$.

\subsection{Input Prompt}
Layer Diffusion is a latent diffusion model designed to generate transparent images by encoding alpha channel transparency into the latent manifold of the pretrained model. By providing appropriate prompts, the model can generate images containing transparency information and even multiple transparent layers simultaneously. This study employs Layer Diffusion to automatically generate portrait RGB images and their corresponding alpha mattes based on the given prompts.

To generate person-related prompts, two approaches were adopted. One involves automated generation by systematically traversing various characteristics of human, and the other utilizes ChatGPT to generate diverse and creative prompts.

\subsubsection{Automated Prompt Generation via Characteristic Traversal}
The first approach focuses on systematically describing various attributes of a person to automate the prompt generation process. These attributes include gender, actions, emotions, age, occupation, whether the person wears glasses or hats, hair length and color, clothing style, and type of attire. A few examples are as follows.
\begin{itemize} \small
{\sf
    \item A young man with black short hair, wearing casual jeans, striking a pose.
    \item A senior woman with blonde long hair and glasses, wearing a formal blouse, looking thoughtful.
    \item An elderly person with white hair, wearing an elegant dress, smiling.
}
\end{itemize}

By systematically traversing these characteristics, this approach ensures that the prompts are diverse and accurately represent a wide range of human features and states.

\subsubsection{Enriching Prompts Using ChatGPT}

To further enhance the diversity of the prompts, ChatGPT was employed to generate additional descriptive scenarios. Based on the provided descriptions, ChatGPT automatically created prompts, enriching the dataset with varied and imaginative examples. Examples of prompts generated by ChatGPT are as follows.

\begin{itemize}
{\small\sf
    \item An elderly man taking off his hat with a thoughtful expression.
    \item An elderly woman placing hands on hips while appearing confused.
    \item A person with a beard clapping hands while appearing confused.
}
\end{itemize}

By leveraging ChatGPT, the range of generated prompts was expanded to include more nuanced character traits, emotional expressions, and scenarios, significantly improving the diversity and representativeness of the dataset.

\subsection{Fast Eye Screening Based on Inverse Alpha}

\begin{figure*}
     \centering
    \includegraphics[scale=.264]{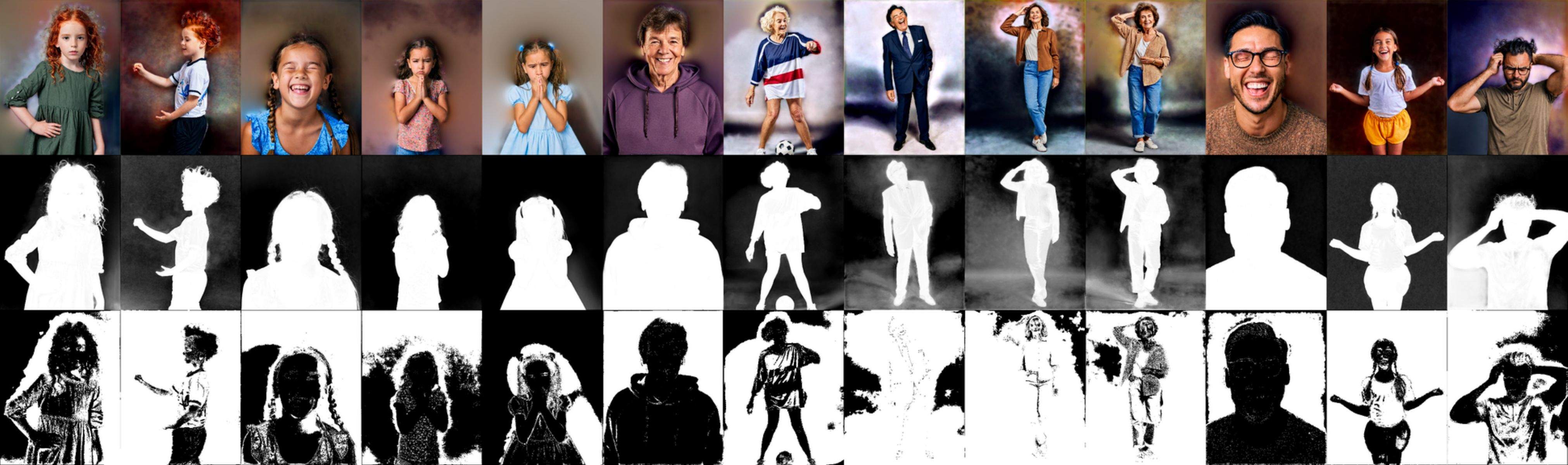}
    \caption{\textbf{These are examples to be deleted}. The first row shows the generated RGB images, the second row displays their corresponding alpha mattes, and the third row illustrates the inverse alpha values.}
    \label{figure-method-2}   
\end{figure*}

The foreground RGB images and their corresponding alpha mattes were manually screened to remove samples with significant deviations. As shown in Figure~\ref{figure-method-2}, these images contained numerous erroneous alpha values, primarily located in the background region and directly connected to the semi-transparent areas along the human edges. This connection not only made it difficult to accurately distinguish whether each pixel belonged to the foreground or background, but also introduced substantial interference into the subsequent training process. Additionally, processing these heavily flawed images would have significantly increased computational resource requirements. Therefore, problematic images were discarded, and only those where the semi-transparent areas were primarily concentrated along the human edges were retained. Several examples are shown in Figure~\ref{figure-method-3}, illustrating images that better represent the semi-transparent characteristics of human edges and provide a more reliable foundation for subsequent optimization steps.

\subsection{Connectivity-Aware Alpha Refinement}
For the images that remain after preliminary screening, the next step is to identify which alpha values need to be preserved and which need to be corrected. A straightforward idea involves employing a trimap, where alpha values corresponding to the foreground pixels are set to $255$, those for the background pixels are set to $0$, and the alpha values in the semi-transparent regions are retained. While this approach can partially optimize the generated alpha matte, creating an accurate trimap is both time-consuming and challenging, particularly when defining the boundaries of semi-transparent regions, such as human edges with intricate details like hair. As a result, this method fails to correct the alpha matte effectively.

The proposed method combines inverse alpha and introduces an observed prior that \textit{in human alpha mattes, semi-transparent regions consistently appear along the edges of the body, these regions naturally form multiple connected areas, with each connected area corresponding to a distinct semi-transparent segment}. Specifically, this region typically results from a gradual transition between the foreground and background, where transparency changes smoothly and continuously without abrupt jumps or breaks. This ensures that the alpha values along the edge region form a spatially coherent gradient. Furthermore, human edges often appear naturally blurred, especially in scenarios with complex backgrounds or varying lighting conditions. The transparency values in these regions reflect the interaction between the foreground and background, maintaining a seamless and consistent transition that avoids hard boundaries.

Based on the above ideas, Connectivity-Aware Alpha Refinement was designed to optimize the alpha matte by preserving the connected semi-transparent regions at the human edge in the inverse alpha and correcting erroneous values in other areas. The process leverages Breadth-First Search (BFS)~\cite{cormen2009introduction} and Depth-First Search (DFS)~\cite{knuth1997art} to achieve efficient processing. Given the alpha matte $\alpha$ and inverse alpha $\bar{\alpha}$ with width W and height H, the steps are as follows.  \\

\begin{itemize}
\item \textbf{Step 1:} Perform BFS on $\alpha$ to identify all connected background regions, denoted as $R_{\text{bg}}$. Each connected background region corresponds to one edge region of the human subject.
    \begin{equation}
        R_{bg}=BFS(\alpha)
    \end{equation}

    \item \textbf{Step 2:} Combine $R_{\text{bg}}$ with $\bar{\alpha}$ and perform DFS to locate the starting points of the semi-transparent regions along the human edge, denoted as $P_{\text{start}}$.
    \begin{equation}
        P_{start}=DFS(\bar{\alpha}, R_{bg})
    \end{equation}

    \item \textbf{Step 3:} Starting from each point in $P_{\text{start}}$, perform BFS on $\bar{\alpha}$ to identify all connected pixels, denoted as $R_{\text{semi}}$. Retain the alpha values of these connected pixels in $\alpha$ as valid.
    \begin{equation}
        R_{semi}=BFS(\bar{\alpha},P_{start})
    \end{equation}

    \item \textbf{Step 4:} Initialize the remaining noisy pixels in $\alpha$ to 255. Based on the connectivity property of the human alpha matte, use BFS to extract the connected human alpha matte $\alpha'$, which represents the optimized result.
\begin{equation}
    \alpha'=BFS(\alpha)
\end{equation}
\end{itemize}

\begin{figure*}
     \centering
    \includegraphics[scale=.23]{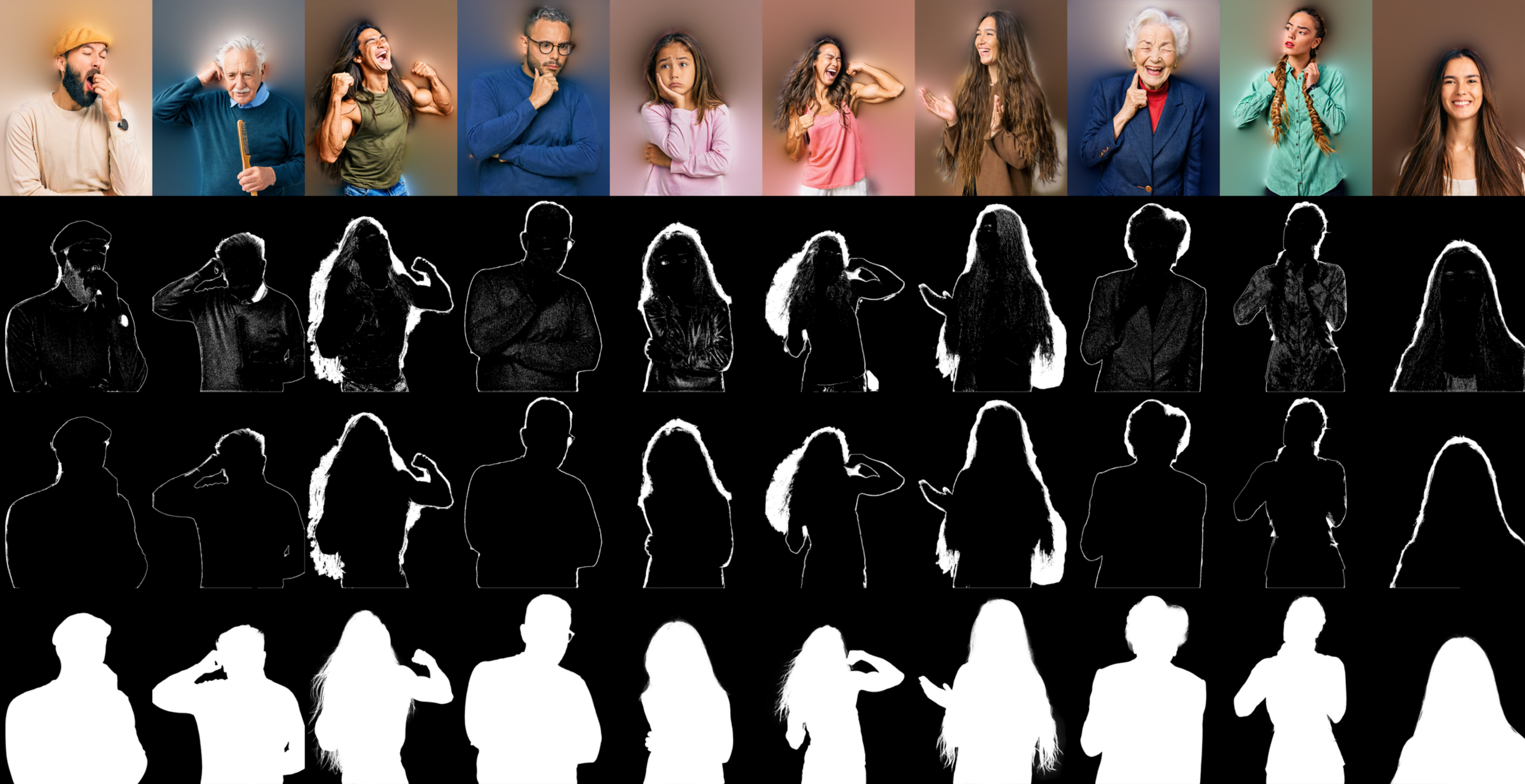}
    \caption{\textbf{Display of Connectivity-Aware Alpha Refinement results}. The first row shows the RGB portrait images, the second row displays the inverse alpha corresponding to the original alpha matte, and the third and fourth rows show the optimized inverse alpha and alpha matte, respectively.}
    \label{figure-method-3}   
\end{figure*}

By integrating BFS and DFS, this algorithm achieves both simplicity and computational efficiency, significantly improving the quality of the alpha matte. The optimized results are shown in Figure~\ref{figure-method-3}.

\subsection{The LD-Portrait-20K Dataset}

\begin{figure*}
     \centering
    \includegraphics[scale=.342]{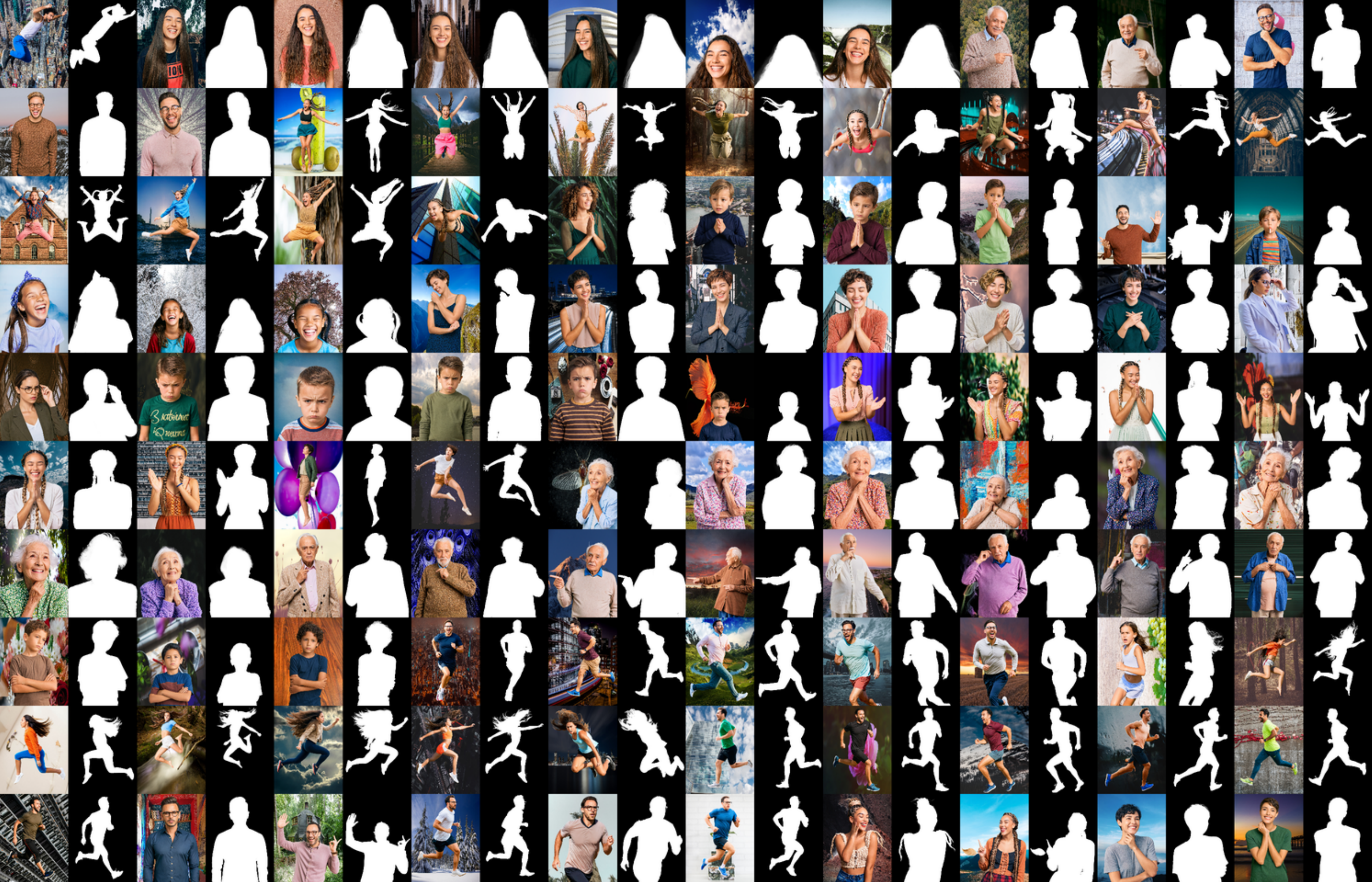}
    \caption{\textbf{Examples from the LD-Portrait-20K Dataset}. 100 examples from the LD-Portrait-20K dataset are presented, including RGB portrait images composited onto different backgrounds and their corresponding alpha mattes, demonstrating the diversity and representativeness of the dataset.}
    \label{figure-method-4}   
\end{figure*}

The portrait matting dataset LD-Portrait-20K was constructed following the outlined steps, comprising $20,051$ portrait RGB images and their corresponding alpha mattes. Examples of the dataset, including samples composited onto different backgrounds, are shown in Figure~\ref{figure-method-4}. The high quality of the foregrounds ensures seamless integration with various background images, highlighting the dataset's usability in diverse scenarios. During the dataset generation process, diverse attributes such as gender, age, pose, expression, and clothing, were deliberately incorporated to ensure diversity and representativeness. This approach not only captures a broad range of person-specific features and scene variations but also utilizes an efficient and controllable generation pipeline. It allows for flexible customization of specific attributes to meet different requirements, overcoming the labor-intensive and restrictive nature of manual annotation in traditional dataset construction. 
As a result, the method significantly enhances both the quality and efficiency of dataset creation. LD-Portrait-20K, together with the proposed dataset generation method, is expected to serve as a valuable resource for advancing portrait matting and video portrait matting tasks in future research.

\begin{table*}[ht]
% \large 
% \normalsize
\caption{Training of Different Datasets on ViTMatte-B and GFM and Comparison on PhotoMatte85 and PPM-100.\label{table1}}
\centering
% \Large
% \normalsize
\begin{threeparttable}
% \begin{tabular}{c|c|cccc|cccc}
\begin{tabular}{@{}lcccccccccc@{}}
\toprule
\multirow{2}{*}{Method} &\multirow{2}{*}{Training Dataset} & \multicolumn{4}{c}{PhotoMatte85}    & \multicolumn{4}{c}{PPM-100} \\
\cmidrule(lr){3-6}\cmidrule(lr){7-10} & & MAD  & MSE  & Grad  & Conn  & MAD  & MSE  & Grad  & Conn  \\
\midrule
\multirow{6}{*}{ViTMatte-B} 
& HHM50K &3.78	&0.87 &5.79 &3.96 &5.63	&1.31	&6.27	&4.10 \\
& P3M-10k &3.39 &0.57 &3.85 &3.03 &5.16	&1.31	&5.80	&\underline{3.53} \\
& ImageMatte & 3.26 &0.52 &3.85 &2.85 &\underline{5.08}	&\underline{1.19}	&5.57	&\underline{3.53} \\
& LD-Portrait-CK &3.44 &0.71 &4.87 &3.45 &5.79	&1.75	&7.97	&4.94 \\
& LD-Portrait-10K &\underline{3.20} &\underline{0.49} &\underline{3.56} &\underline{2.63} &5.16	&1.20	&\underline{5.33}	&3.77 \\
& LD-Portrait-20K & \textbf{3.13} & \textbf{0.42} & \textbf{3.05} & \textbf{2.42} &\textbf{5.06}	&\textbf{1.12}	&\textbf{5.15}	&\textbf{3.51} \\
\midrule
\multirow{7}{*}{GFM} 
& HHM50K &6.49 &4.25 &14.11 &12.02 &67.44 &63.40	&22.40 &41.59\\
& P3M-10k &11.05 &9.17 &14.27 &21.28 &\underline{17.61}	&\underline{13.75} &\underline{12.54} &\underline{9.97}\\
& ImageMatte &8.40 &6.58 &11.26 &15.86 &23.73 &19.68 &16.99	&15.40 \\
& LD-Portrait-CK &5.70 &4.08 &11.87 &10.77 &24.22 &21.96 &20.44 &21.85 \\
& LD-Portrait-Ori &8.94 &6.56 &11.48 &16.08  &43.52 &38.54 &19.17 &29.95 \\
& LD-Portrait-10K &\underline{4.78} &\underline{2.94} &\underline{10.52} &\underline{8.61} &17.76 &15.41 &17.14 &13.31 \\
& LD-Portrait-20K & \textbf{3.48} & \textbf{1.81} & \textbf{8.06} & \textbf{6.08} &\textbf{14.65} &\textbf{12.32} &\textbf{12.33}	&\textbf{9.4} \\

\bottomrule
\end{tabular}
\end{threeparttable} 
\end{table*}

\section{Experiments}
This section presents a comparison of LD-Portrait-20K with other datasets, an evaluation of the method's performance against the chroma keying algorithm, an ablation study on dataset size, and an examination of the practical value of the dataset in video matting tasks.

\textbf{Models.} Two publicly available image matting models were utilized: the trimap-based model ViTMatte-B~\cite{yao2024vitmatte} and the trimap-free model GFM~\cite{li2022bridging}, both widely used in image matting tasks. All experiments were conducted with a single RTX 3090 GPU to ensure consistency.

\textbf{Datasets.} This experiment compared the LD-Portrait-20K dataset with three existing portrait matting datasets:

\begin{itemize}
    \item HHM50K\cite{sun2023ultrahigh}. A high-resolution portrait matting dataset manually annotated using Photoshop, containing exactly $52,887$ portrait images with fine-grained alpha mattes.
    \item P3M-10K\cite{ma2023rethinking}. A portrait matting dataset containing $9,421$ images, specifically designed for privacy-preserving tasks.
    \item ImageMatte. A dataset inspired by RVM~\cite{lin2022robust}, constructed by combining AIM~\cite{xu2017deep} and Distinctions-646~\cite{qiao2020attention}, containing $420$ portrait images. Each image was randomly composited with $100$ background images from BG-20k~\cite{li2022bridging} to generate a total of $42,000$ composite images.
    \item LD-Portrait-20K. Each foreground image was randomly composited with $5$ background images from BG-20k~\cite{li2022bridging}, resulting in a total of $100,255$ composite images.
\end{itemize}

Three validation datasets were selected to evaluate the matting models trained on these datasets:

\begin{itemize}
    \item PhotoMatte85~\cite{lin2021real}. This dataset contains $85$ portrait foreground images and their corresponding alpha mattes. Each image is randomly fused with $10$ background images from BGMv2~\cite{lin2021real} to create a total of $850$ test images.
    \item AIM~\cite{xu2017deep}. A dataset of $10$ portrait images, each randomly fused with $10$ background images from BGMv2~\cite{lin2021real}, producing $100$ test images.
    \item PPM-100~\cite{ke2022modnet}. A validation dataset specifically designed for portrait matting, containing $100$ real-world portrait images with their corresponding alpha mattes.
\end{itemize}

To accommodate GPU memory constraints, PhotoMatte85 was resized to half of its original size and PPM-100 to one-third of its original size. For a fair comparison with state-of-the-art portrait matting models~\cite{wang2024eformer}, AIM images were resized to $512\times 512$.

\textbf{Evaluation.} To quantitatively evaluate the matting results, four commonly used metrics were selected: Mean Absolute Difference (MAD), Mean Squared Error (MSE), Gradient Loss (Grad), and Connectivity Loss (Conn)\cite{rhemann2009perceptually}. Lower values of these metrics indicate higher quality of the alpha mattes. For easier comparison, the MAD, MSE, Grad, and Conn metrics were scaled by $10^3$, $10^3$, $10^{-3}$, and $10^{-3}$, respectively. Given that only the AIM dataset includes trimap annotations, the metrics were computed exclusively within the unknown regions of the AIM dataset for the ViTMatte-B training results. For PhotoMatte85 and PPM-100, however, the metrics were calculated across the entire image\cite{xu2017deep,qiao2020attention,yu2021mask,chen2013knn,li2020natural,cai2022transmatting,qiao2023hierarchical}.

\subsection{Evaluation of Dataset Effectiveness}

To validate the effectiveness of the proposed dataset, LD-Portrait-20K was trained alongside three comparison datasets on both ViTMatte-B and GFM models, with comparative testing conducted on the three aforementioned validation datasets. To ensure a fair comparison, the number of training epochs was adjusted based on dataset size: $10$ epochs for LD-Portrait-20K, $20$ for HHM50K, $100$ for P3M-10k, and $20$ for ImageMatte, across both models.

Table~\ref{table1} and Table~\ref{table2} present the quantitative comparison results. The best results are highlighted in bold, and the second-best results are underlined. Across the three validation datasets, models trained on LD-Portrait-20K consistently outperform those trained on other datasets across most evaluation metrics. Notably, the LD-Portrait-20K-trained model achieves significantly better Grad metric values in the majority of comparisons, demonstrating its capability to enhance matting performance, particularly in capturing fine details and handling complex scenarios. These results establish LD-Portrait-20K as a robust dataset for improving the quality and reliability of matting models.

Table~\ref{table2} highlights testing results on the AIM validation set, including comparisons with the current state-of-the-art portrait matting model, EFormer~\cite{wang2024eformer}. As EFormer’s code and weights are not publicly available, the experimental setup described in its paper, including resolution and dataset selection, was strictly adhered to, and its reported results were referenced directly. The ViTMatte-B model trained on LD-Portrait-20K outperformed EFormer across multiple metrics, further demonstrating the significant value of the proposed dataset in advancing portrait matting research.

\begin{figure*}
    \centering
    \subfloat{
    % \hfill
    \begin{minipage}[t]{0.11\textwidth}
     \centerline{\includegraphics[scale=.1]{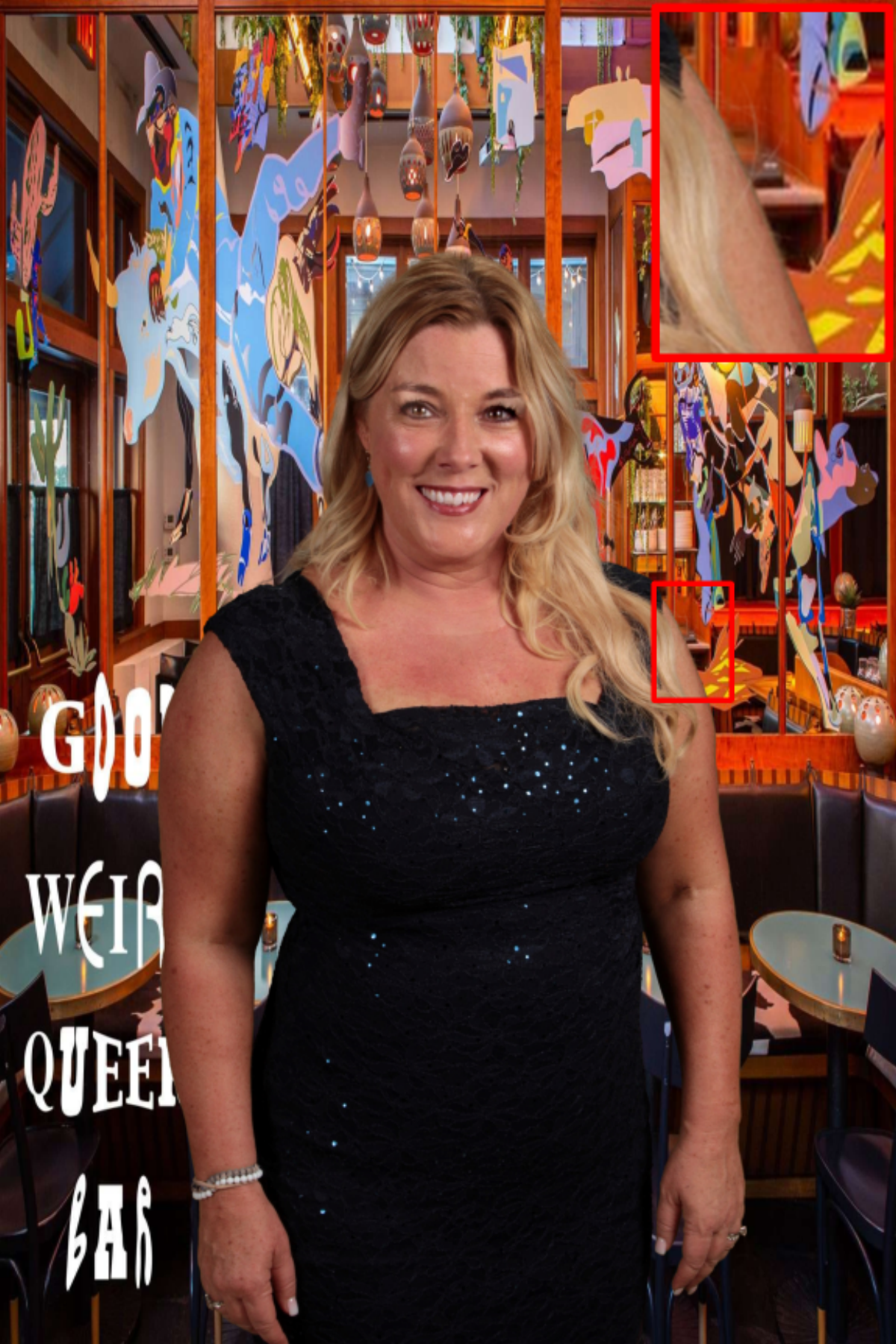}}\vspace{2pt}   
     \centerline{\includegraphics[scale=.1]{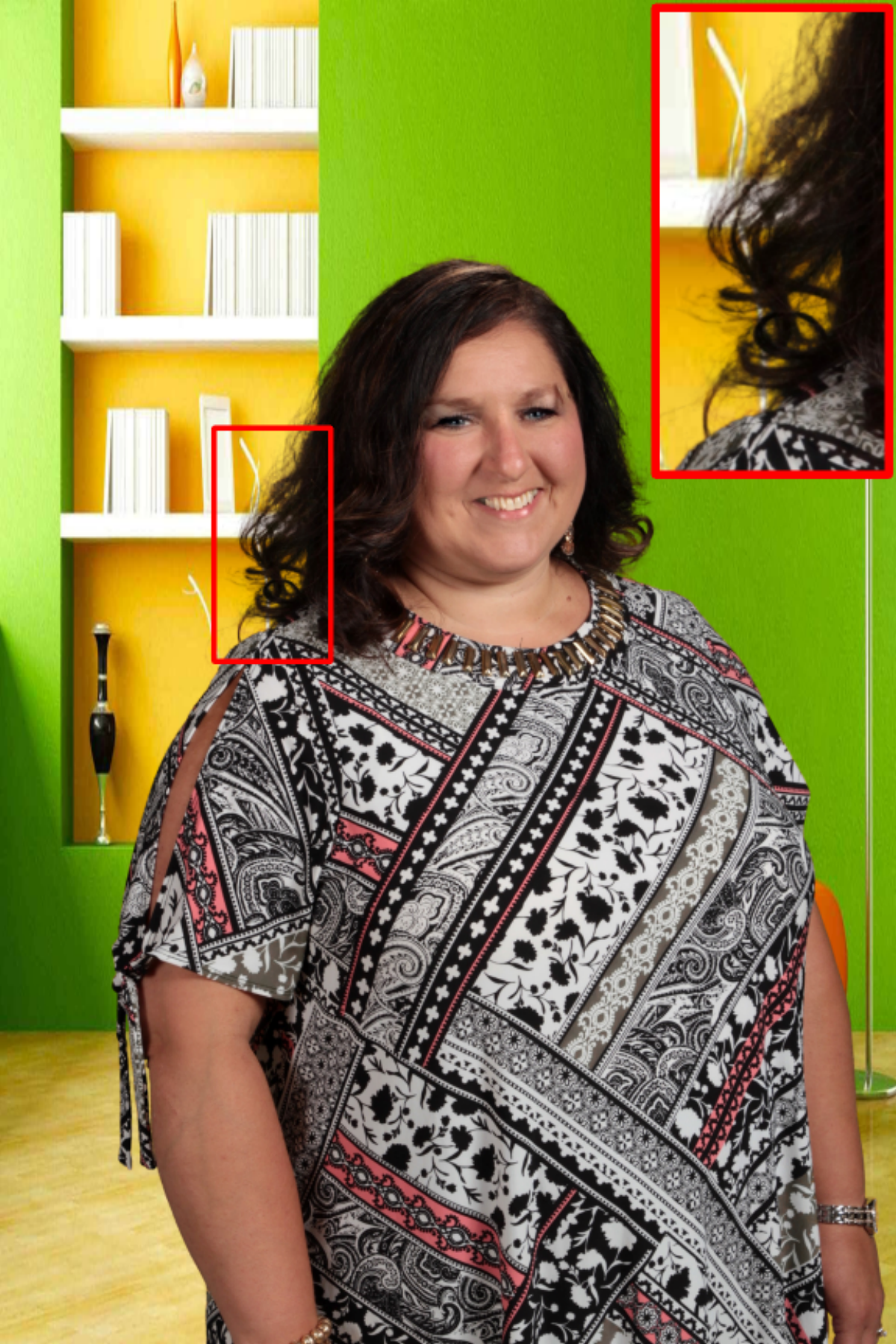}}\vspace{2pt} 
     \centerline{\includegraphics[scale=.1]{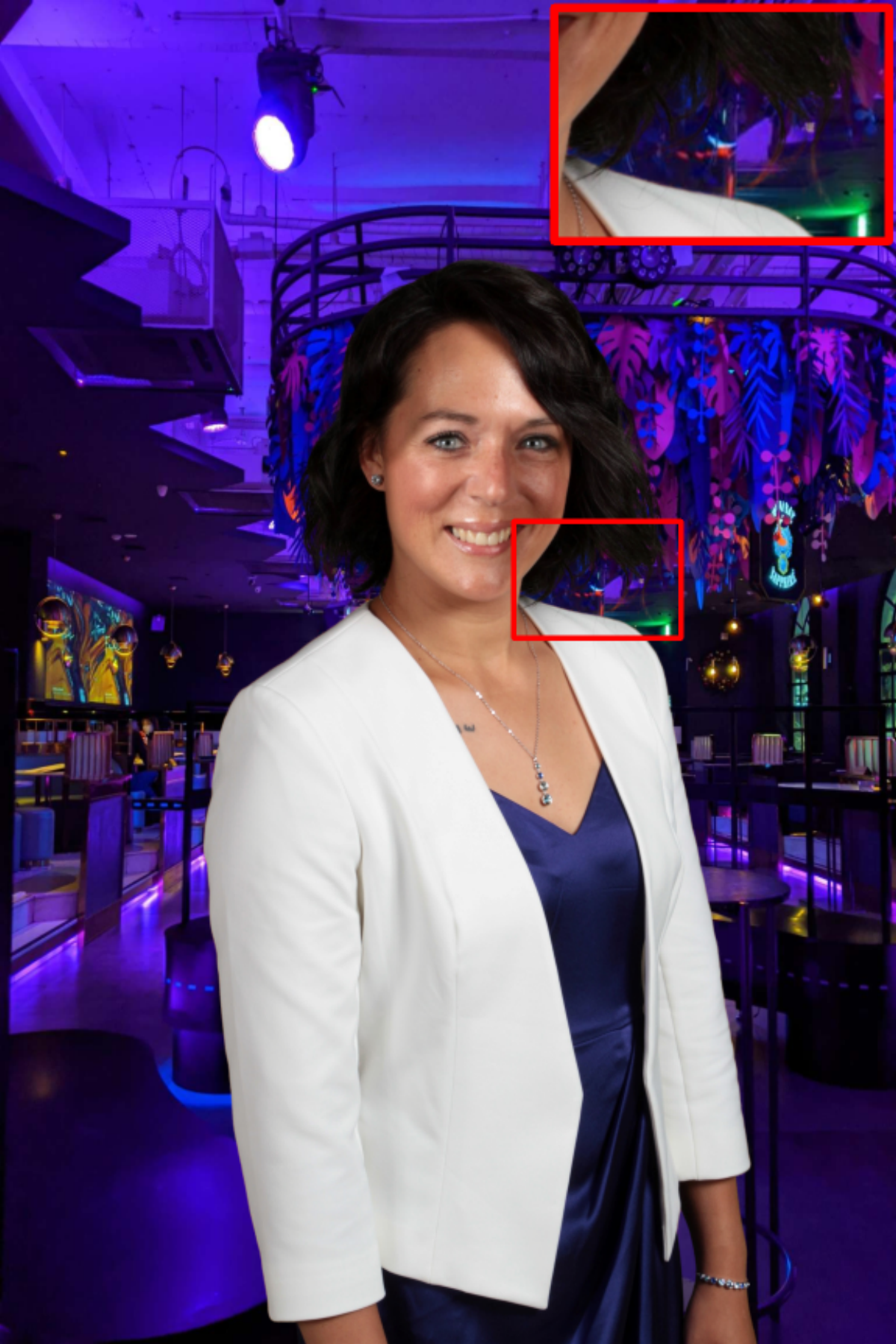}}\vspace{2pt}   
     % \centerline{\includegraphics[scale=.1]{figures/experiment/PH85-4/4_color.pdf}}\vspace{2pt}  
     \centerline{\includegraphics[scale=.1125]{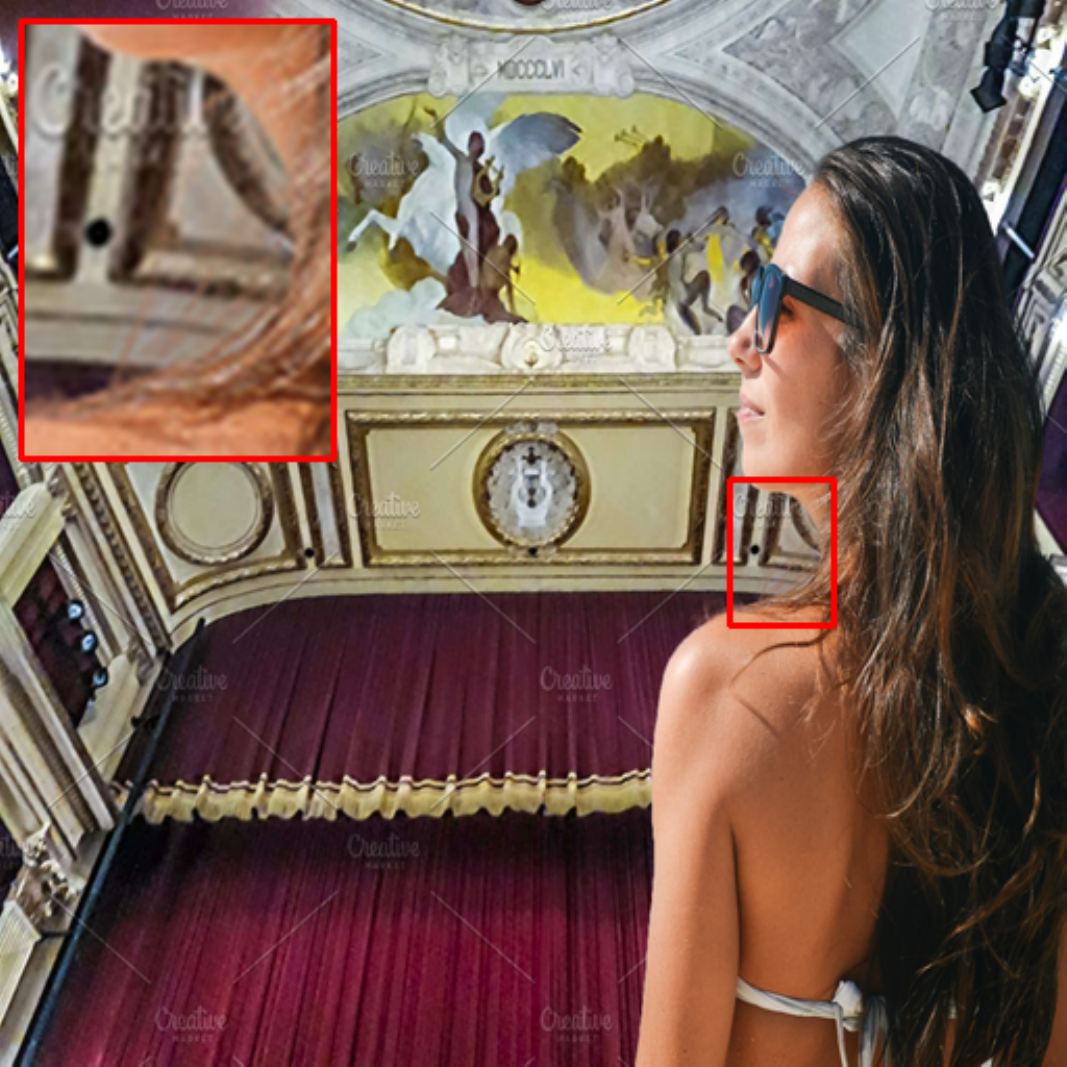}}\vspace{2pt}   
     \centerline{\includegraphics[scale=.1125]{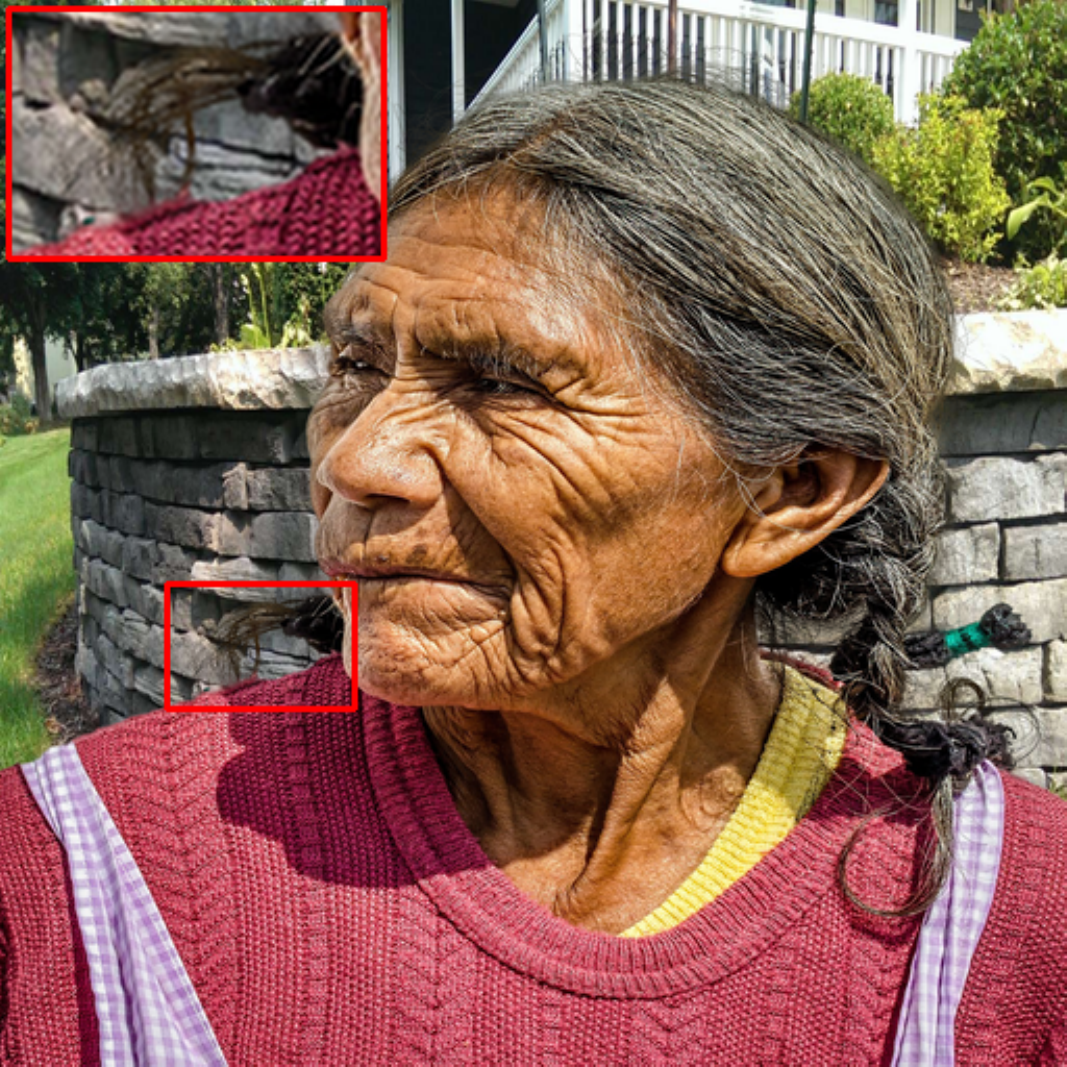}}\vspace{2pt} 
     \centerline{\includegraphics[scale=.1125]{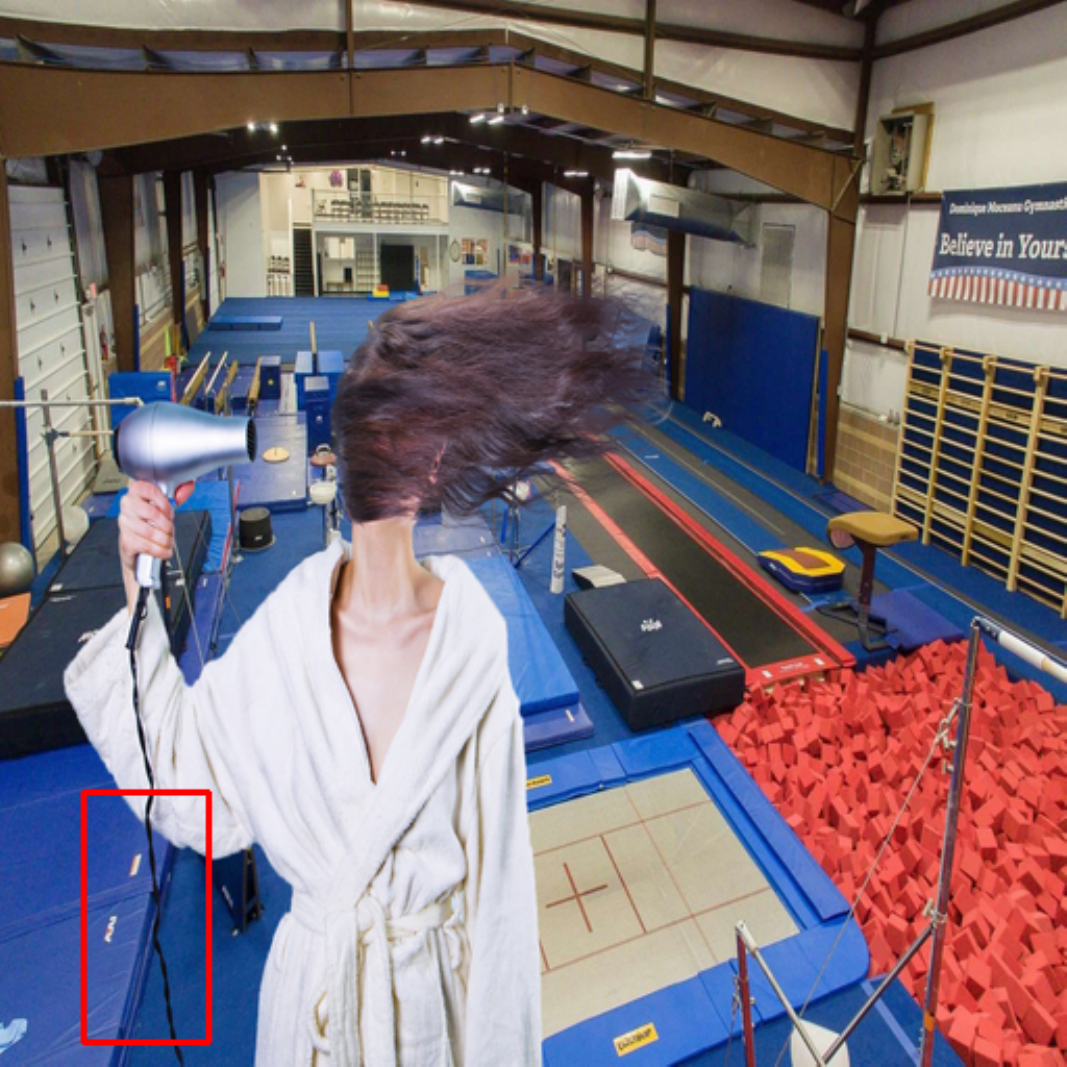}}\vspace{2pt}   
     \centerline{\includegraphics[scale=.1689]{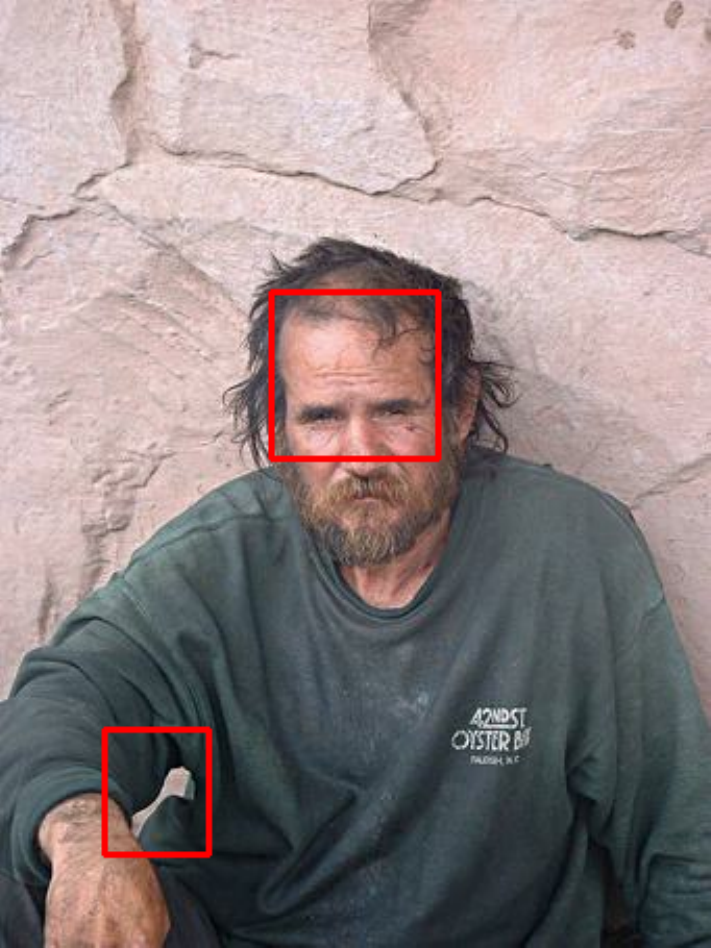}}\vspace{2pt}
     \centerline{\includegraphics[scale=.0473684]{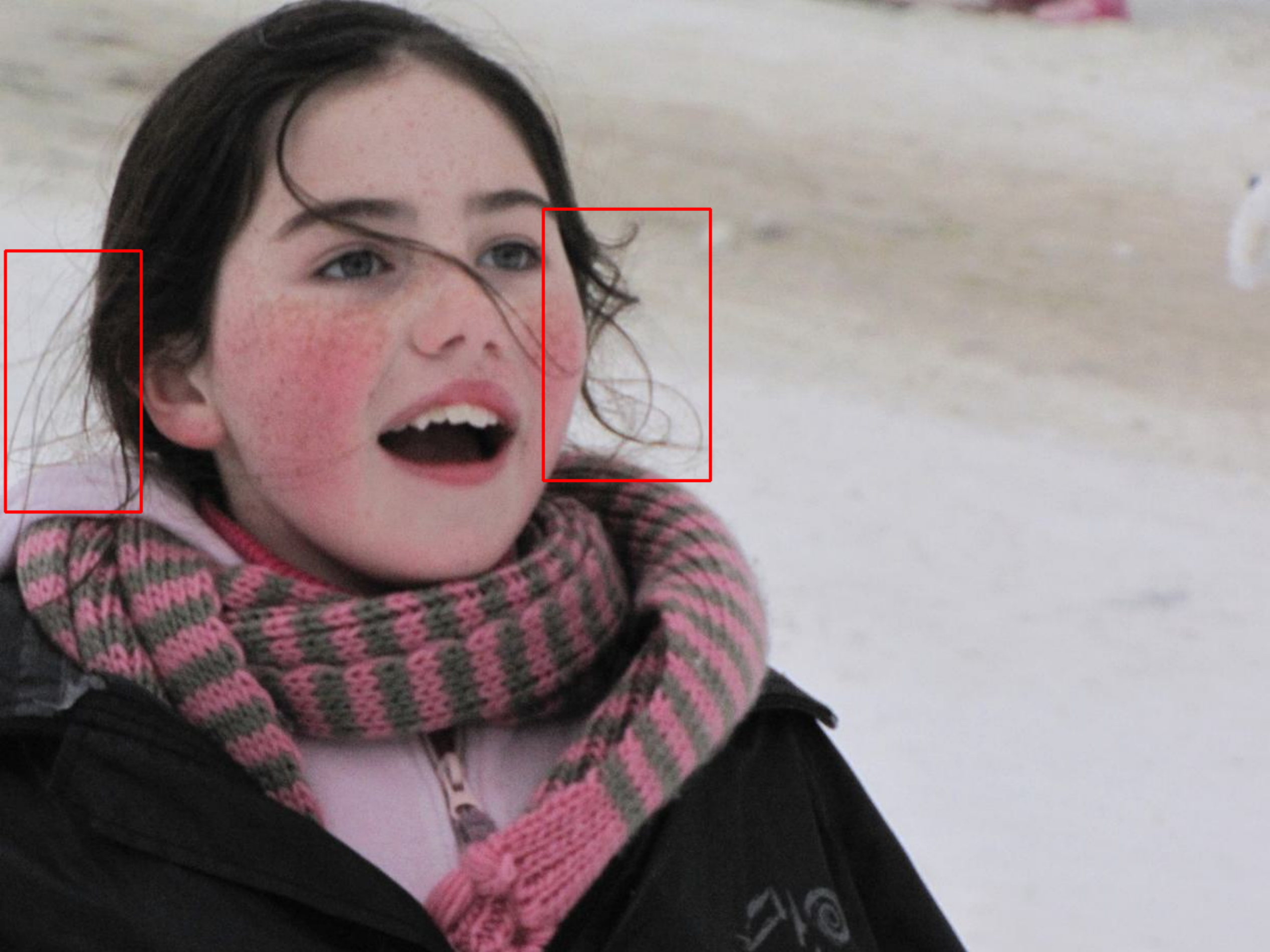}}\vspace{2pt}
     \centerline{\includegraphics[scale=.1056880734]{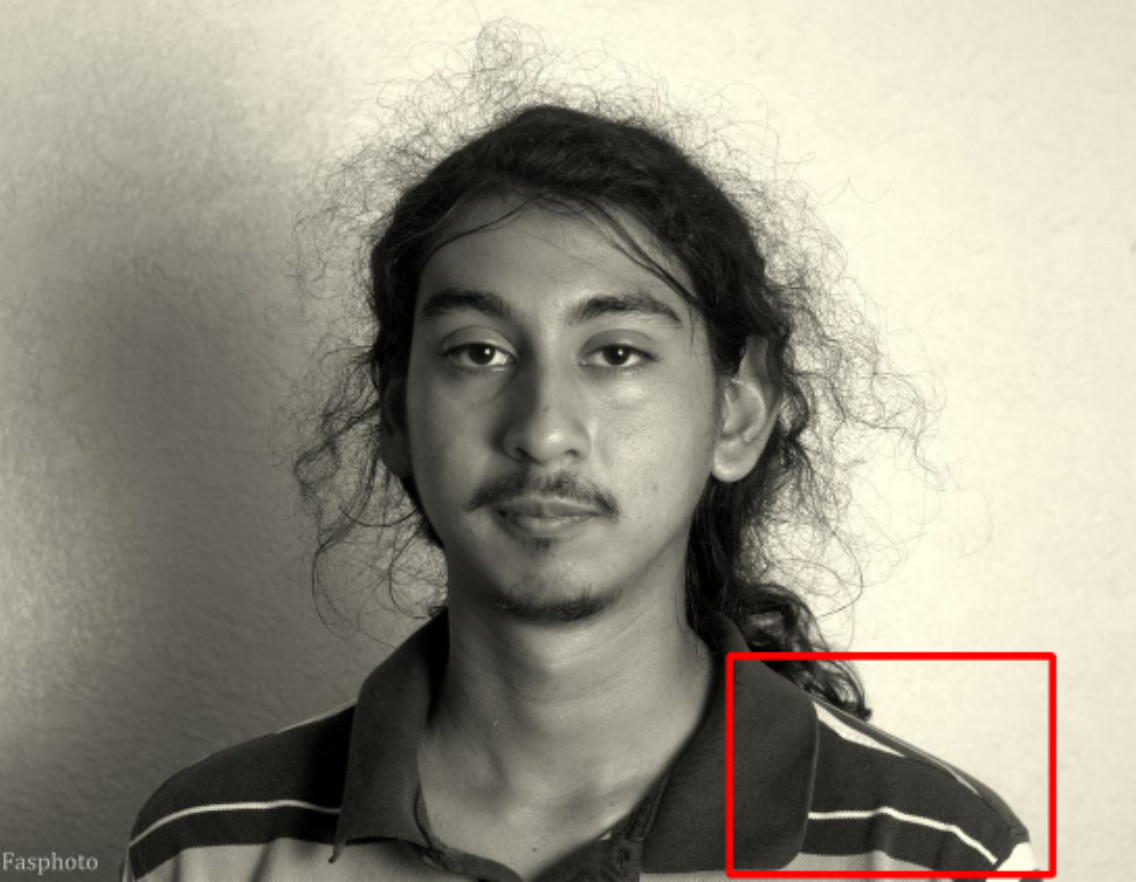}}  
     \centerline{Image}
    \end{minipage}
    \hfill
    \begin{minipage}[t]{0.11\textwidth}
    \centerline{\includegraphics[scale=.1]{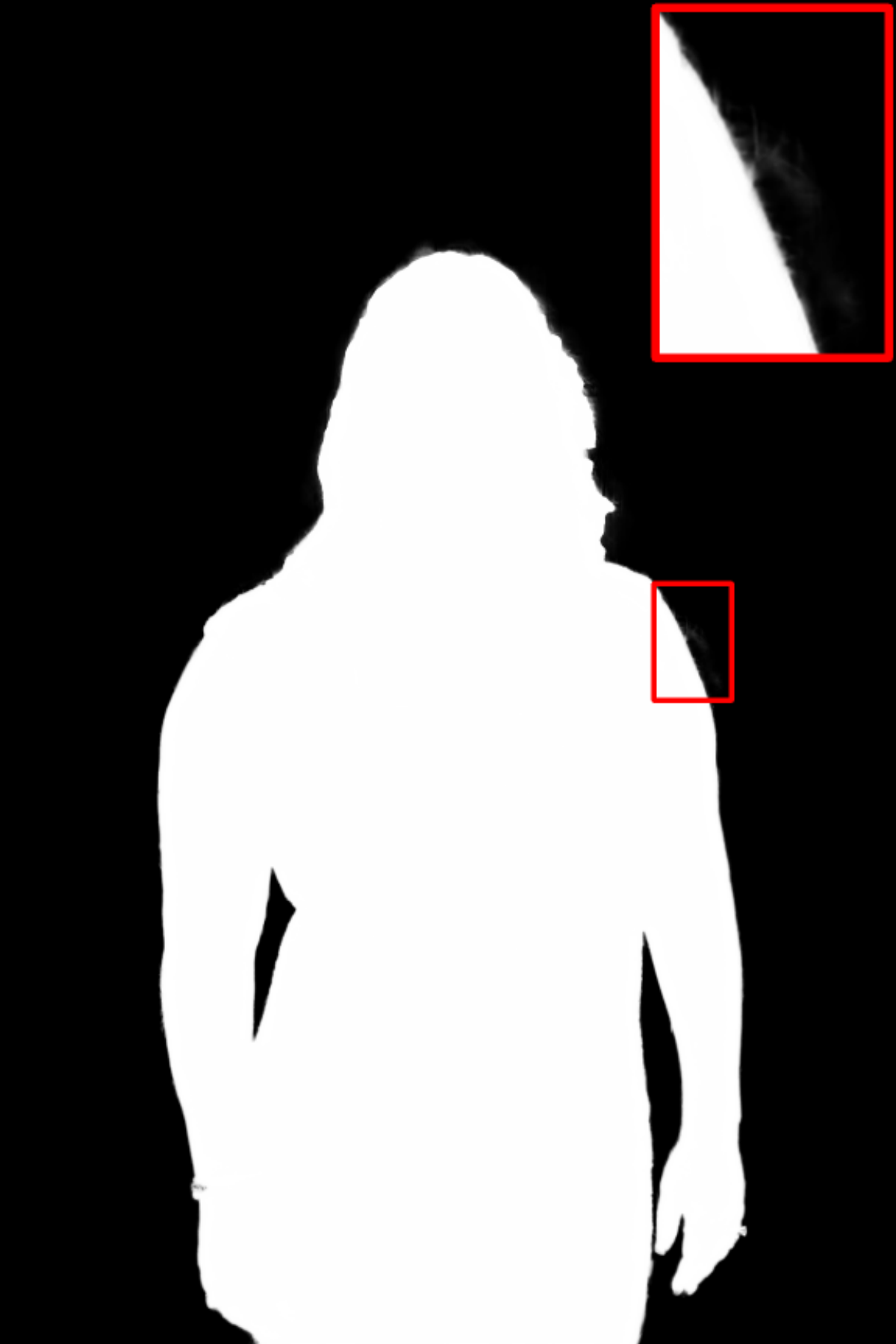}}\vspace{2pt}   
     \centerline{\includegraphics[scale=.1]{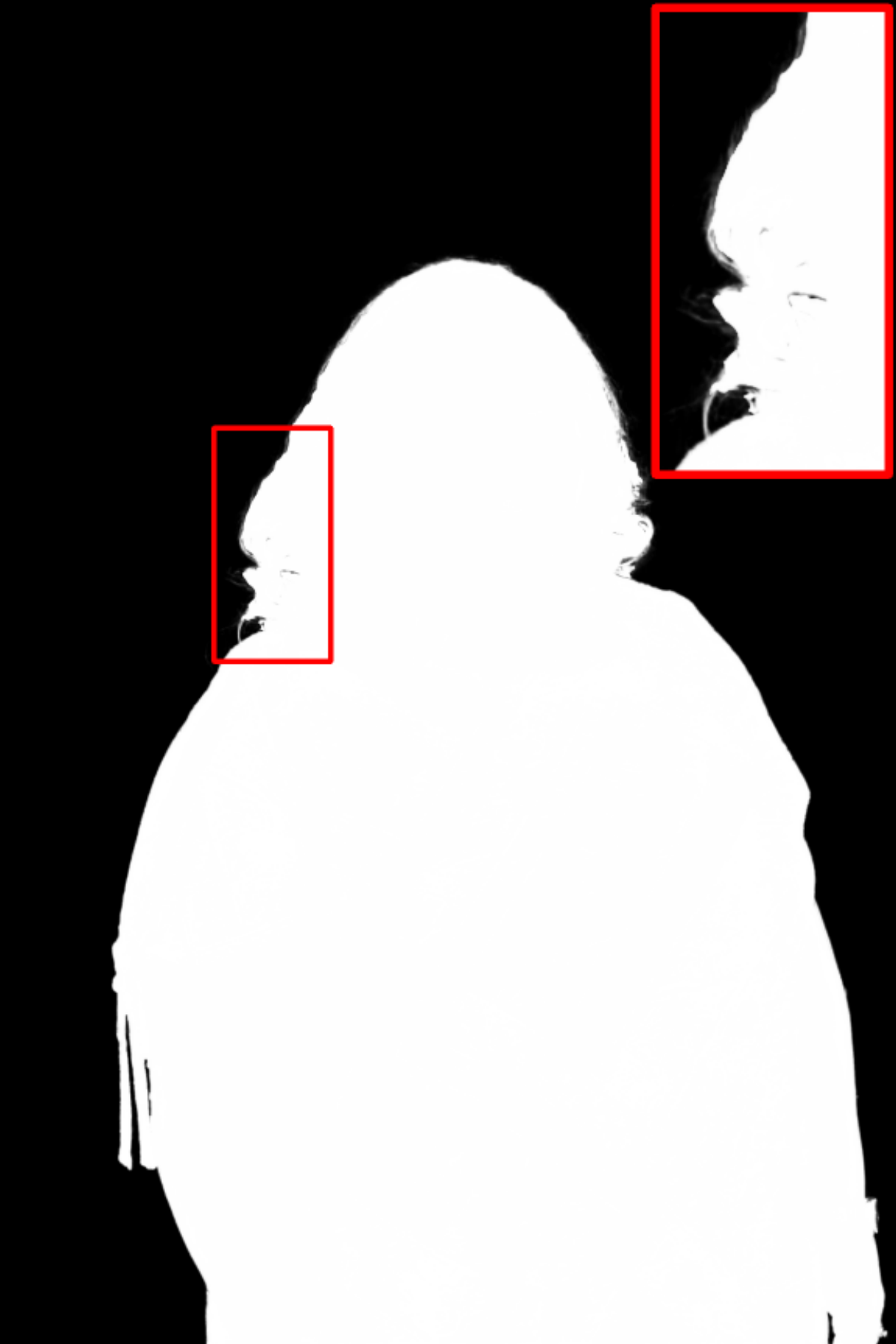}}\vspace{2pt} 
     \centerline{\includegraphics[scale=.1]{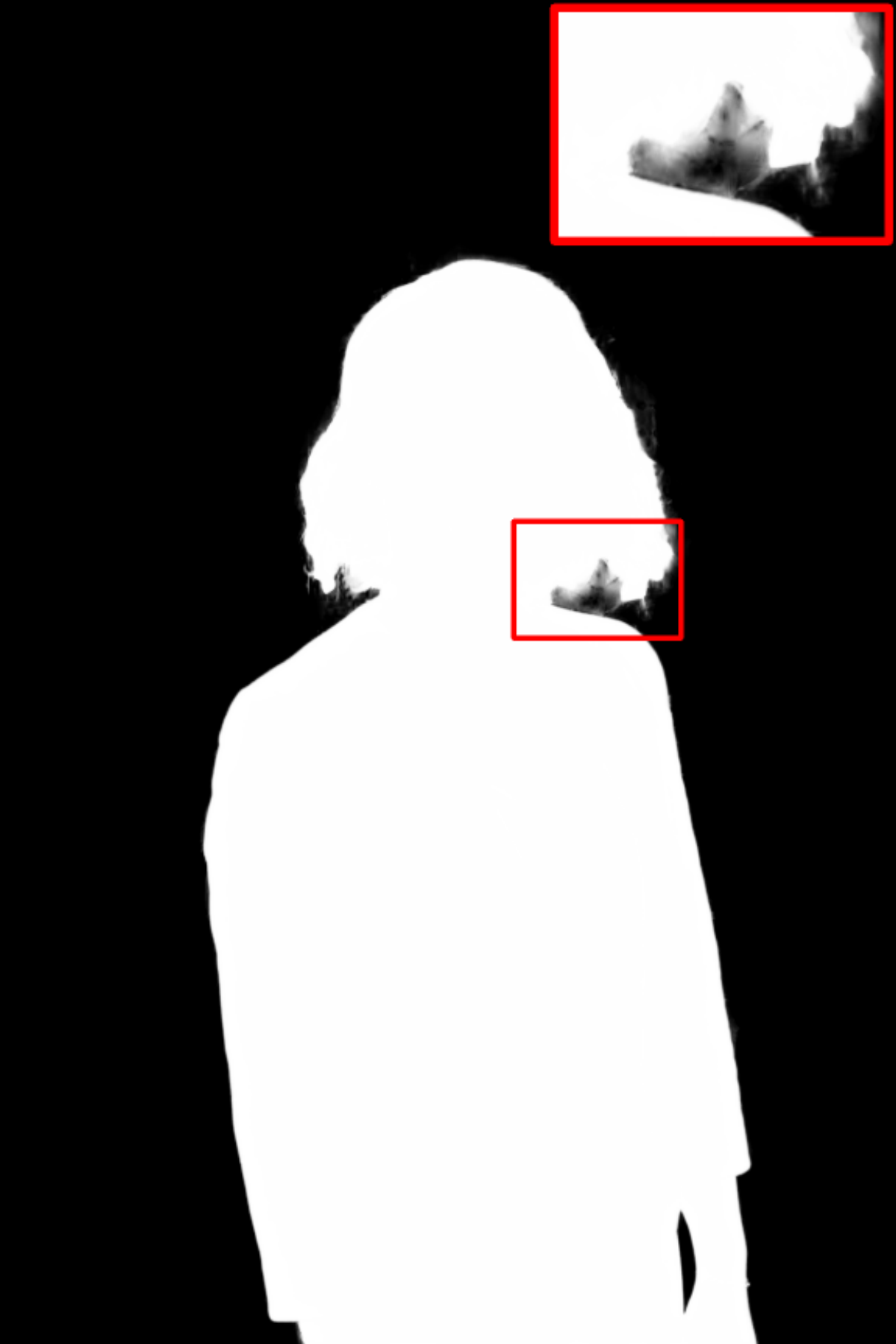}}\vspace{2pt}   
     % \centerline{\includegraphics[scale=.1]{figures/experiment/PH85-4/4_HHM50K.pdf}}\vspace{2pt}  
     \centerline{\includegraphics[scale=.1125]{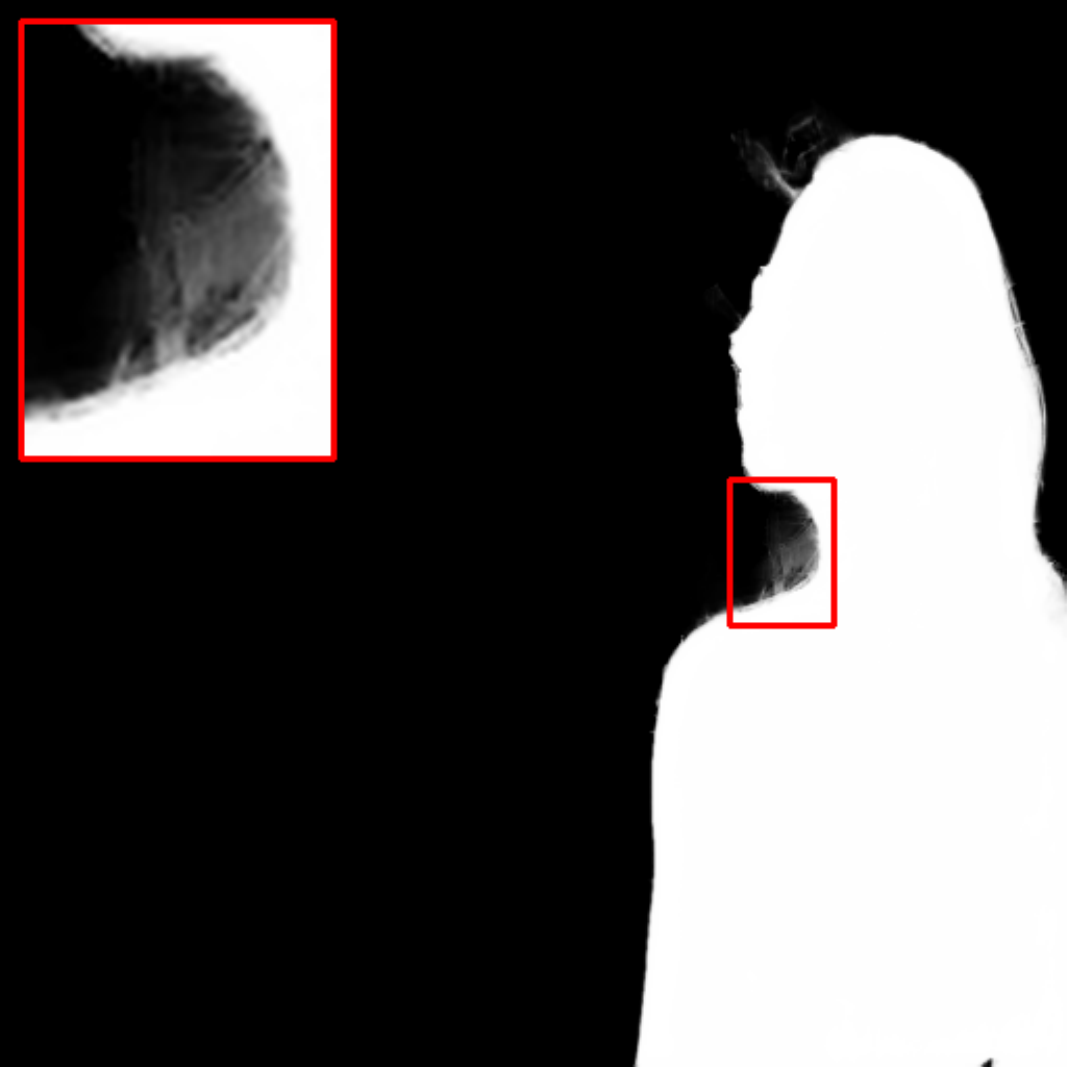}}\vspace{2pt}   
     \centerline{\includegraphics[scale=.1125]{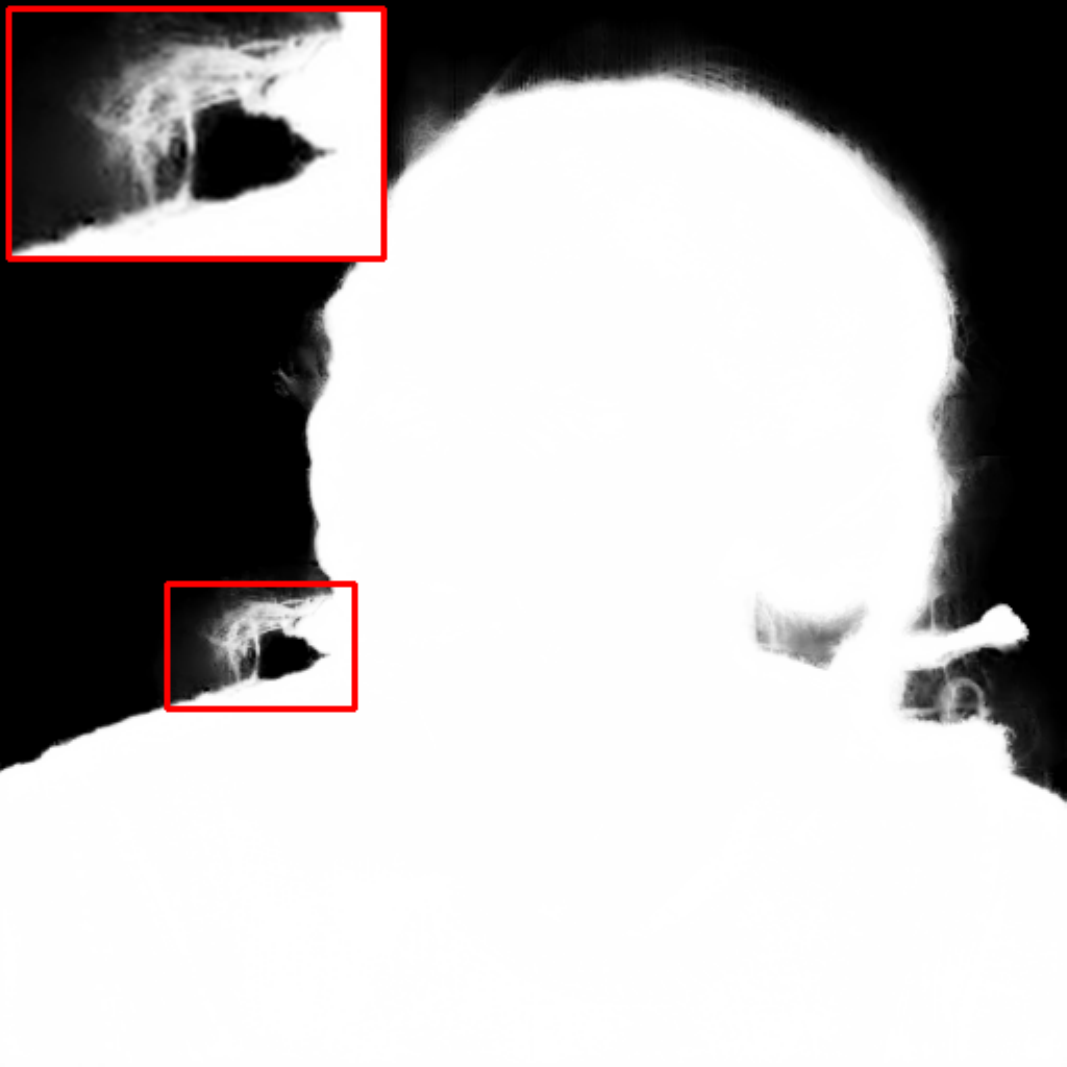}}\vspace{2pt} 
     \centerline{\includegraphics[scale=.1125]{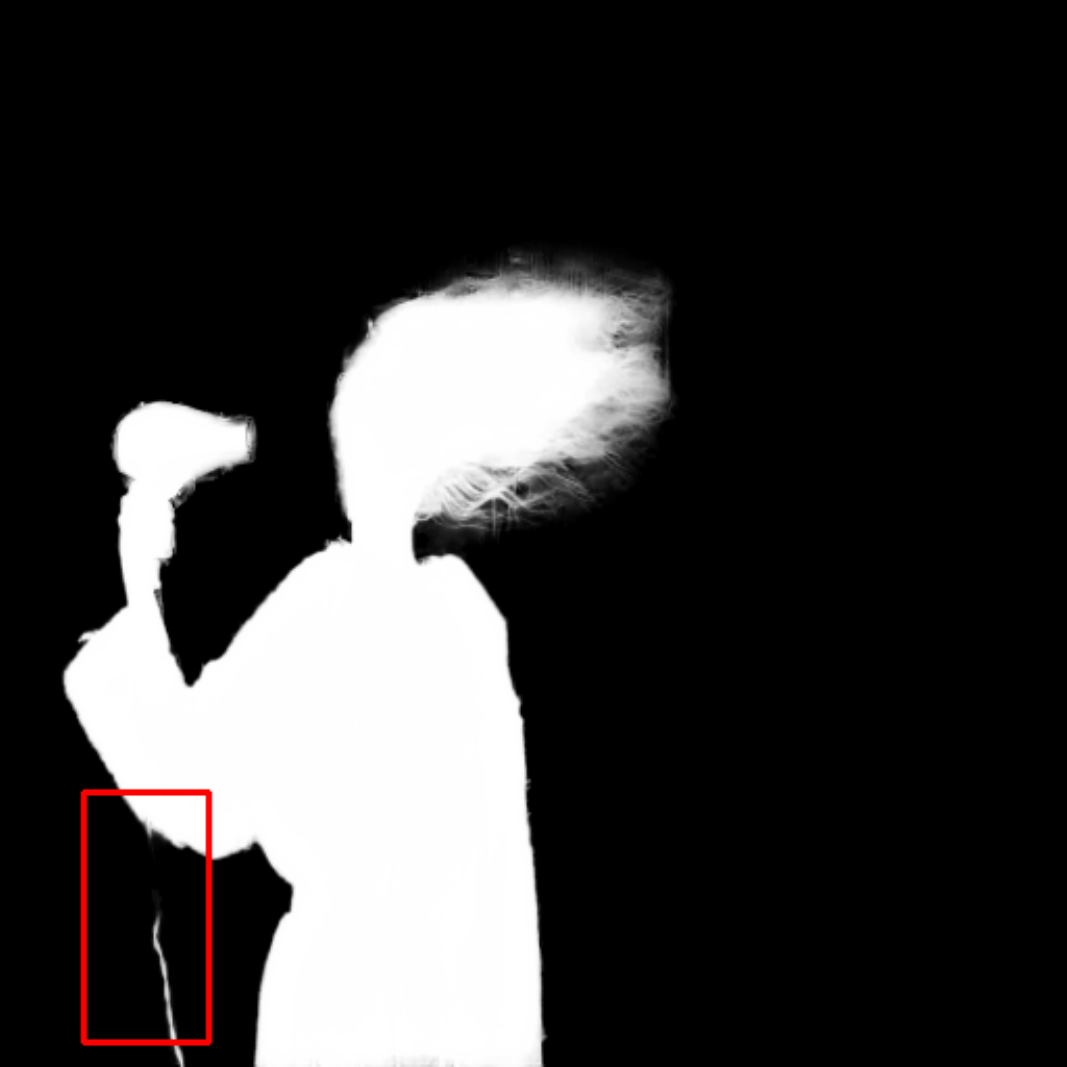}}\vspace{2pt}   
     \centerline{\includegraphics[scale=.1689]{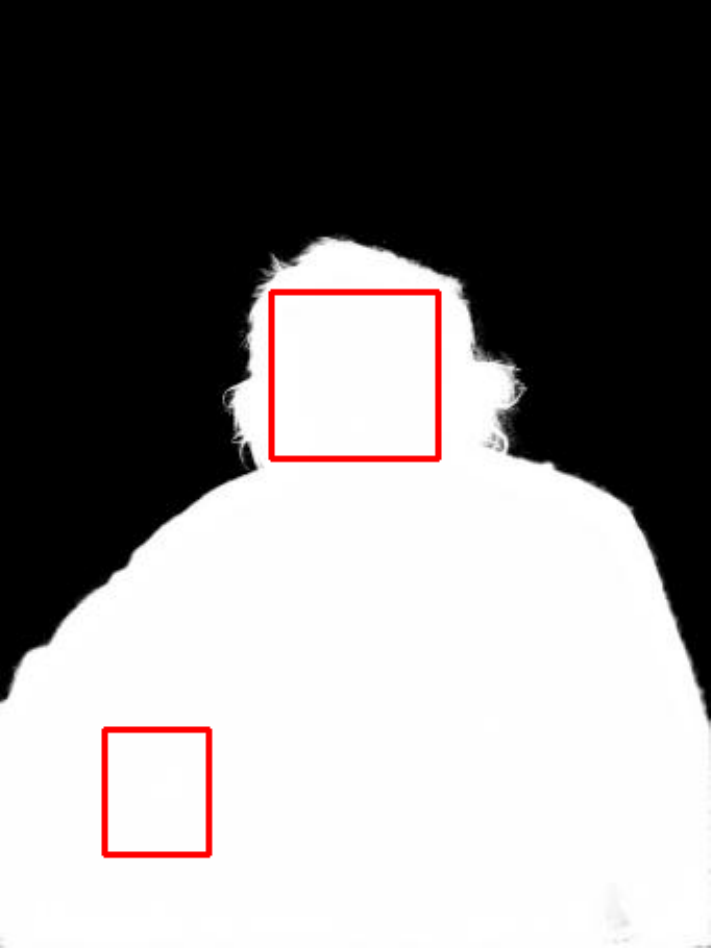}}\vspace{2pt}
     \centerline{\includegraphics[scale=.0473684]{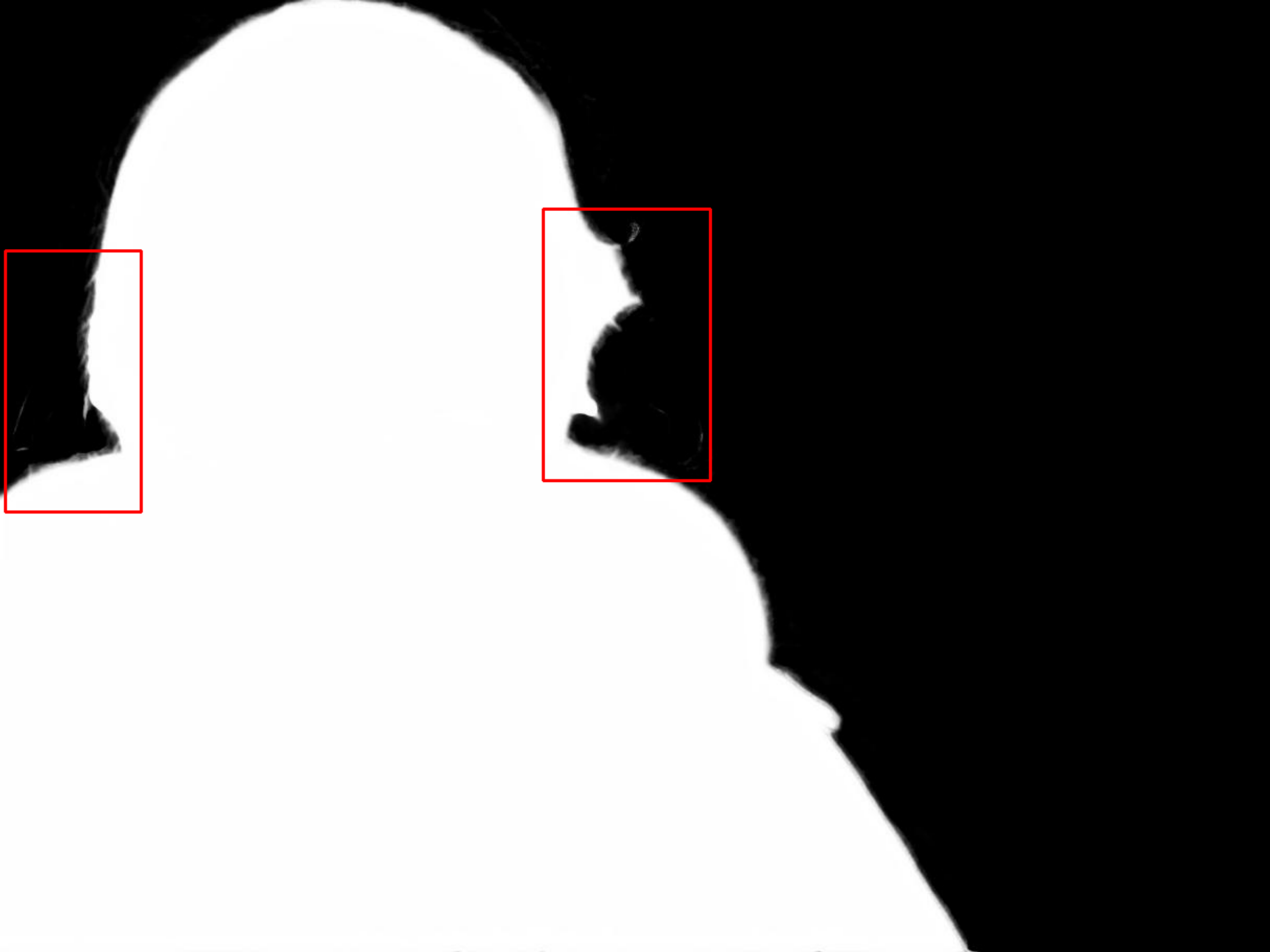}}\vspace{2pt}
     \centerline{\includegraphics[scale=.1056880734]{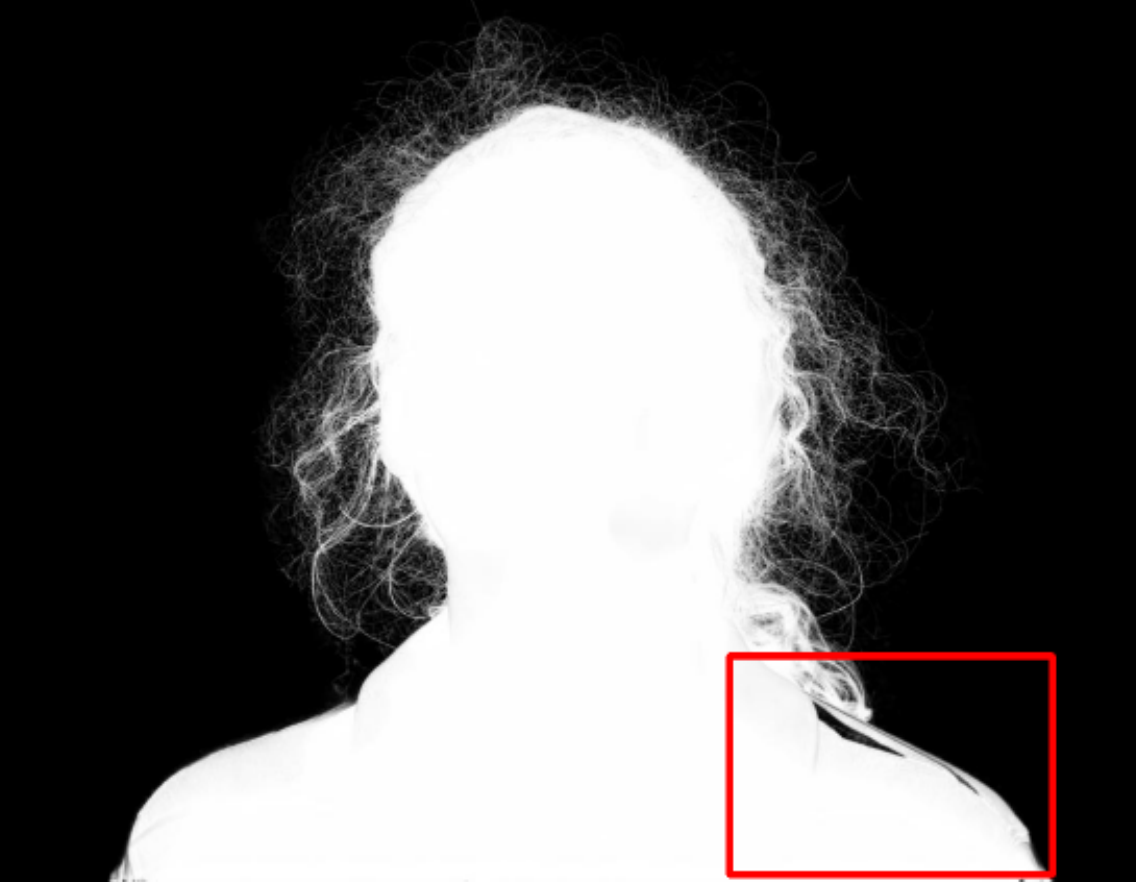}}  
    \centerline{HHM50K}
    \end{minipage}
    \hfill
    \begin{minipage}[t]{0.11\textwidth}
    \centerline{\includegraphics[scale=.1]{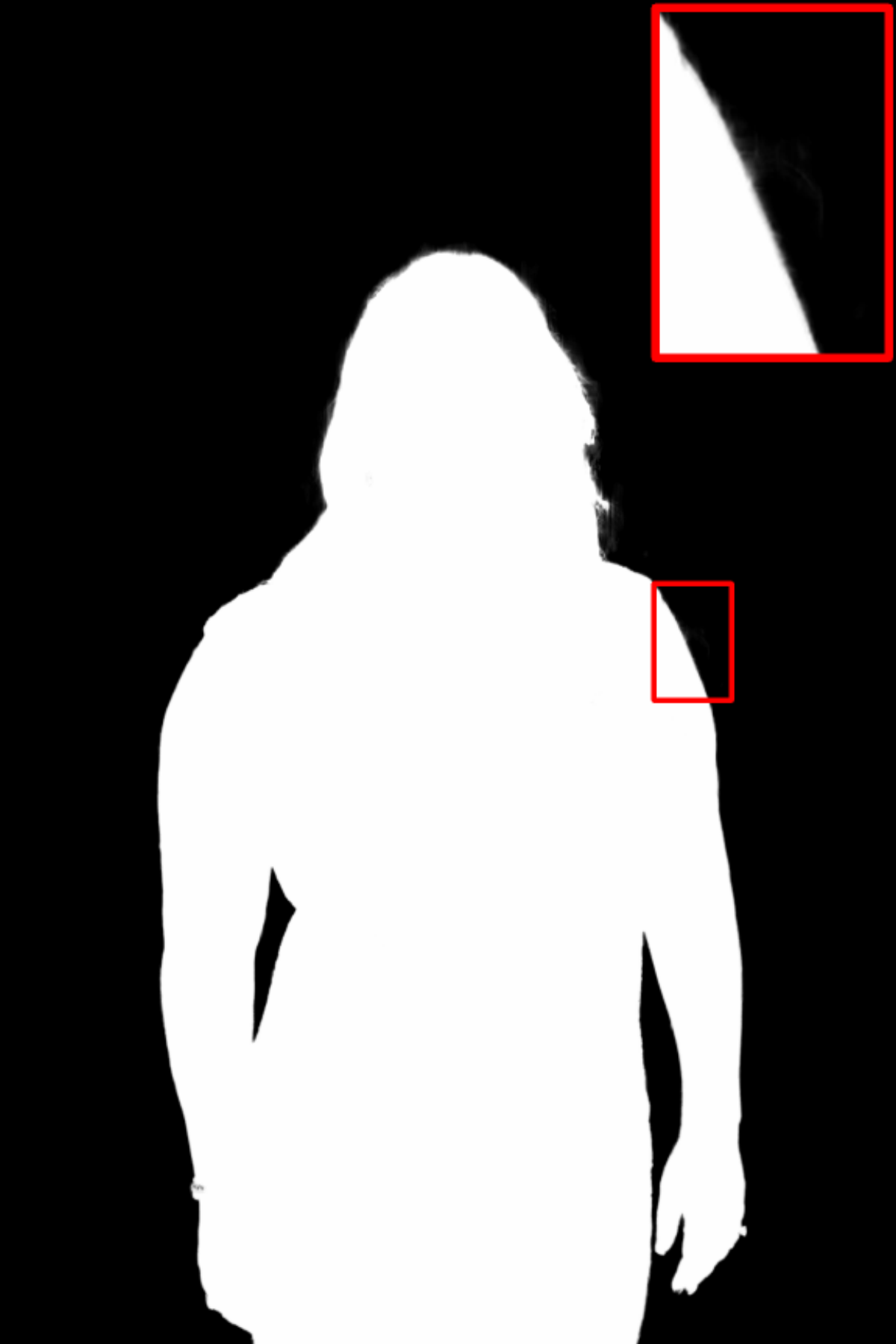}}\vspace{2pt}   
     \centerline{\includegraphics[scale=.1]{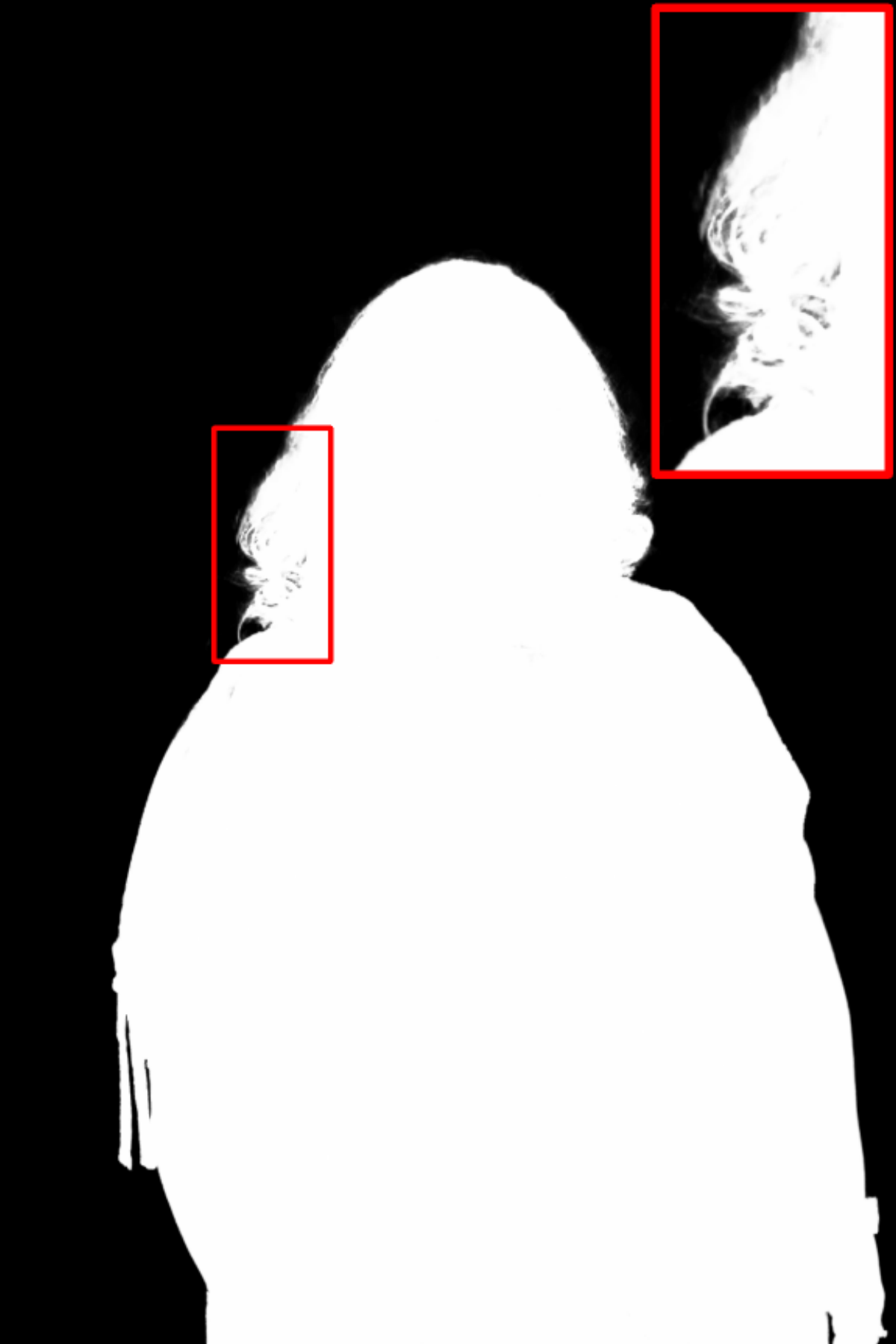}}\vspace{2pt} 
     \centerline{\includegraphics[scale=.1]{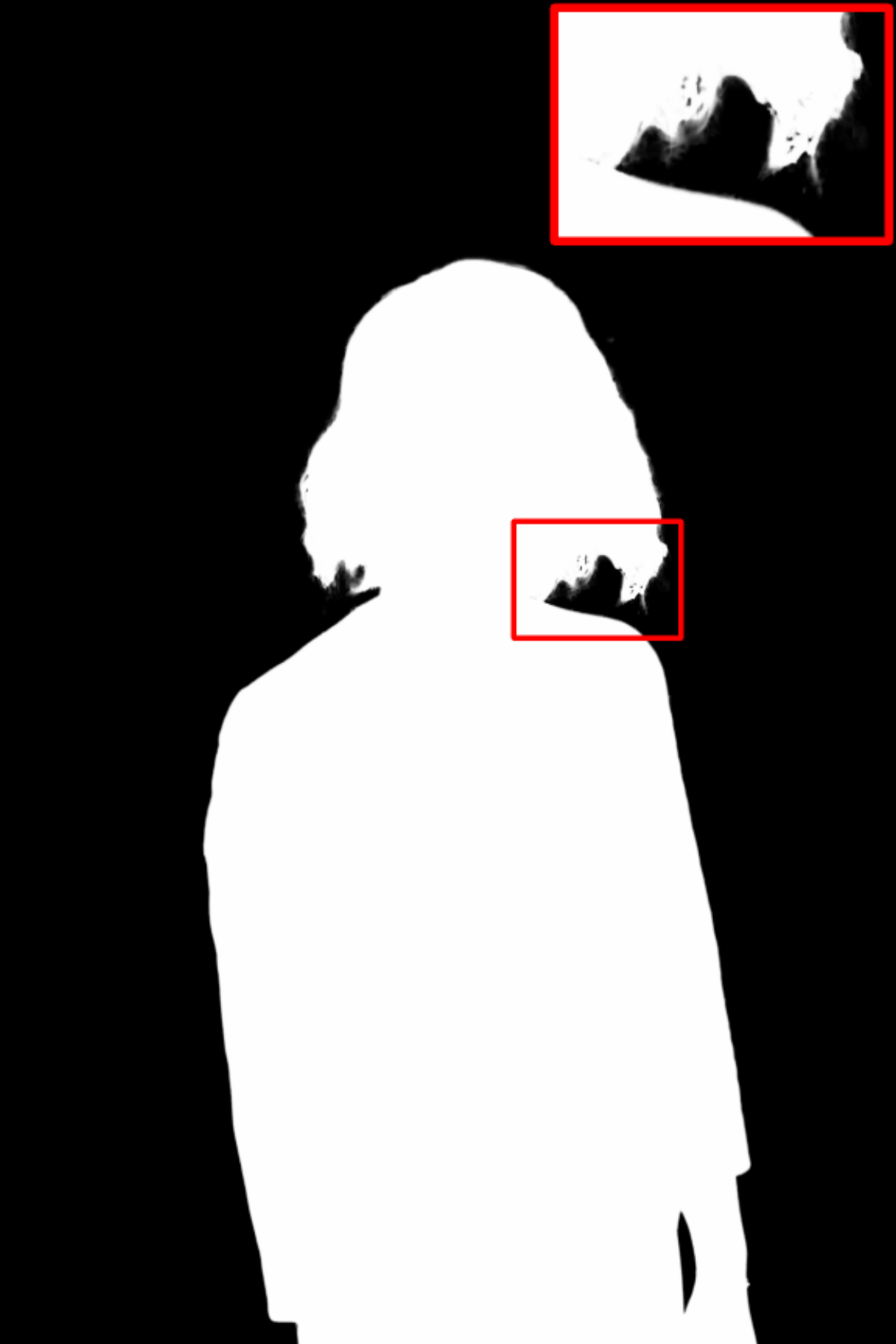}}\vspace{2pt}   
     % \centerline{\includegraphics[scale=.1]{figures/experiment/PH85-4/4_P3M_10k.pdf}}\vspace{2pt}  
     \centerline{\includegraphics[scale=.1125]{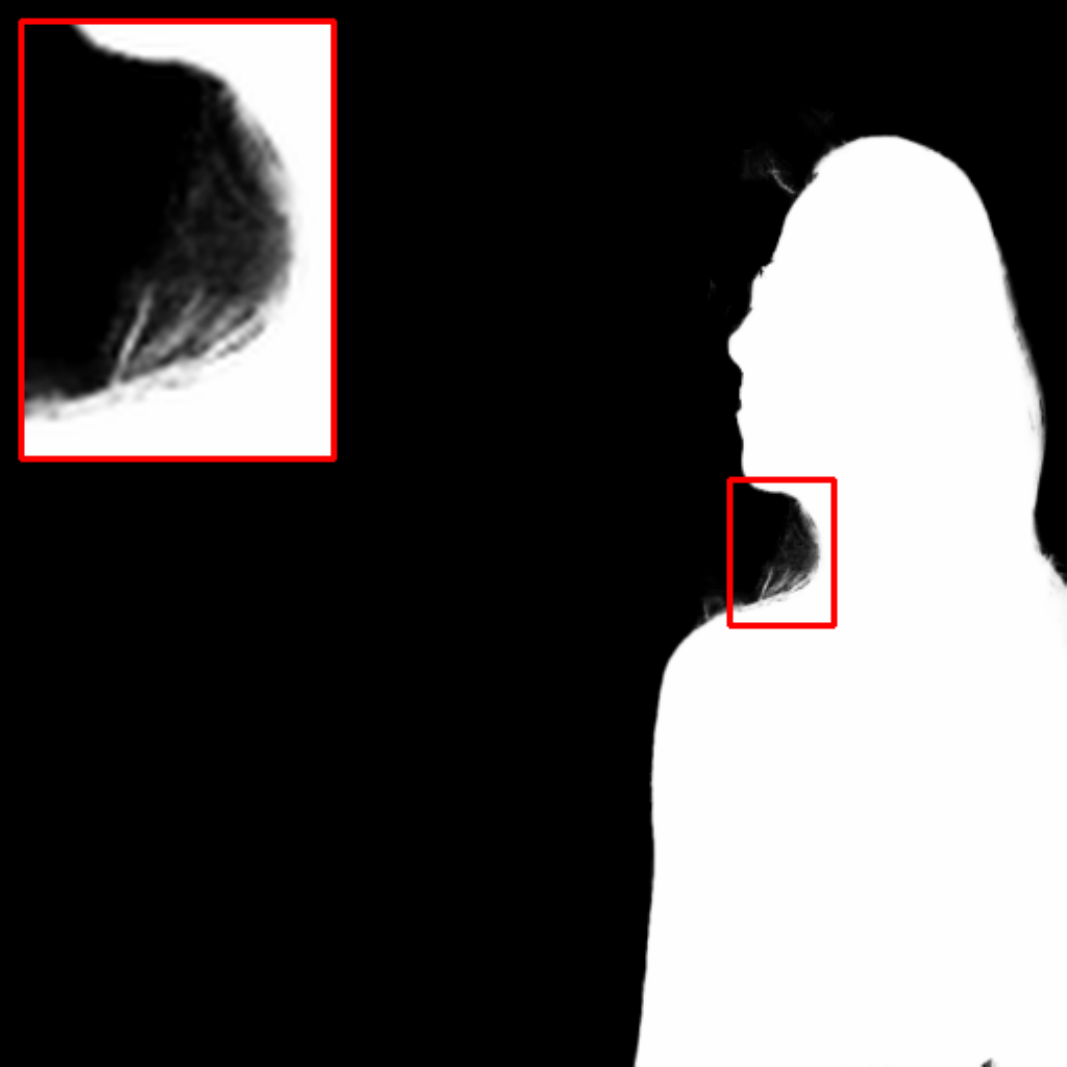}}\vspace{2pt}   
     \centerline{\includegraphics[scale=.1125]{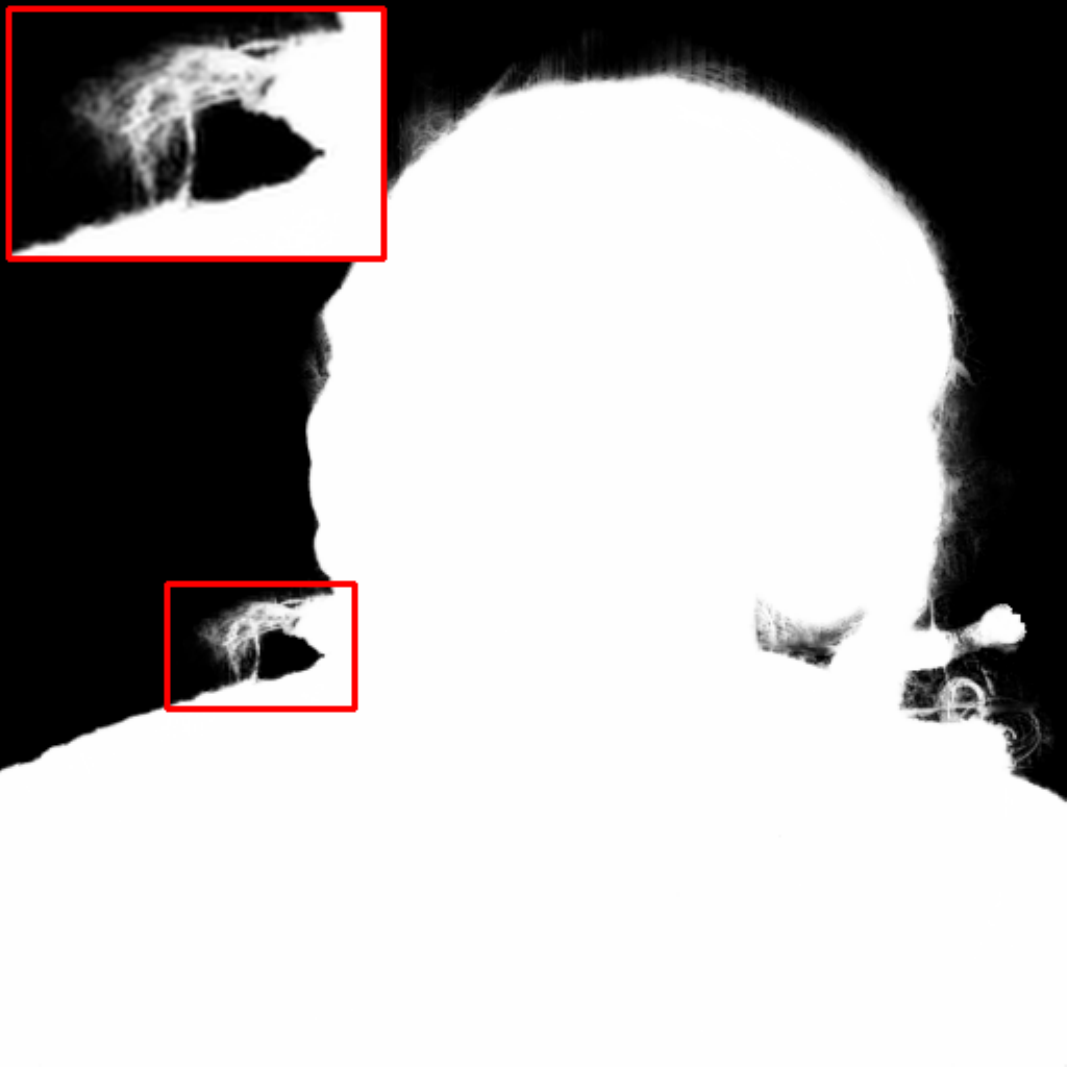}}\vspace{2pt} 
     \centerline{\includegraphics[scale=.1125]{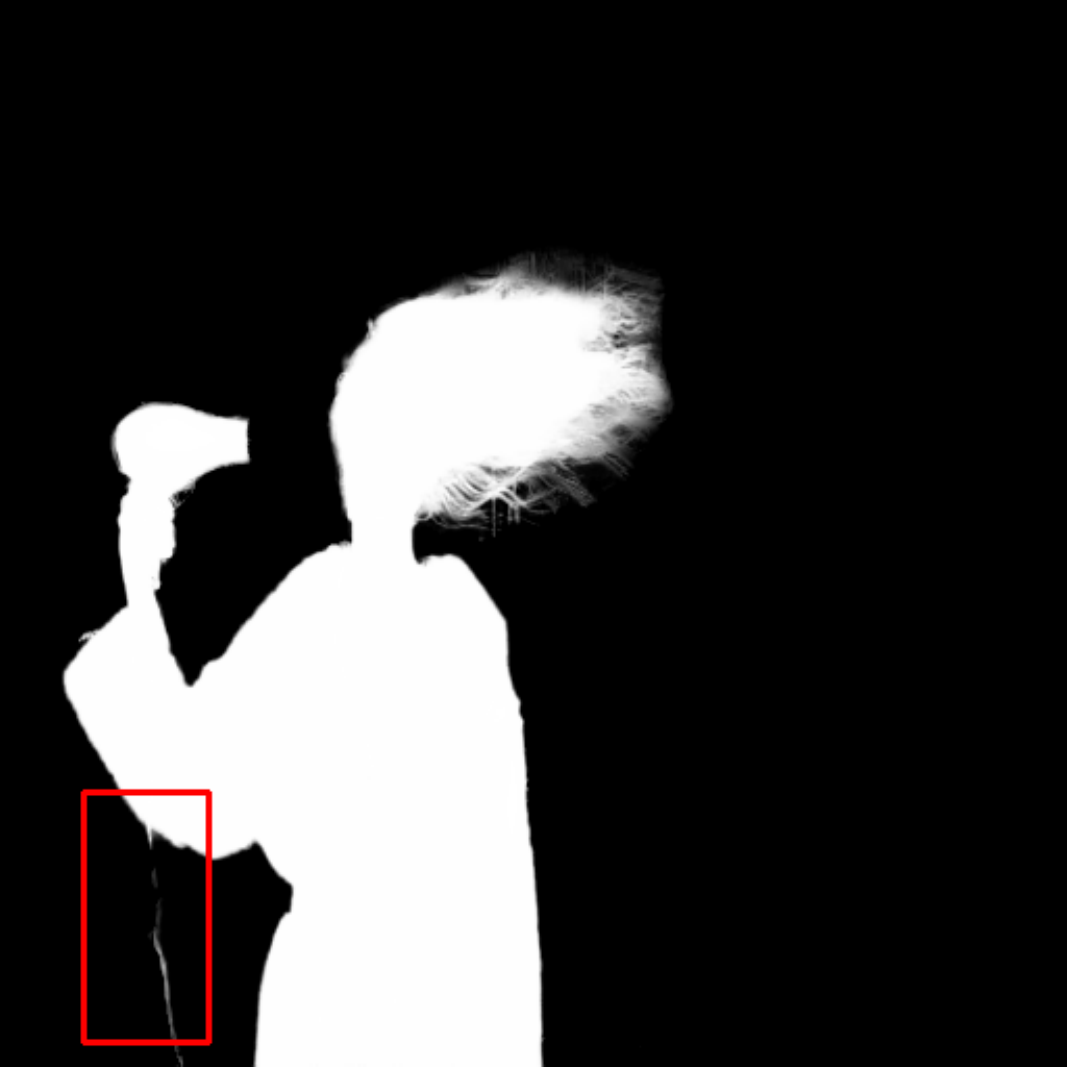}}\vspace{2pt}   
     \centerline{\includegraphics[scale=.1689]{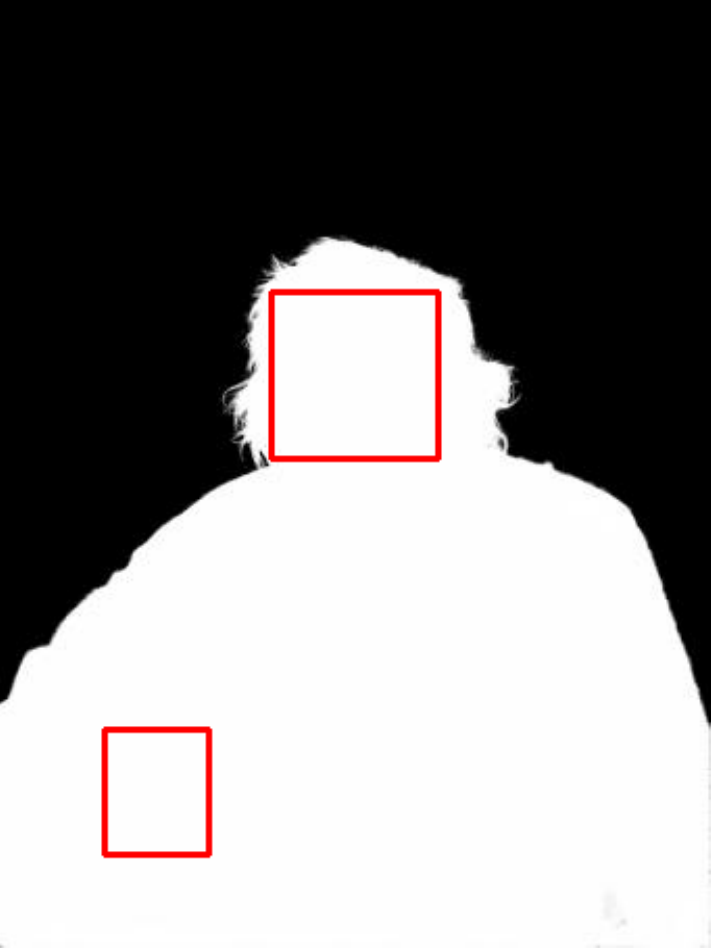}}\vspace{2pt}
     \centerline{\includegraphics[scale=.0473684]{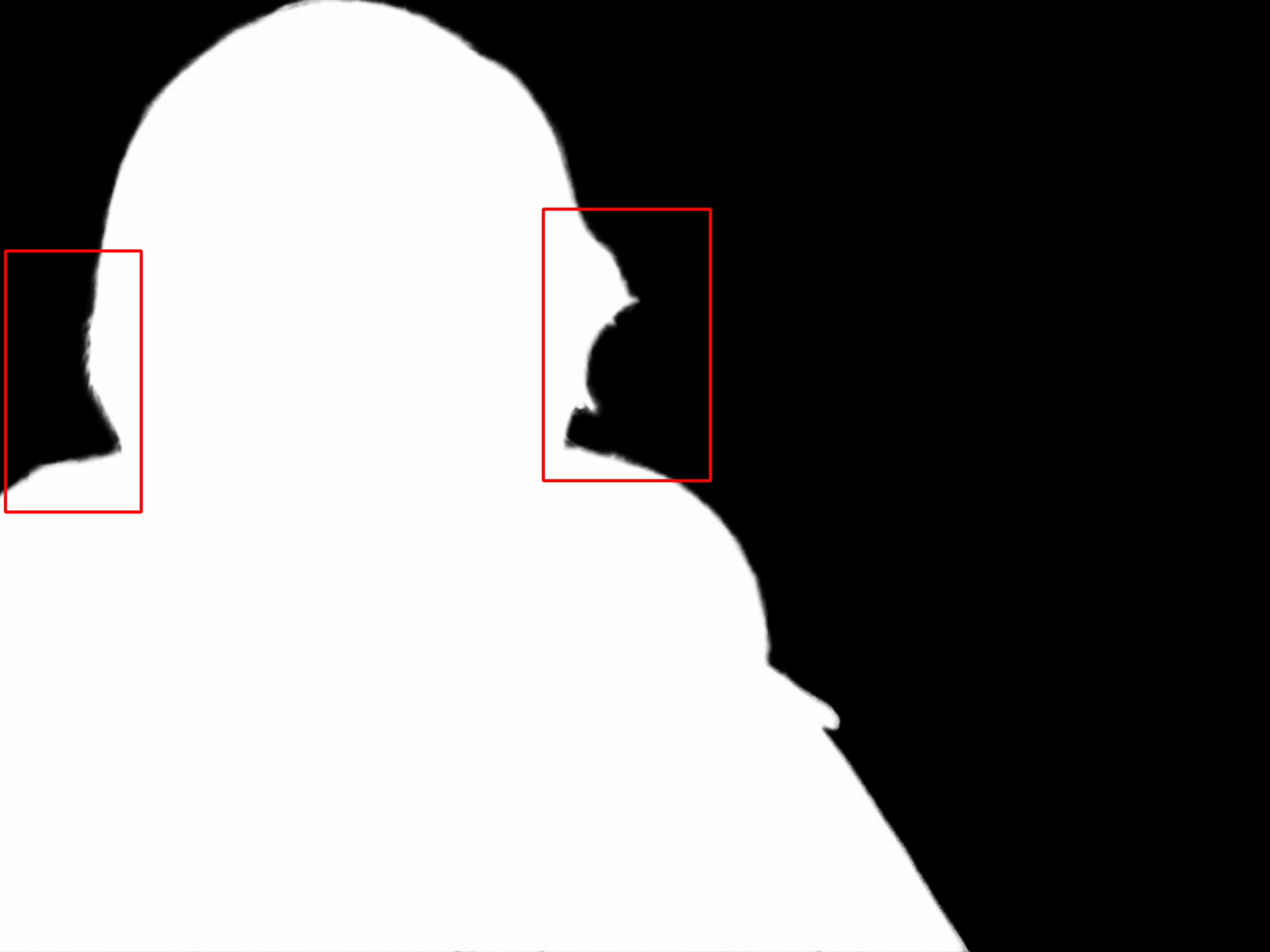}}\vspace{2pt}
     \centerline{\includegraphics[scale=.1056880734]{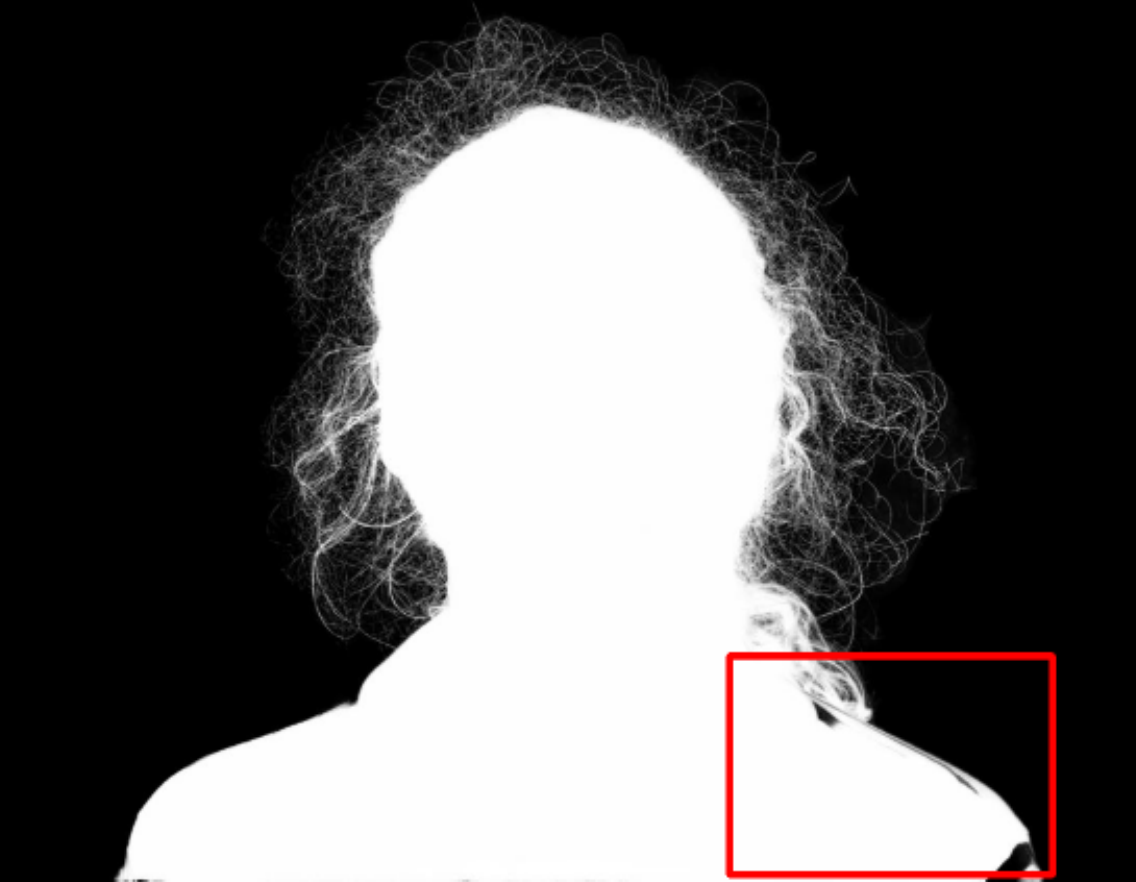}}  
    \centerline{P3M-10k}
    \end{minipage}
    \hfill
    \begin{minipage}[t]{0.11\textwidth}
    \centerline{\includegraphics[scale=.1]{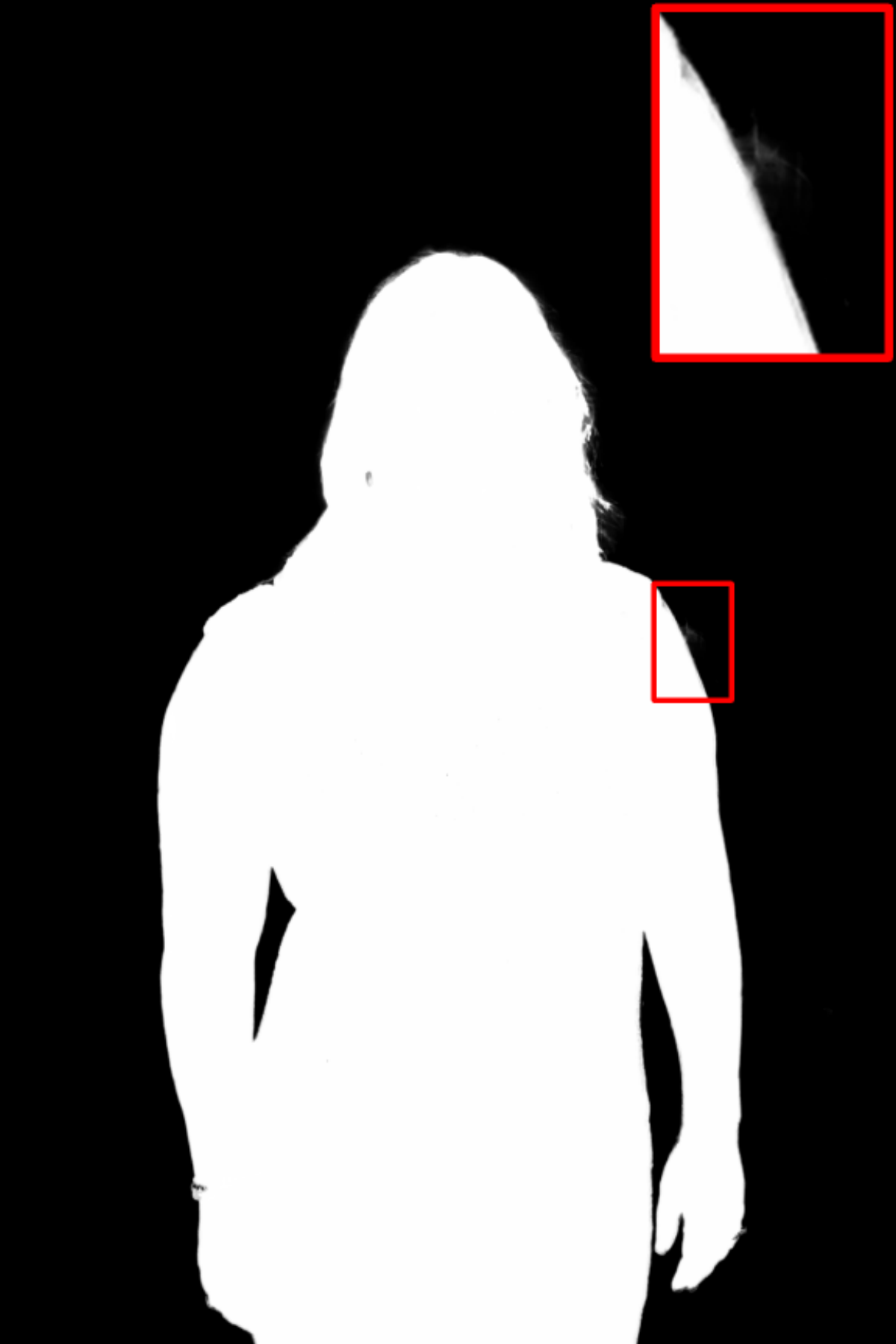}}\vspace{2pt}   
     \centerline{\includegraphics[scale=.1]{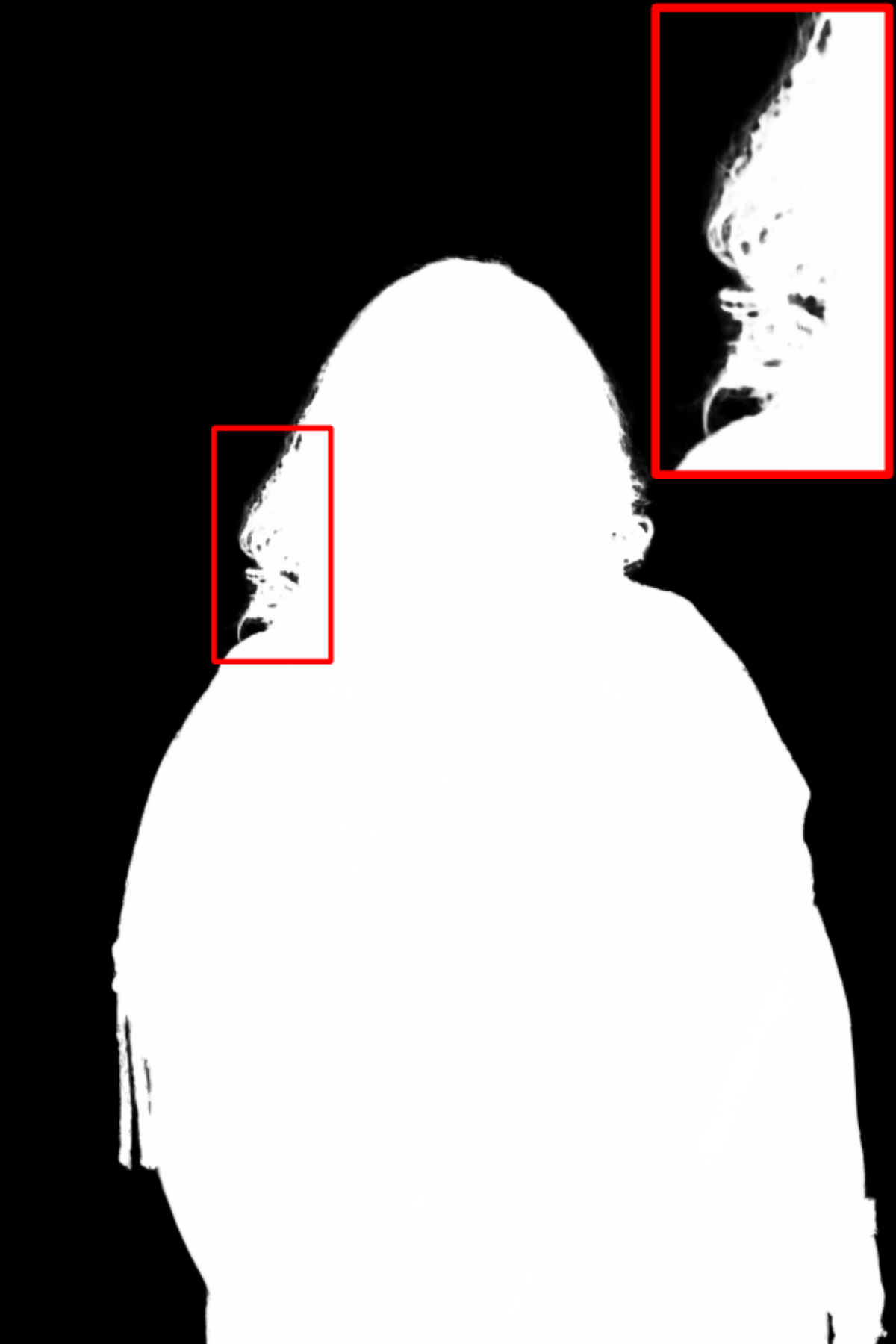}}\vspace{2pt} 
     \centerline{\includegraphics[scale=.1]{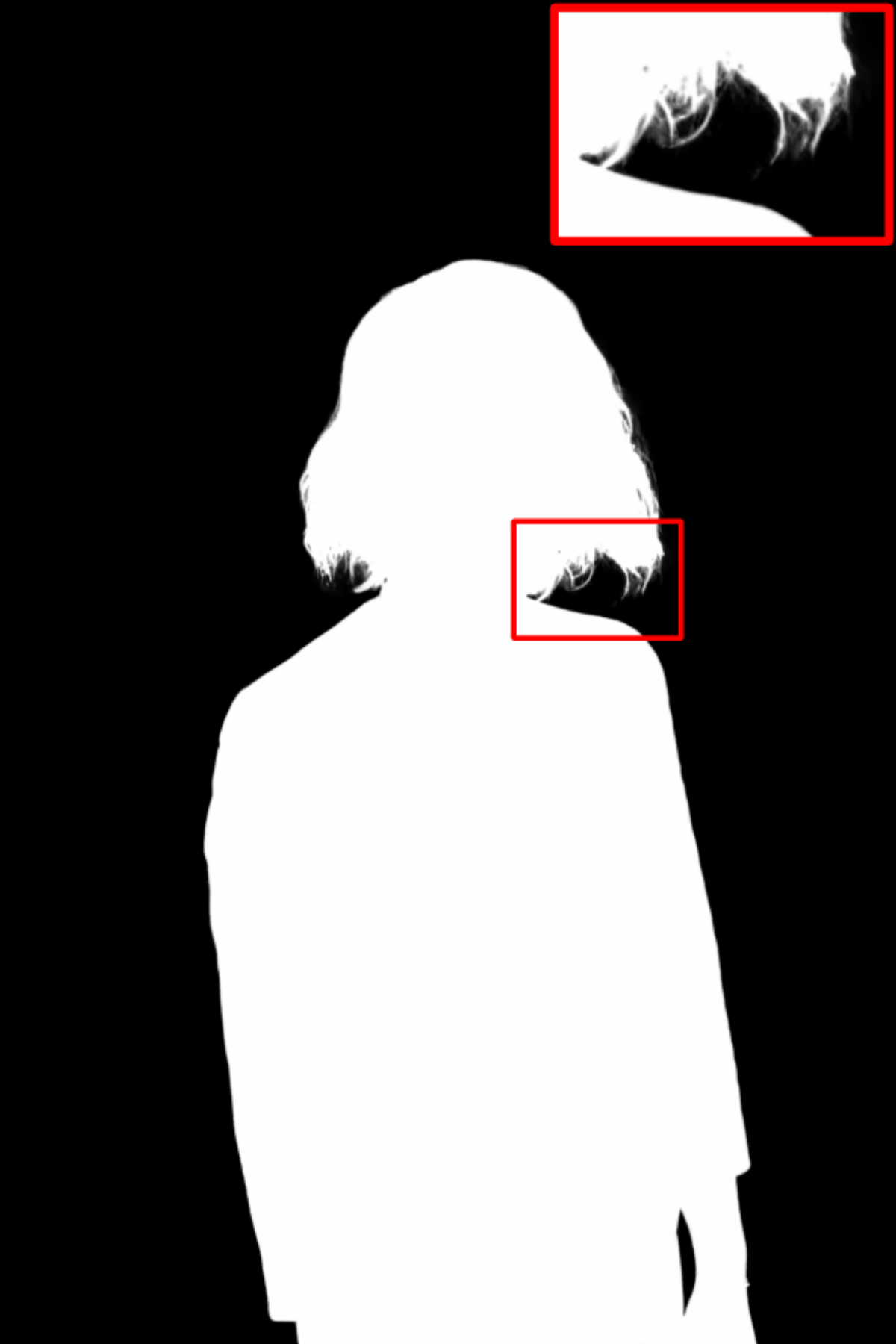}}\vspace{2pt}   
     % \centerline{\includegraphics[scale=.1]{figures/experiment/PH85-4/4_ImageMatte.pdf}}\vspace{2pt}  
     \centerline{\includegraphics[scale=.1125]{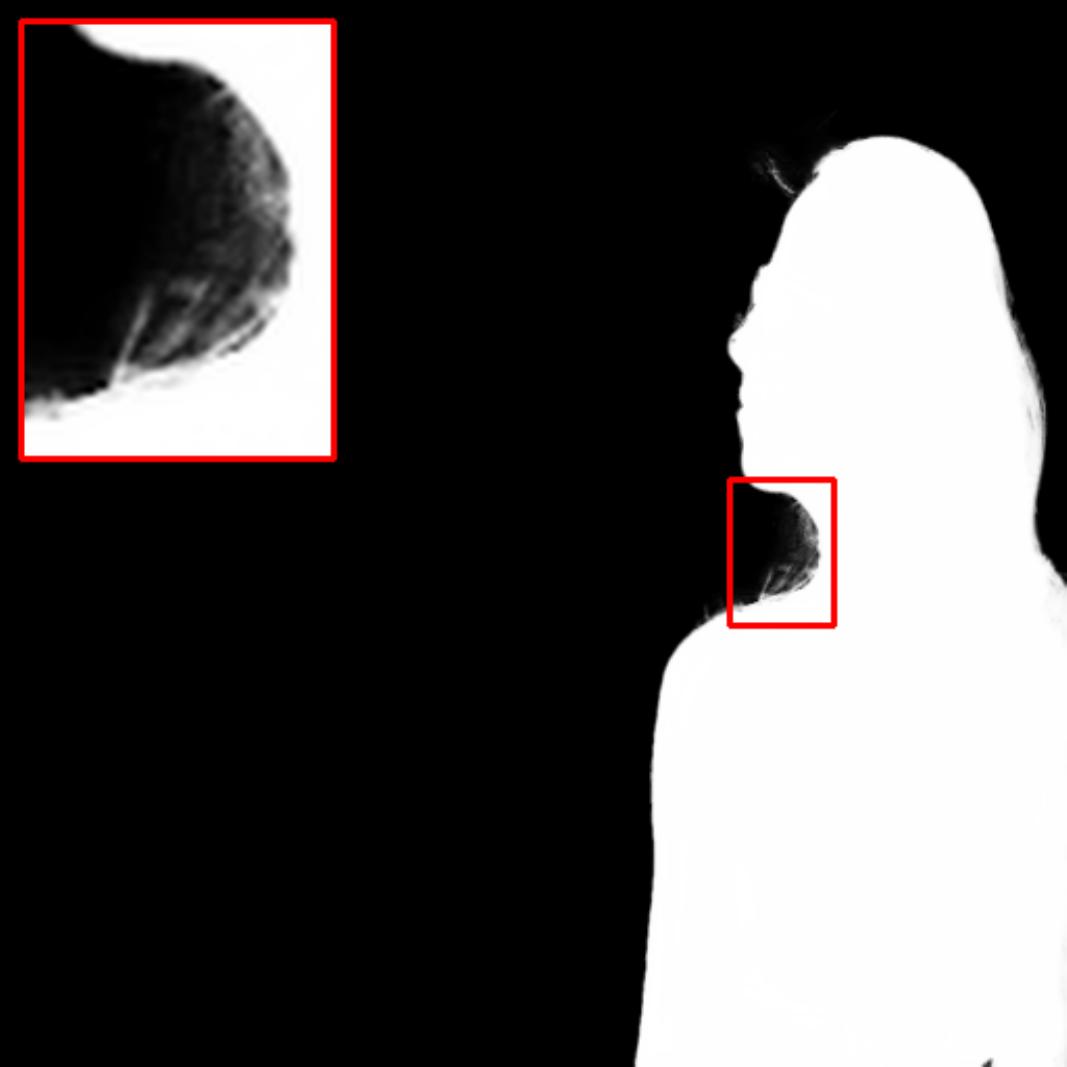}}\vspace{2pt}   
     \centerline{\includegraphics[scale=.1125]{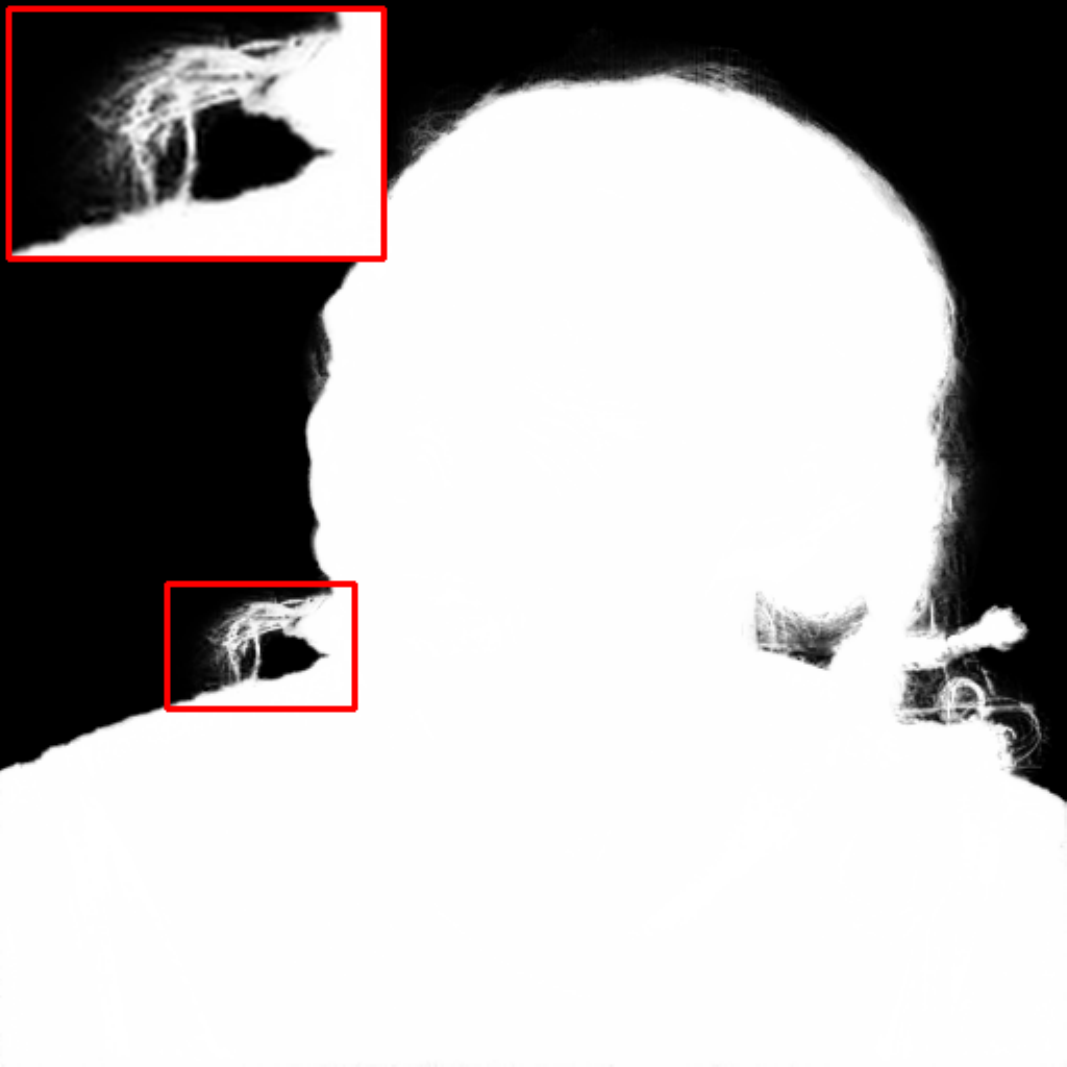}}\vspace{2pt} 
     \centerline{\includegraphics[scale=.1125]{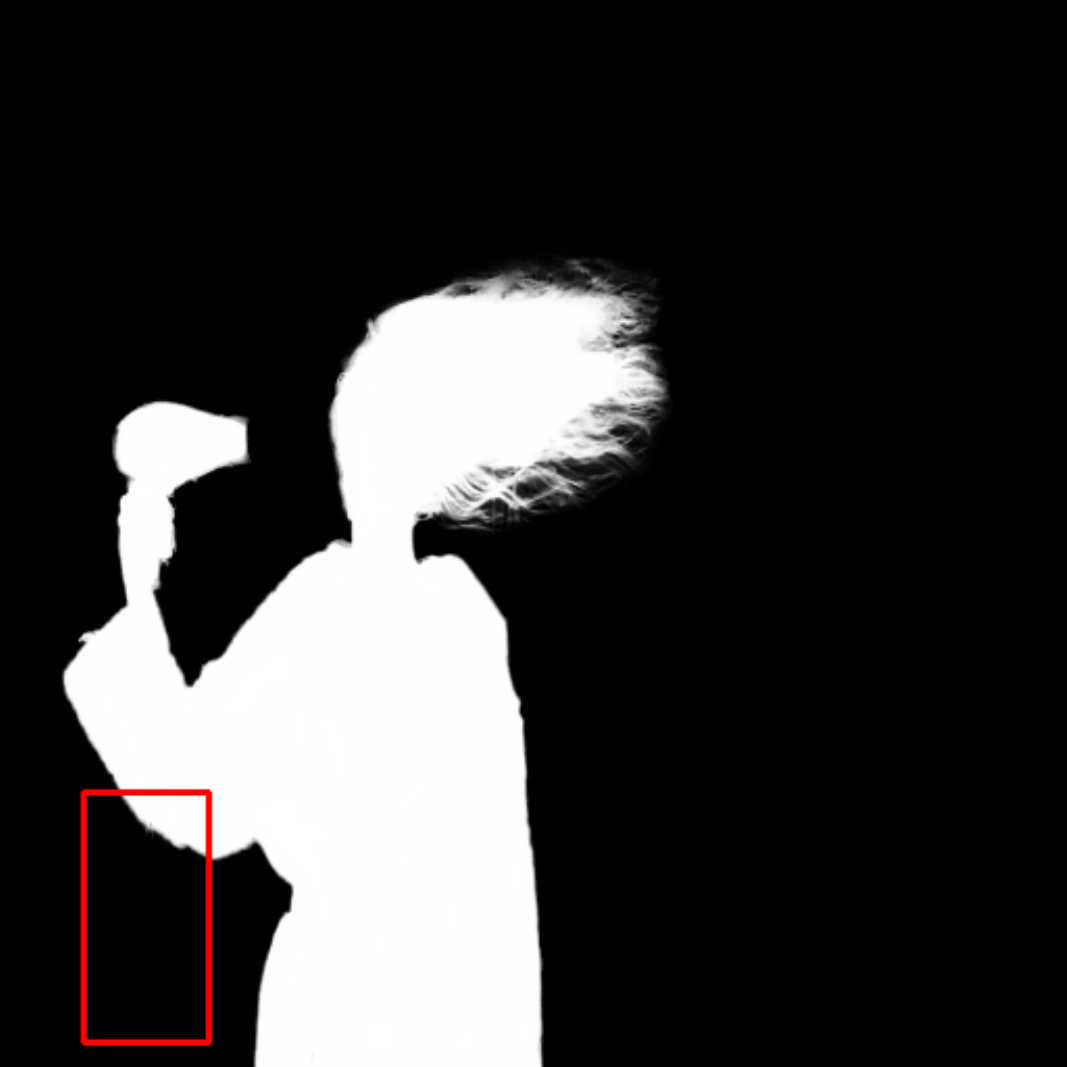}}\vspace{2pt}   
     \centerline{\includegraphics[scale=.1689]{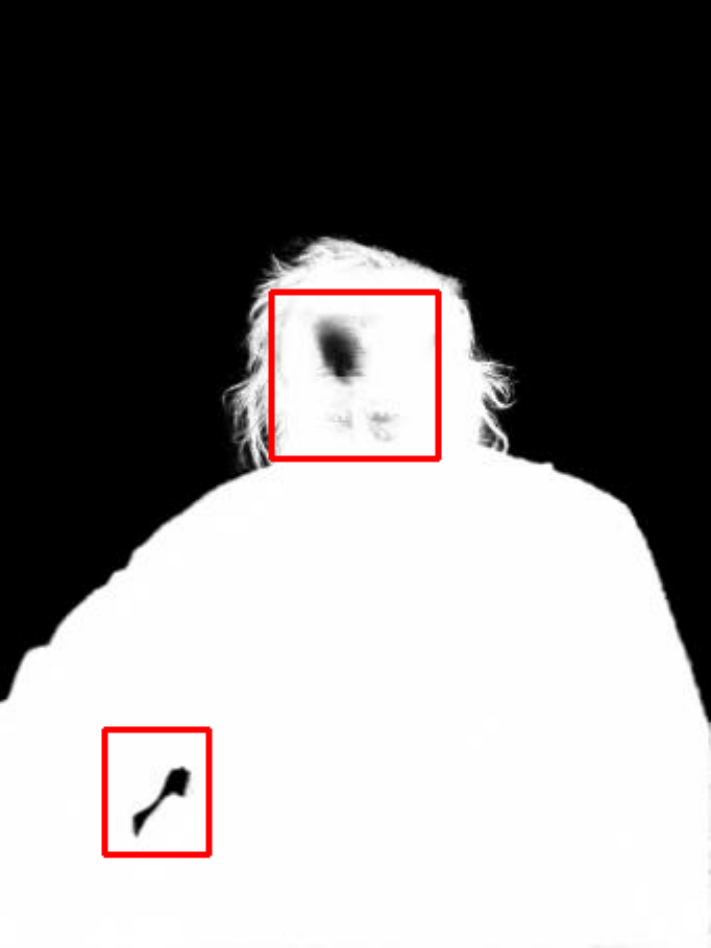}}\vspace{2pt}
     \centerline{\includegraphics[scale=.0473684]{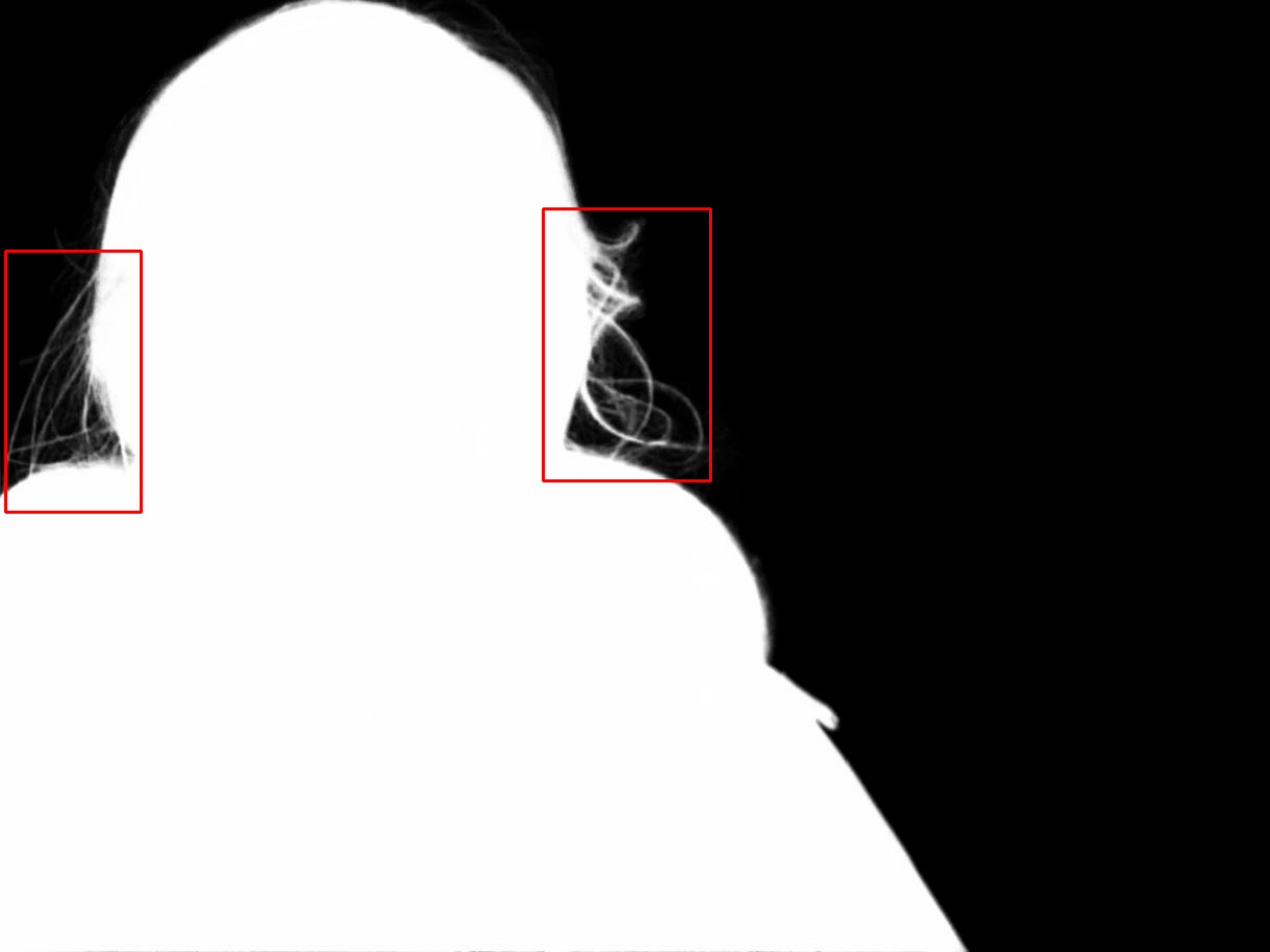}}\vspace{2pt}
     \centerline{\includegraphics[scale=.1056880734]{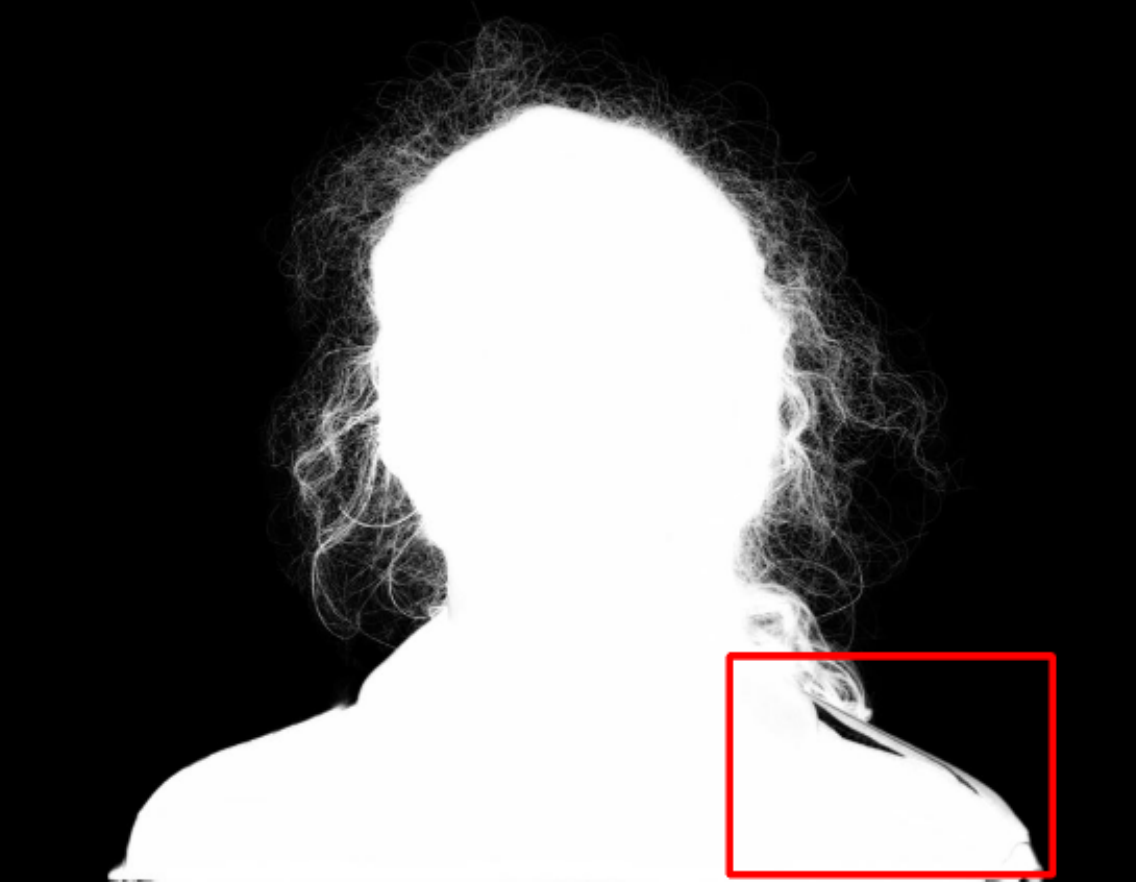}}  
    \centerline{ImageMatte}
    \end{minipage}
    \hfill
    \begin{minipage}[t]{0.11\textwidth}
    \centerline{\includegraphics[scale=.1]{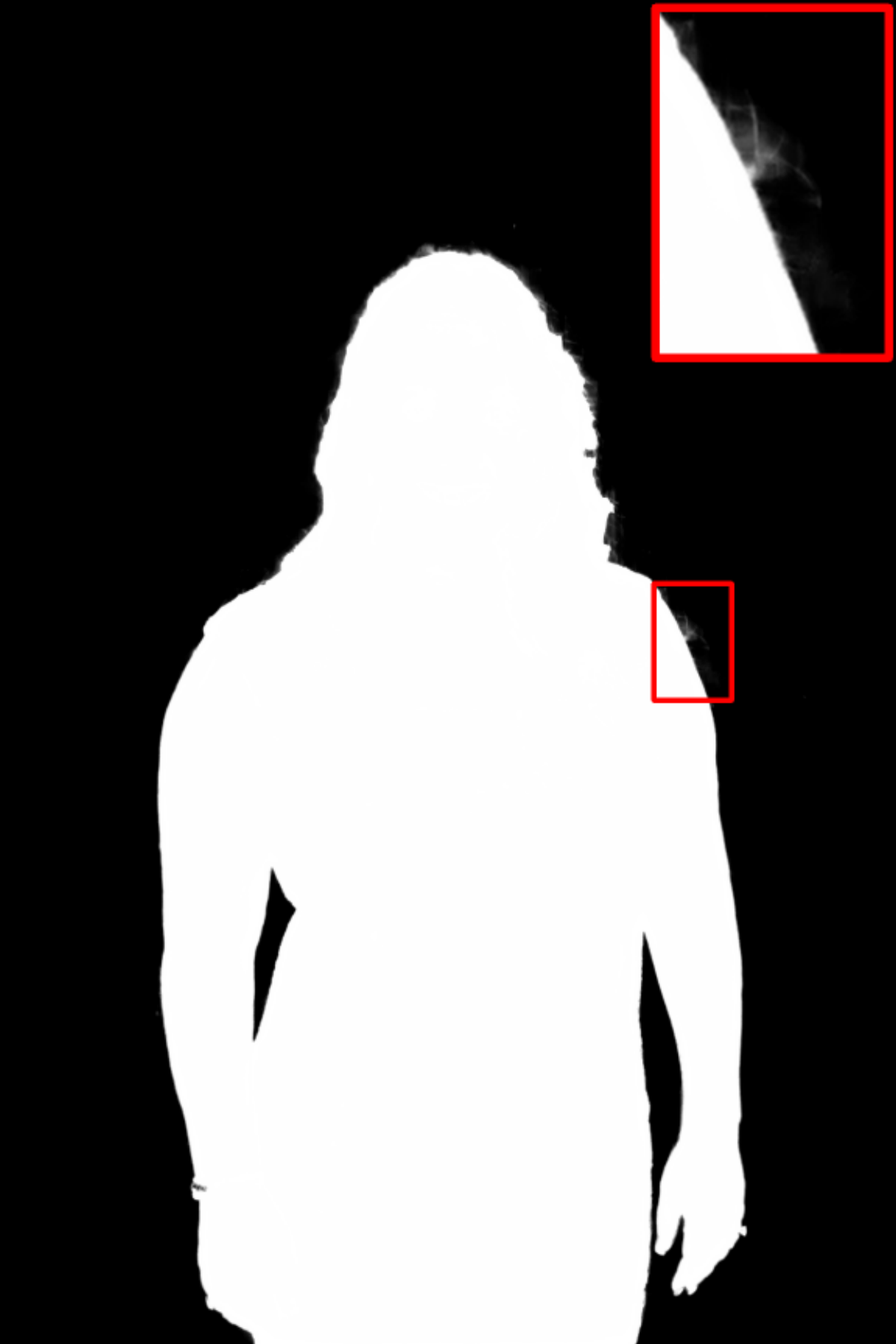}}\vspace{2pt}   
     \centerline{\includegraphics[scale=.1]{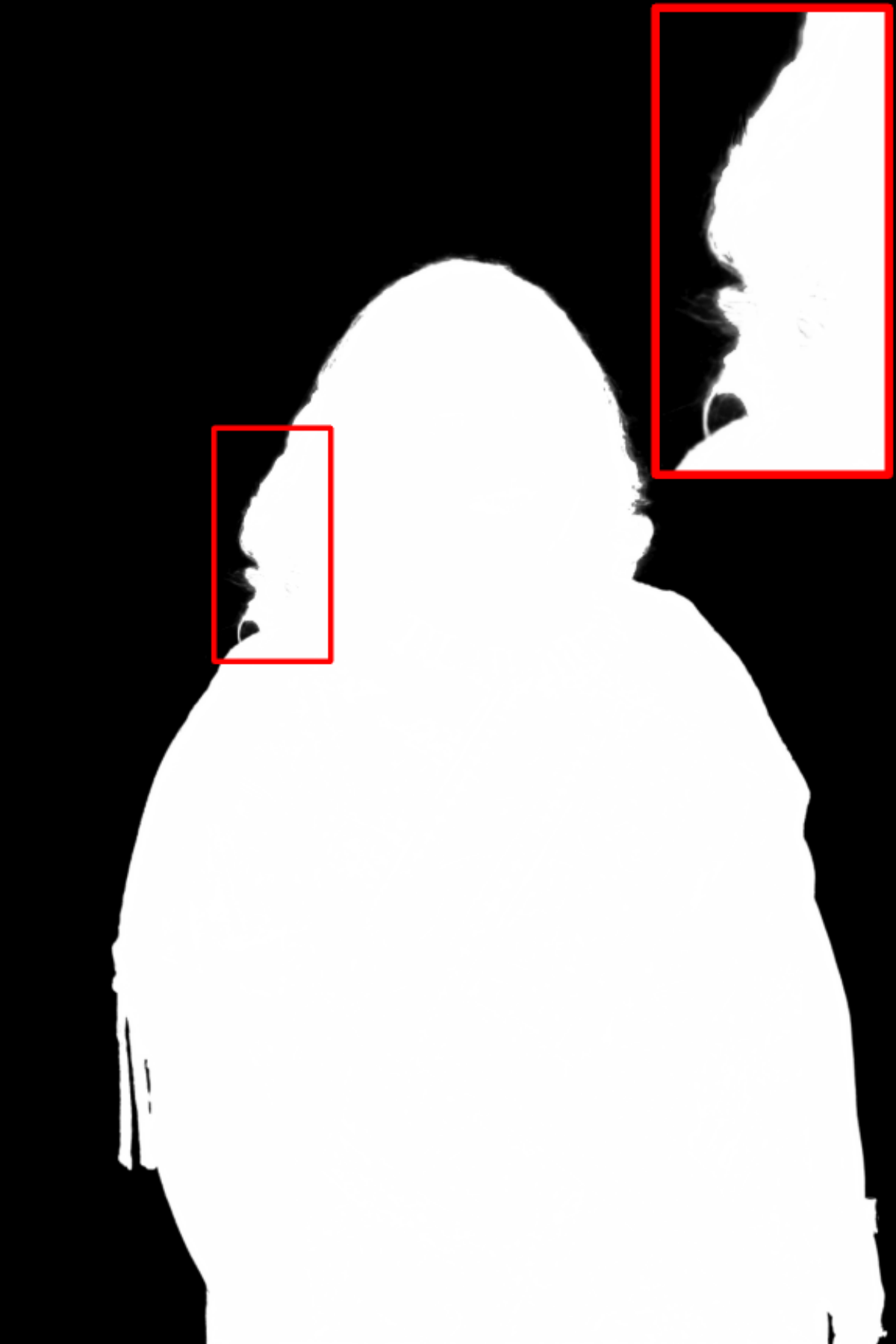}}\vspace{2pt} 
     \centerline{\includegraphics[scale=.1]{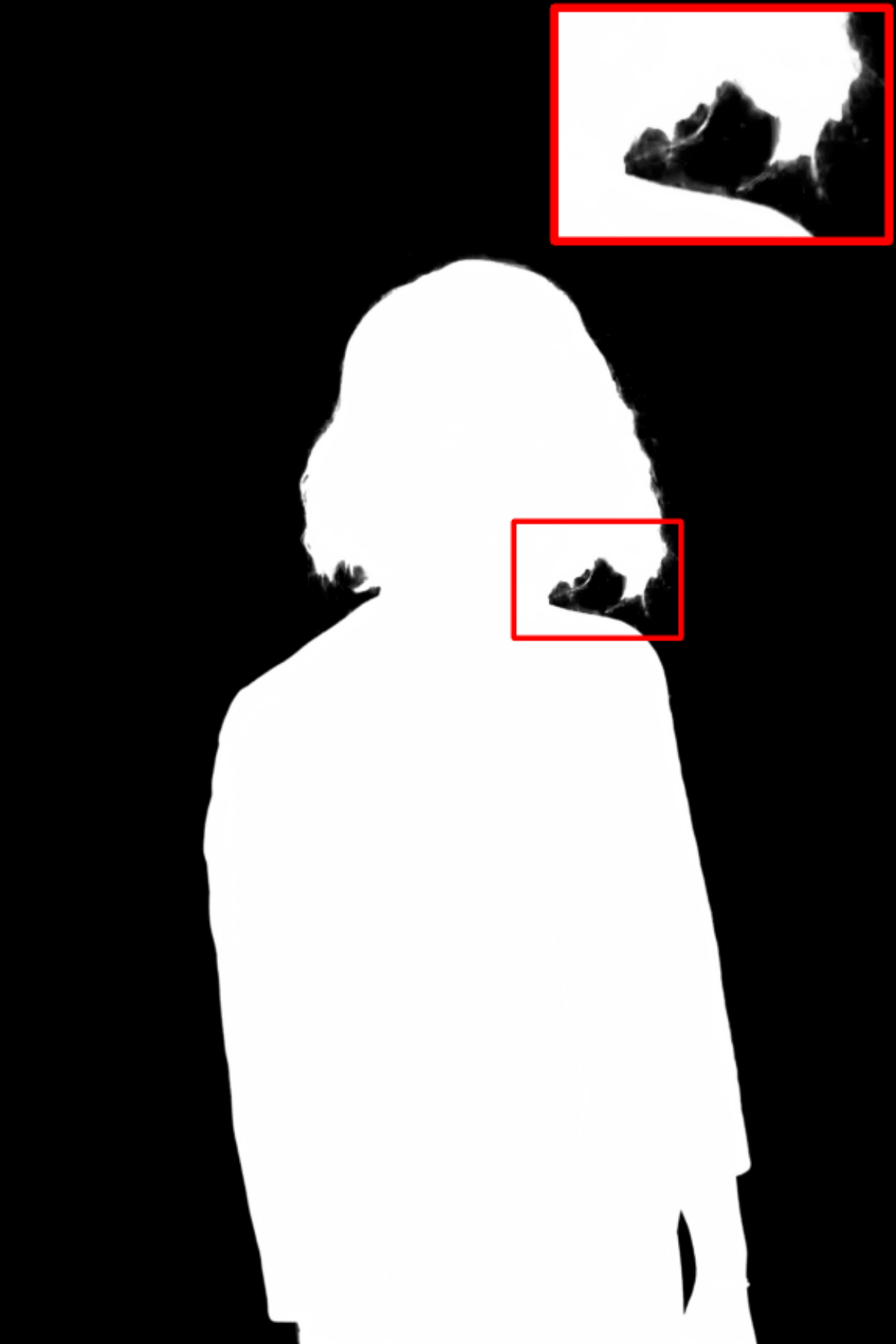}}\vspace{2pt}   
     % \centerline{\includegraphics[scale=.1]{figures/experiment/PH85-4/4_LD_M_CK.pdf}}\vspace{2pt}  
     \centerline{\includegraphics[scale=.1125]{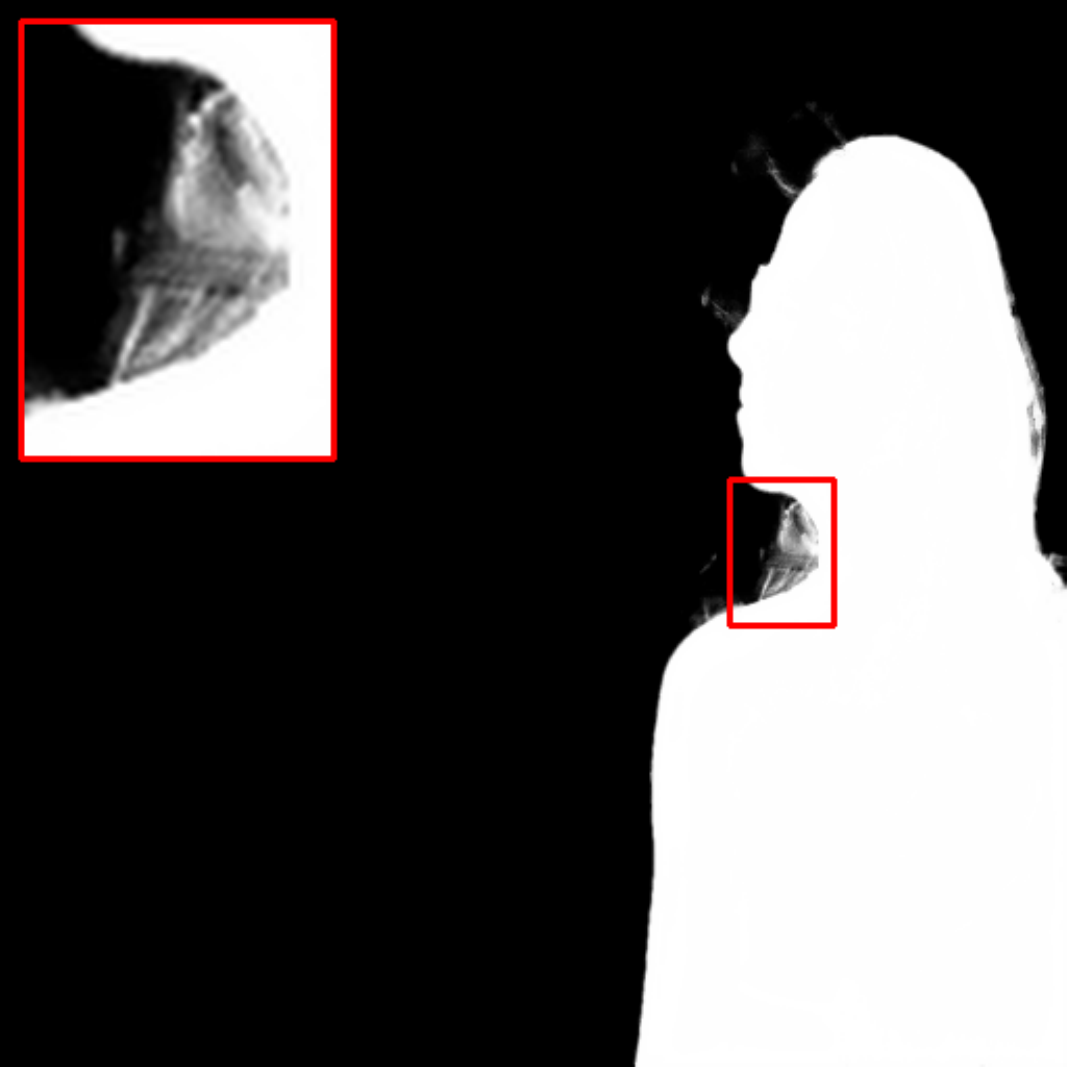}}\vspace{2pt}   
     \centerline{\includegraphics[scale=.1125]{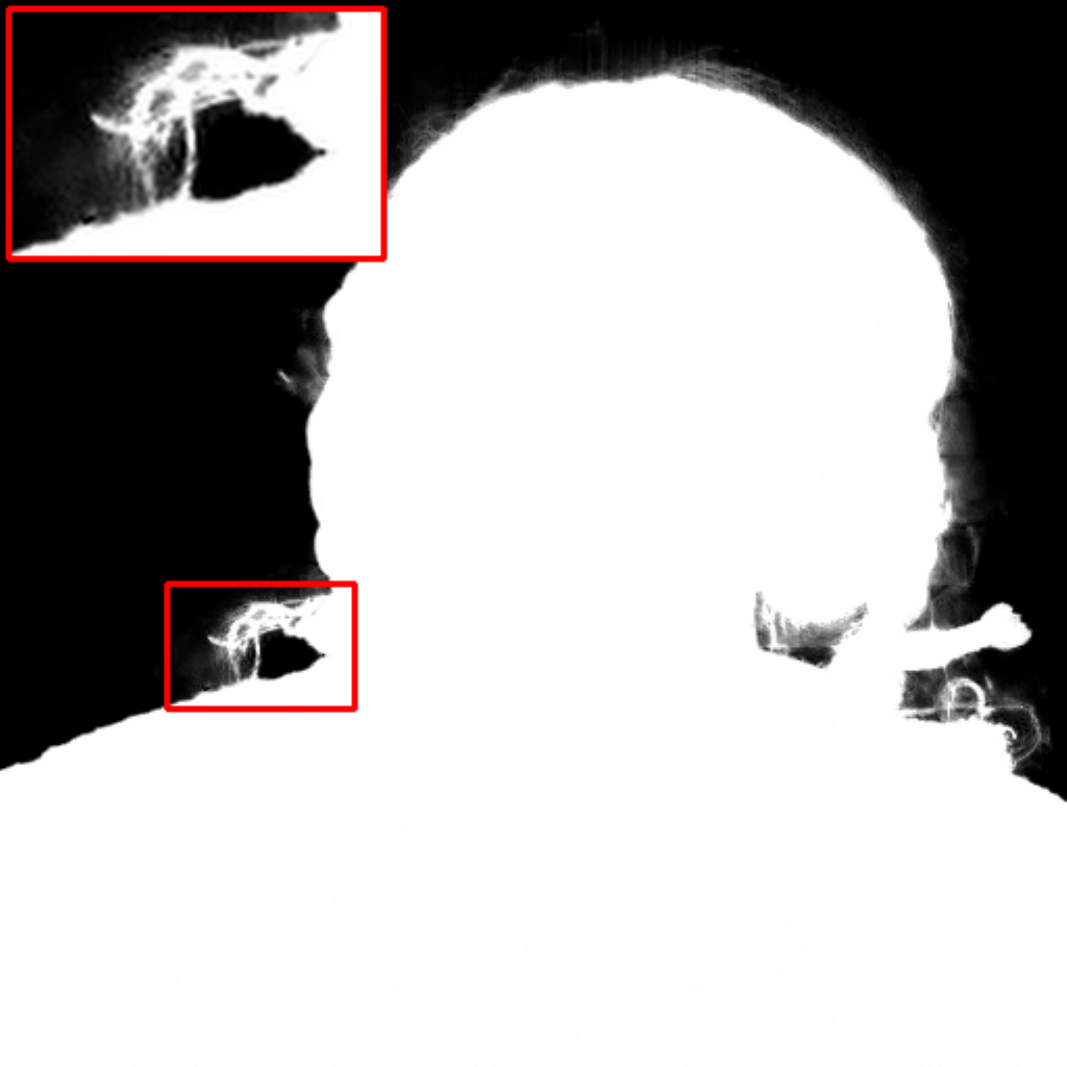}}\vspace{2pt} 
     \centerline{\includegraphics[scale=.1125]{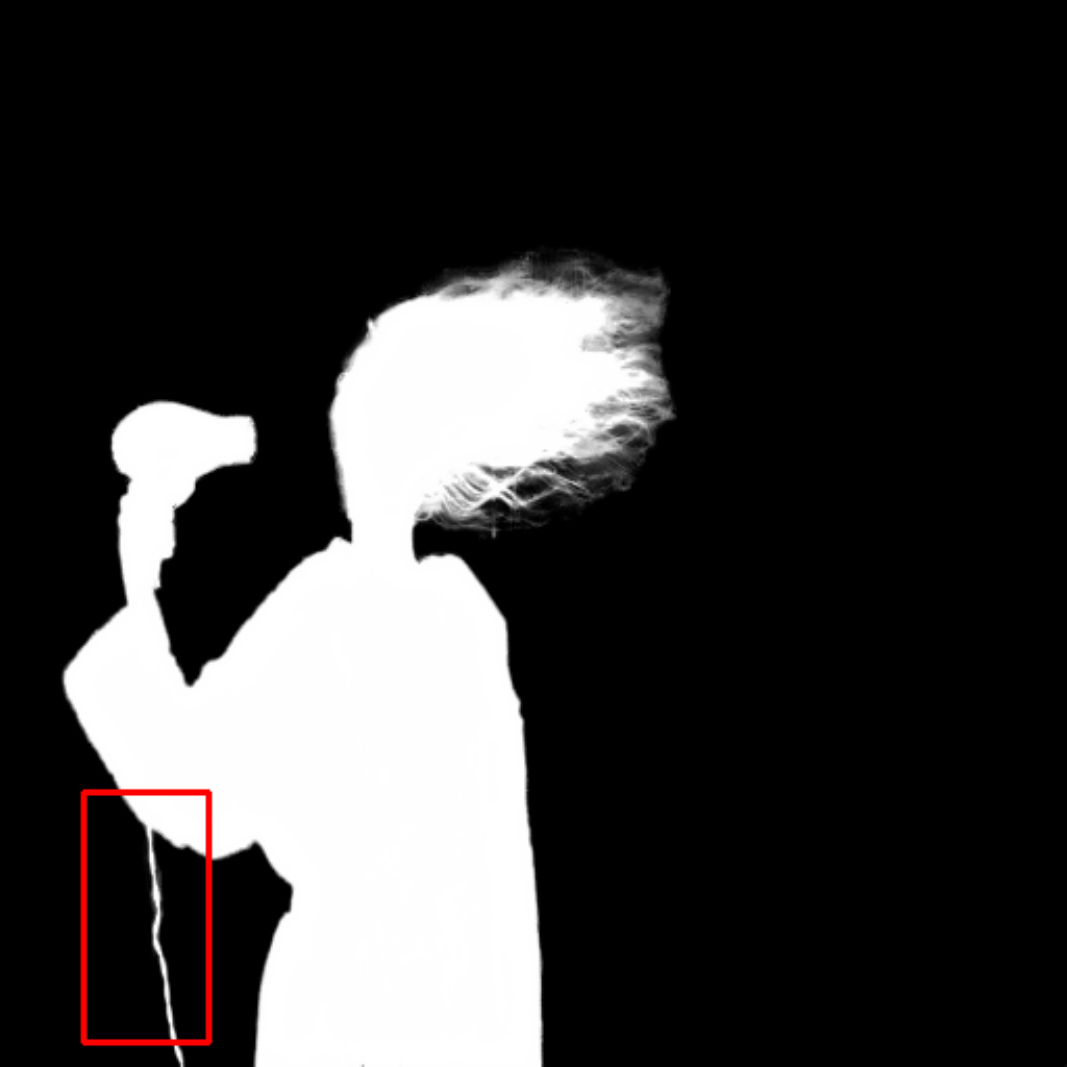}}\vspace{2pt}   
     \centerline{\includegraphics[scale=.1689]{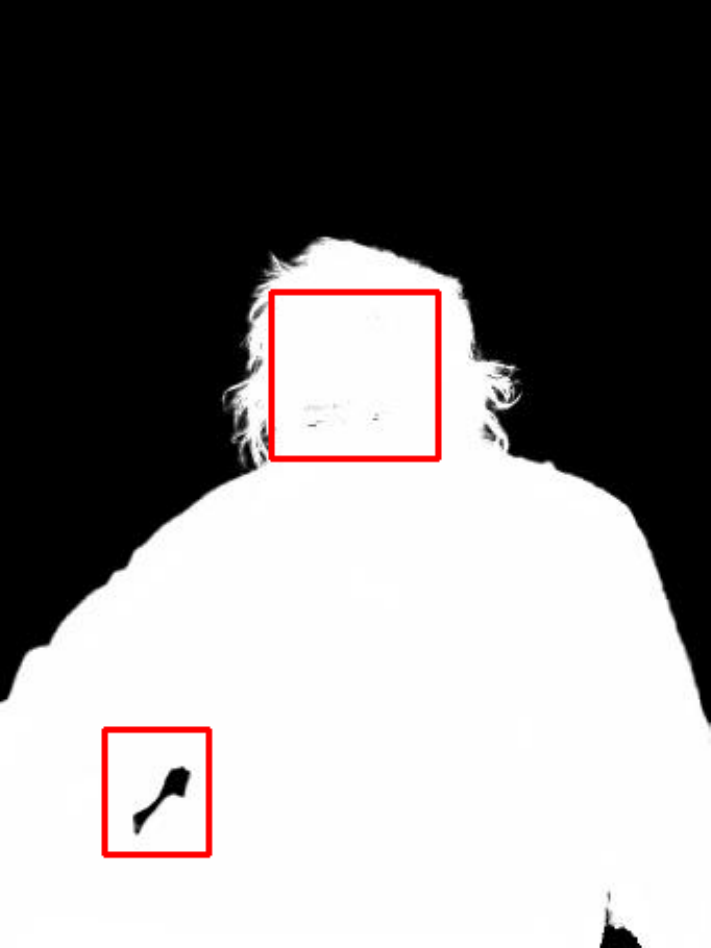}}\vspace{2pt}
     \centerline{\includegraphics[scale=.0473684]{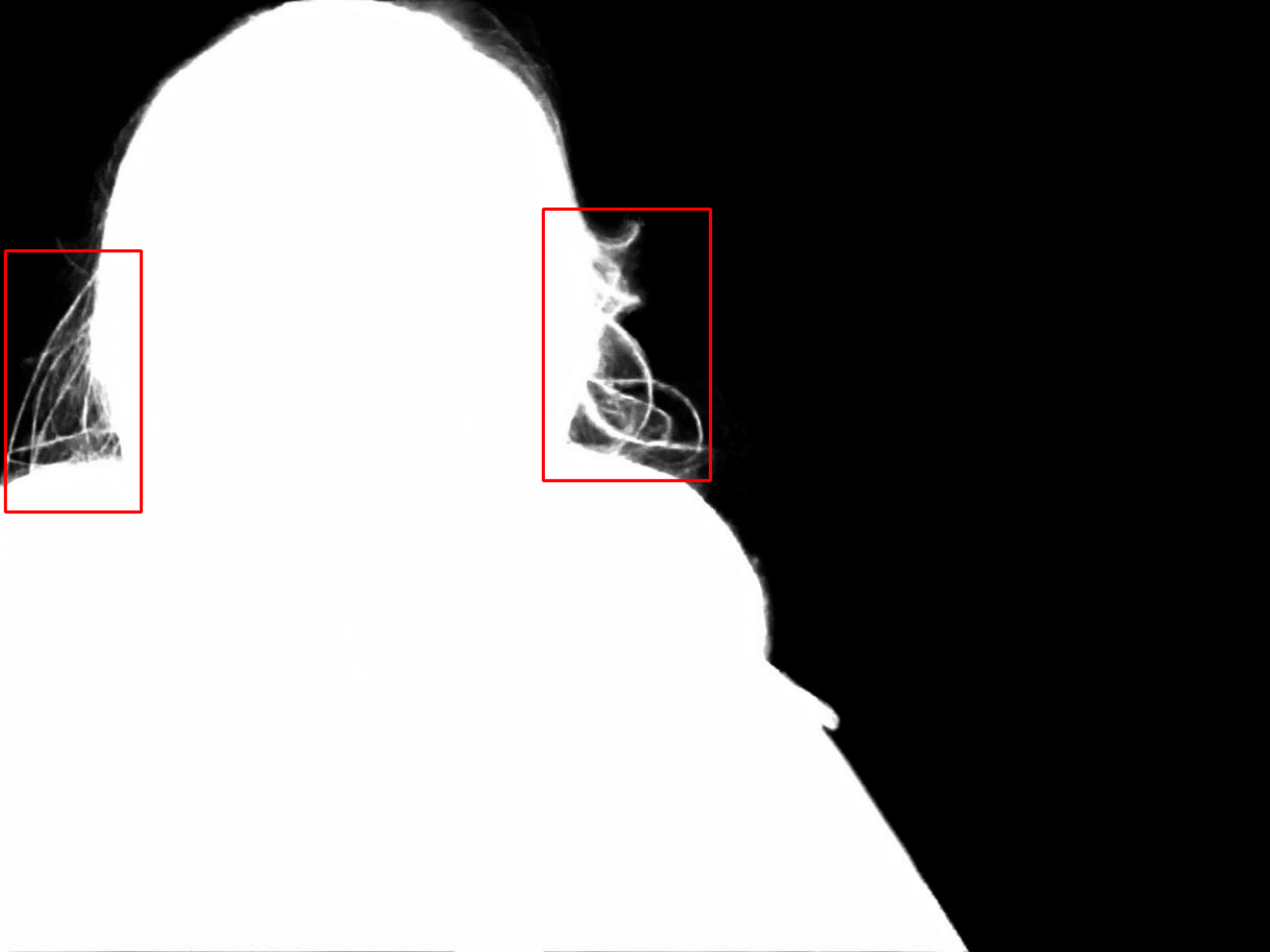}}\vspace{2pt}
     \centerline{\includegraphics[scale=.1056880734]{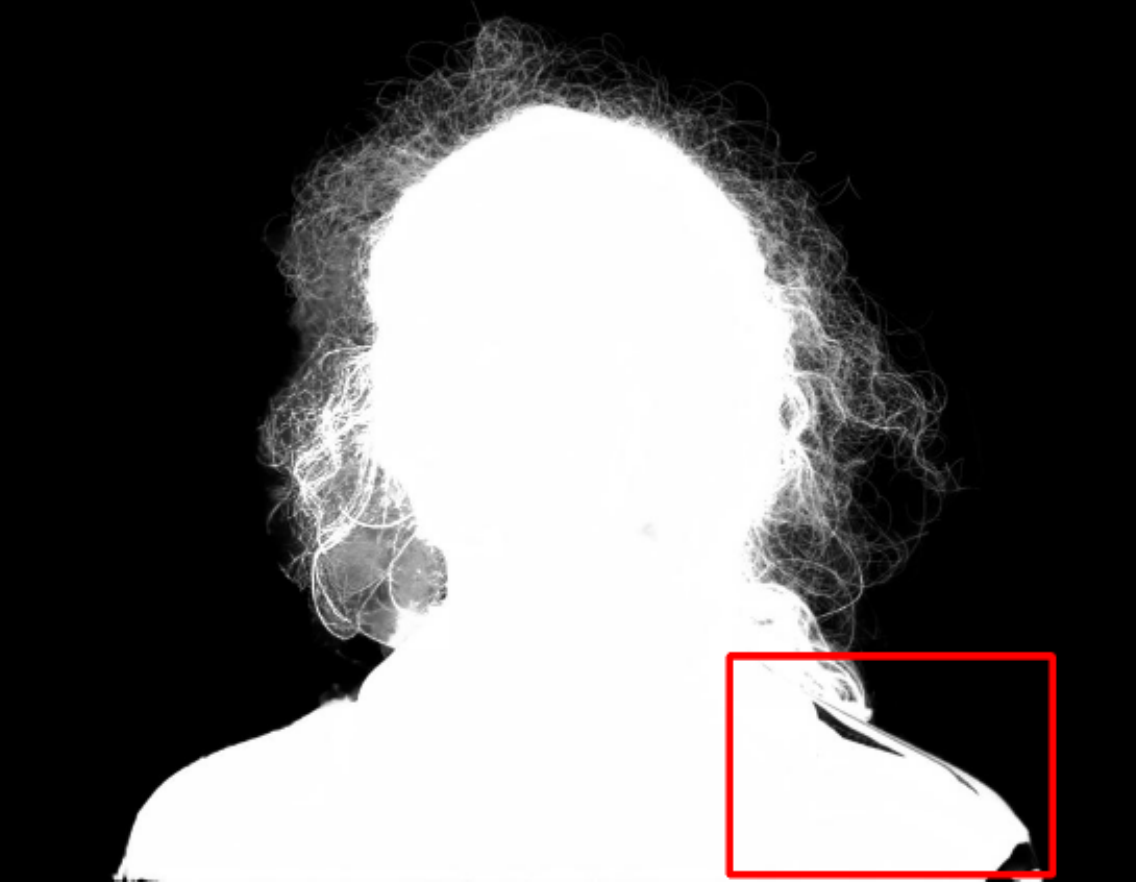}}  
    \centerline{LD-P-CK}
    \end{minipage}
    \hfill
    \begin{minipage}[t]{0.11\textwidth}
    \centerline{\includegraphics[scale=.1]{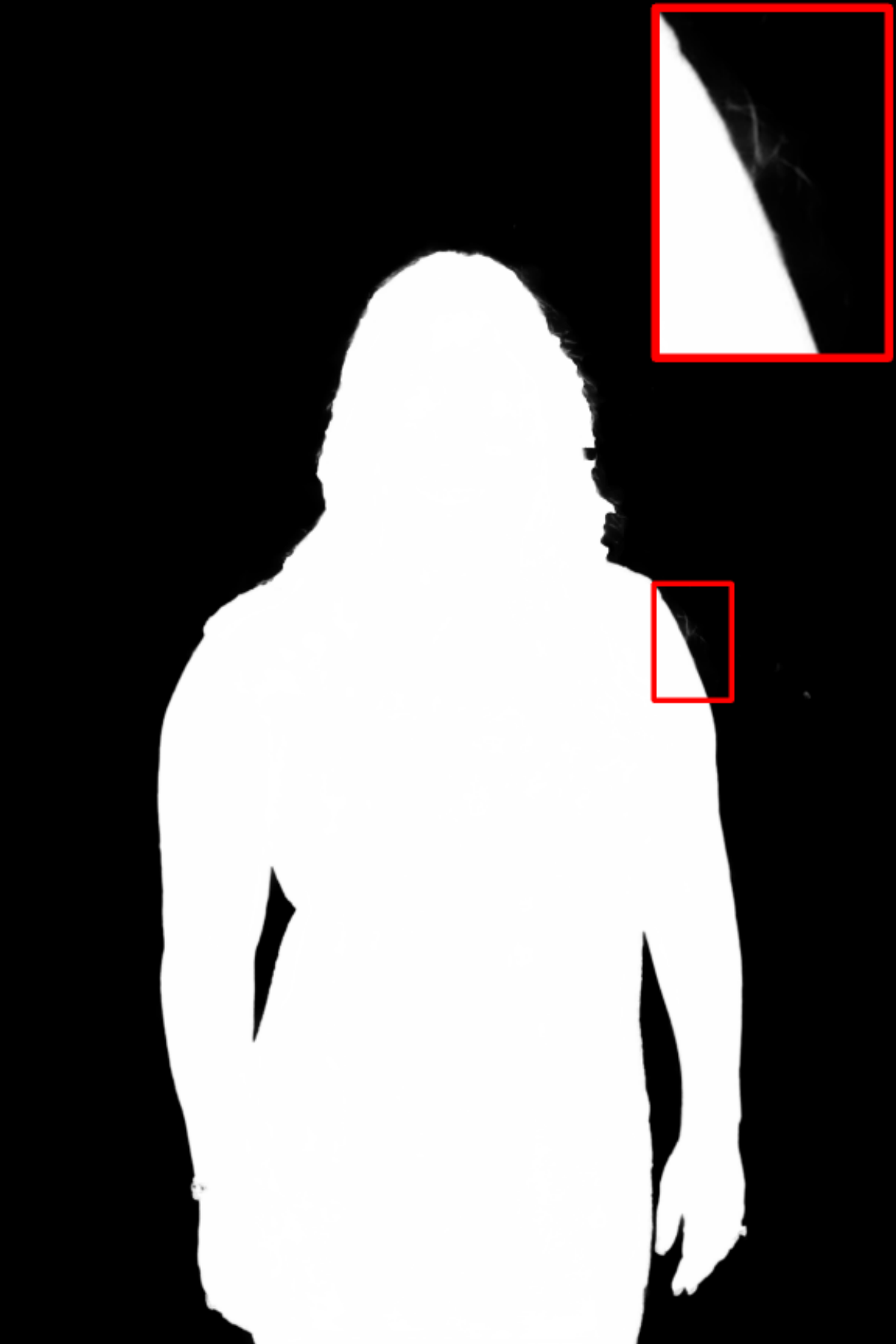}}\vspace{2pt}   
     \centerline{\includegraphics[scale=.1]{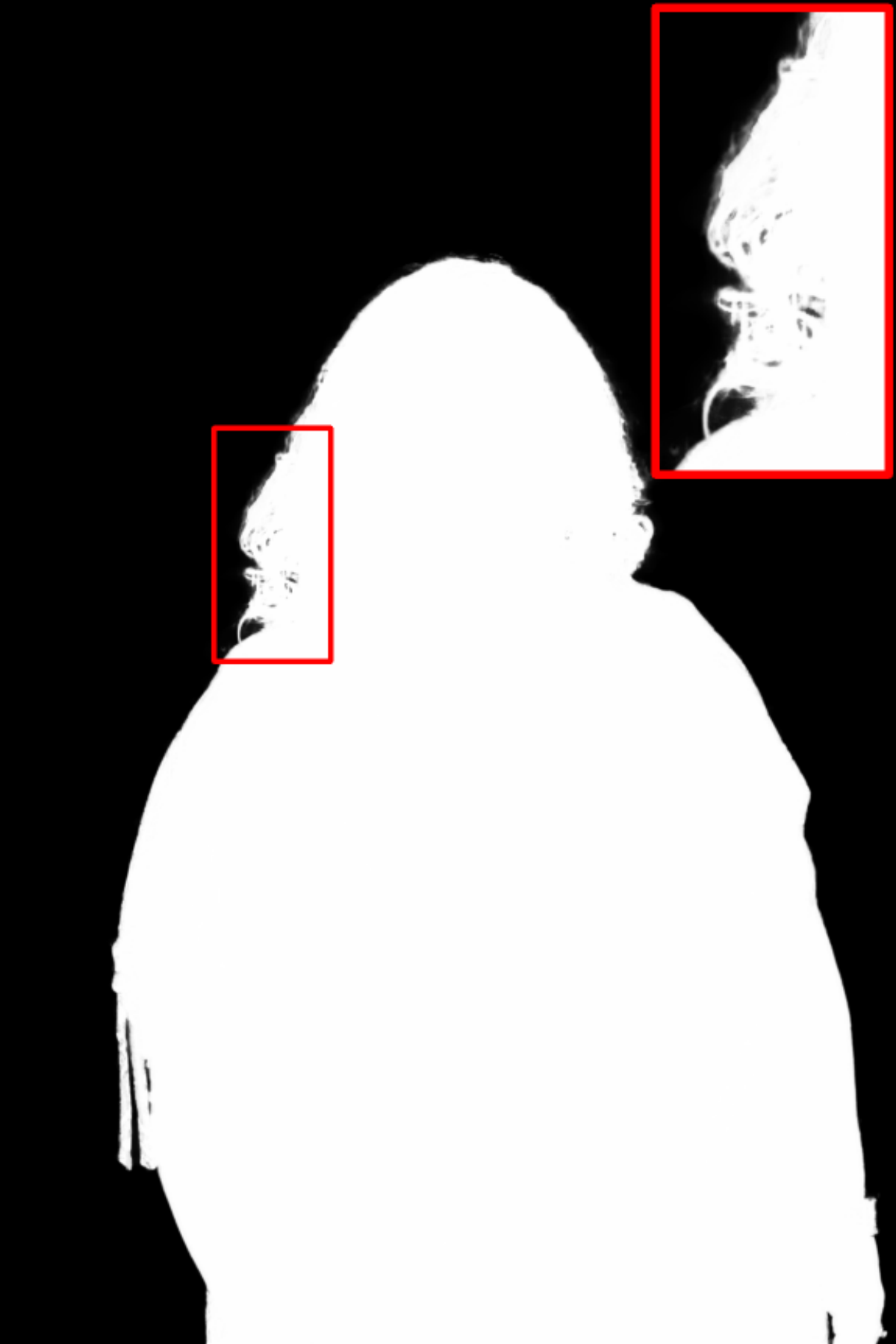}}\vspace{2pt} 
     \centerline{\includegraphics[scale=.1]{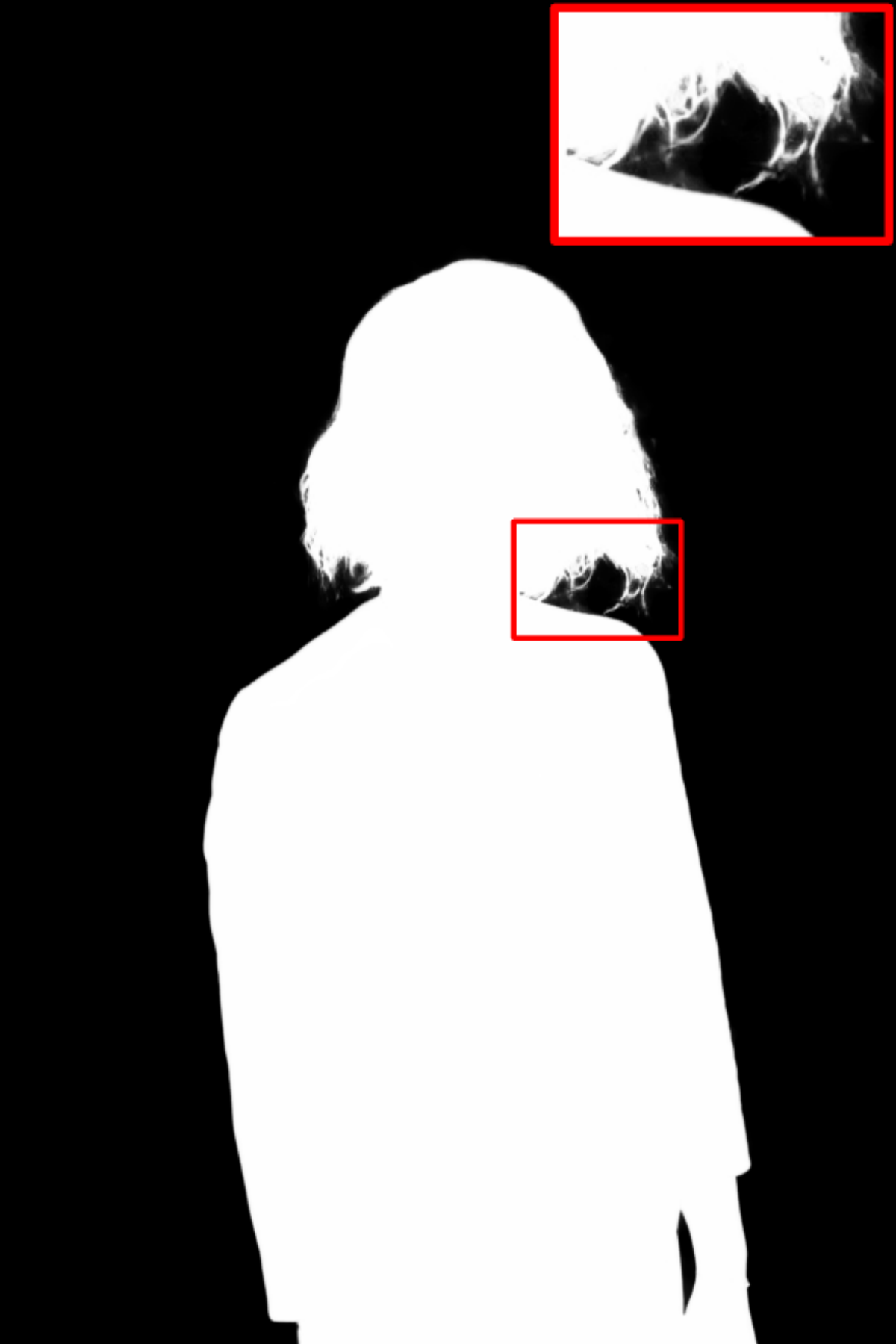}}\vspace{2pt}   
     % \centerline{\includegraphics[scale=.1]{figures/experiment/PH85-4/4_LD_M_10k.pdf}}\vspace{2pt}  
     \centerline{\includegraphics[scale=.1125]{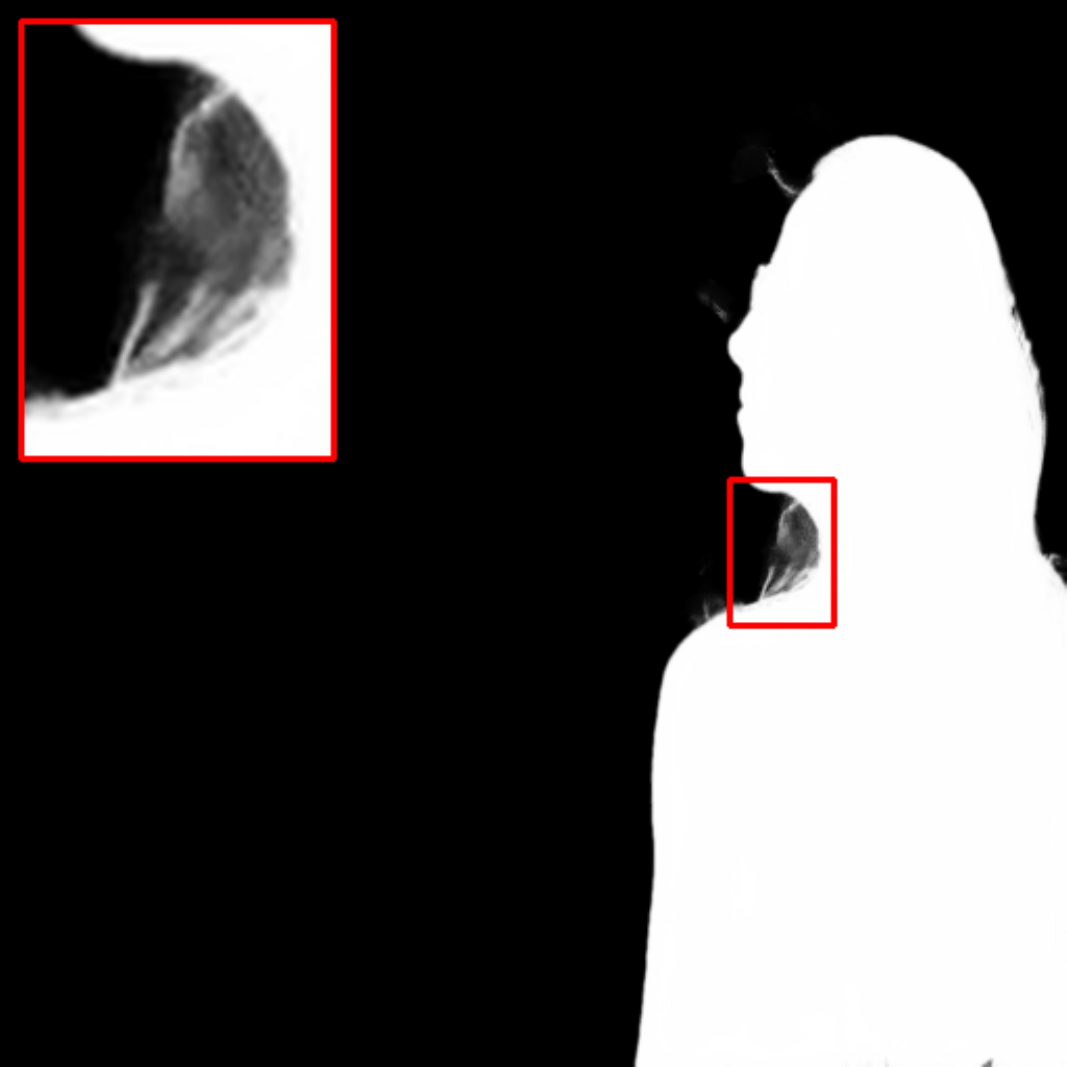}}\vspace{2pt}   
     \centerline{\includegraphics[scale=.1125]{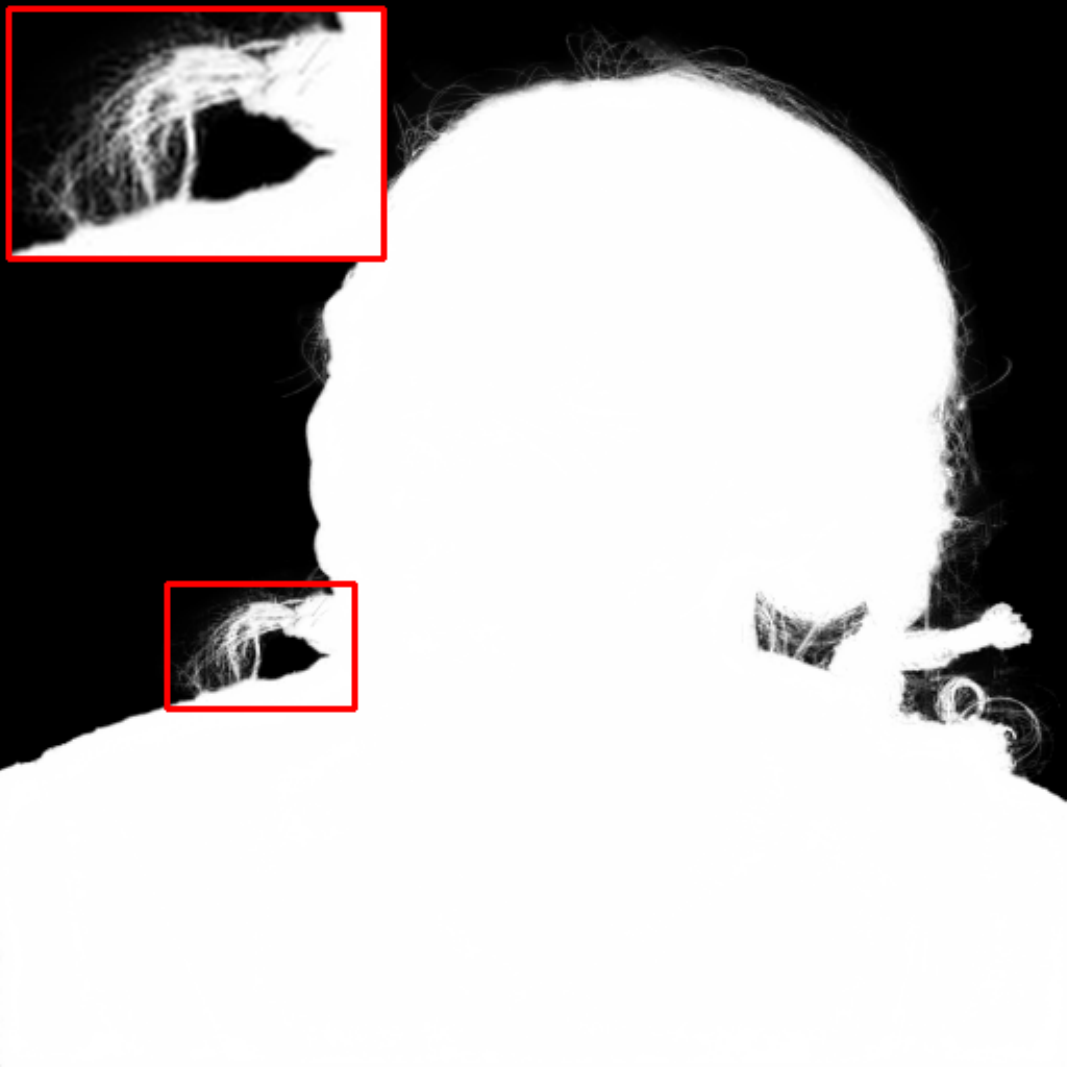}}\vspace{2pt} 
     \centerline{\includegraphics[scale=.1125]{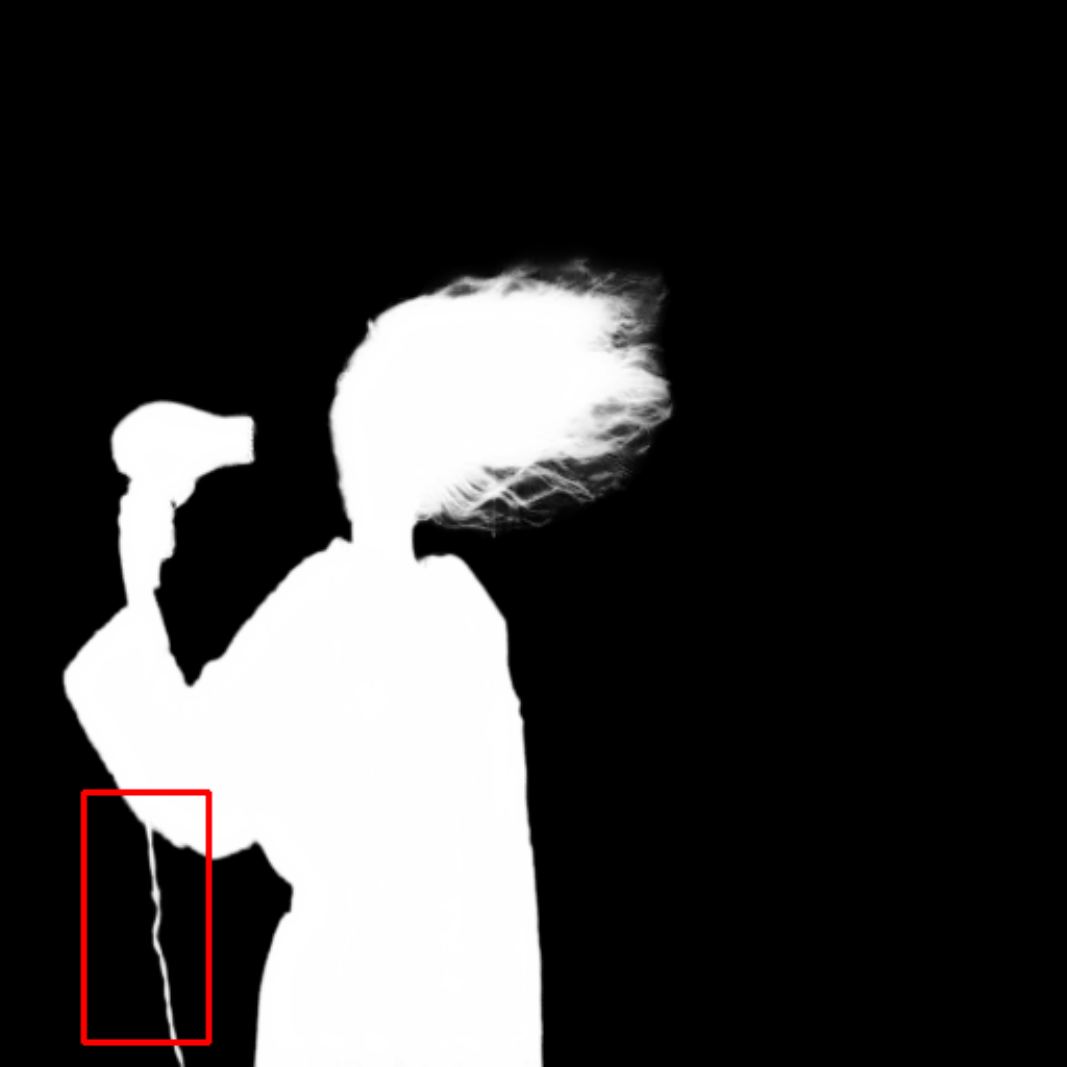}}\vspace{2pt}   
     \centerline{\includegraphics[scale=.1689]{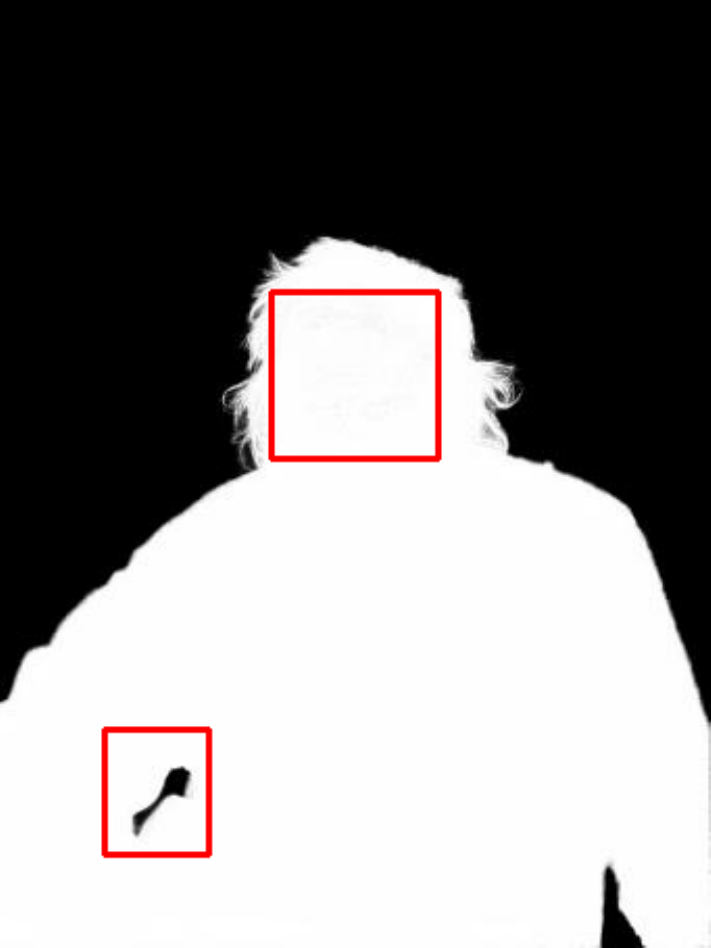}}\vspace{2pt}
     \centerline{\includegraphics[scale=.0473684]{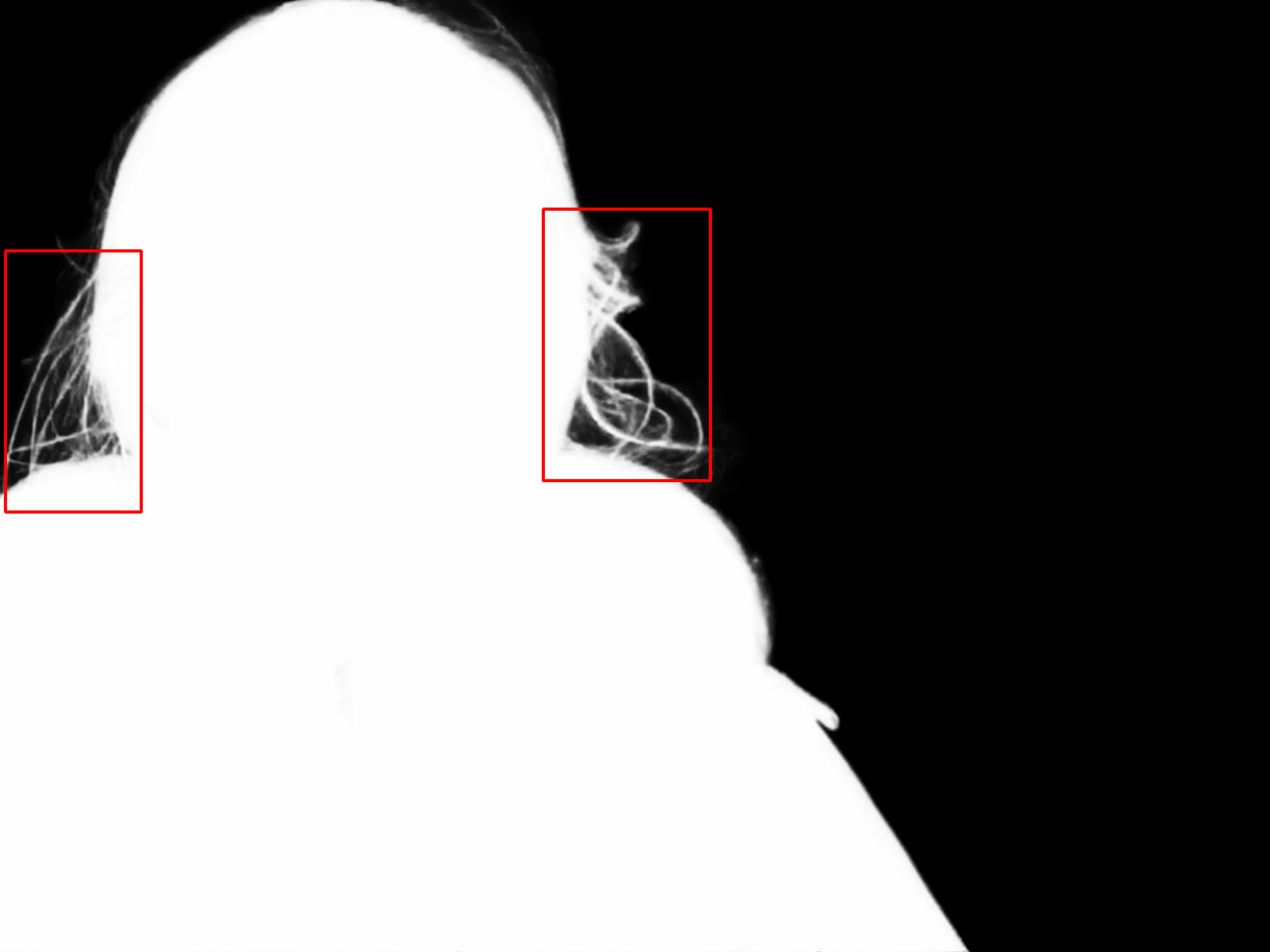}}\vspace{2pt}
     \centerline{\includegraphics[scale=.1056880734]{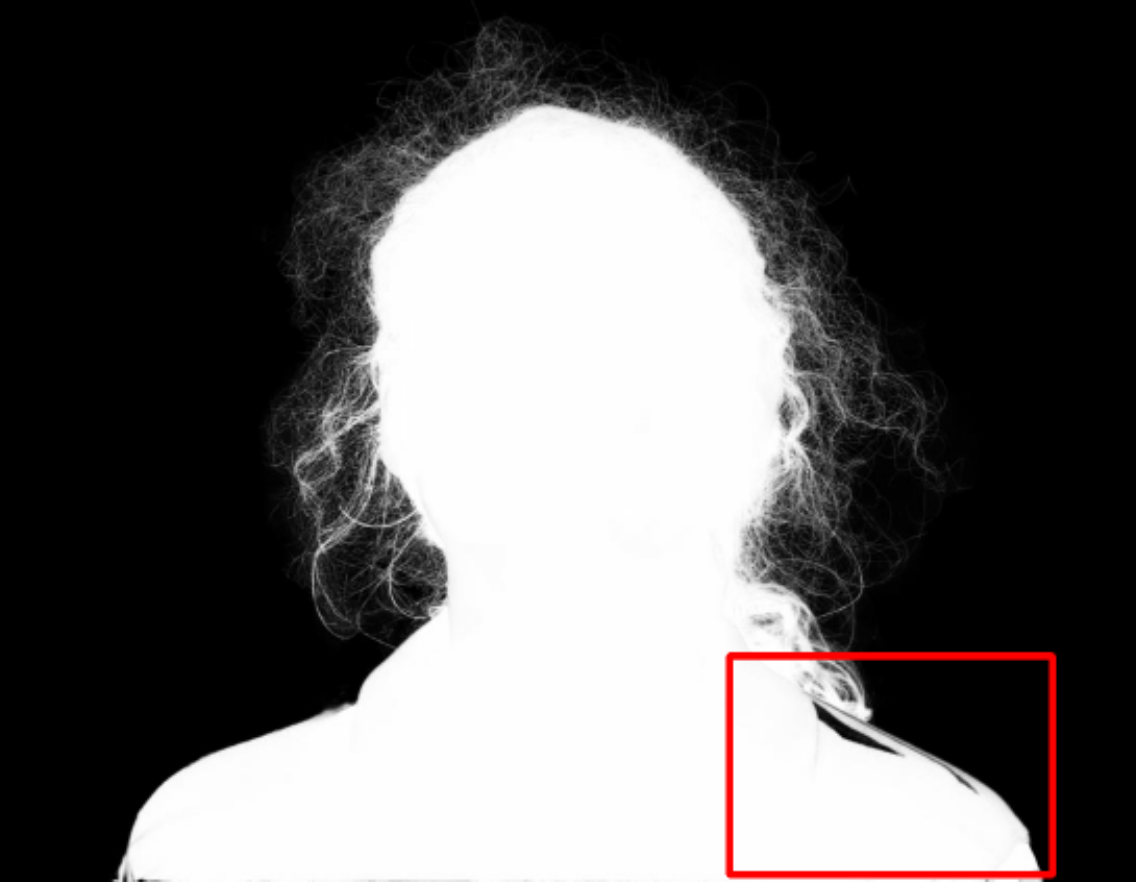}}  
    \centerline{LD-P-10K}
    \end{minipage}
    \hfill
    \begin{minipage}[t]{0.11\textwidth}
    \centerline{\includegraphics[scale=.1]{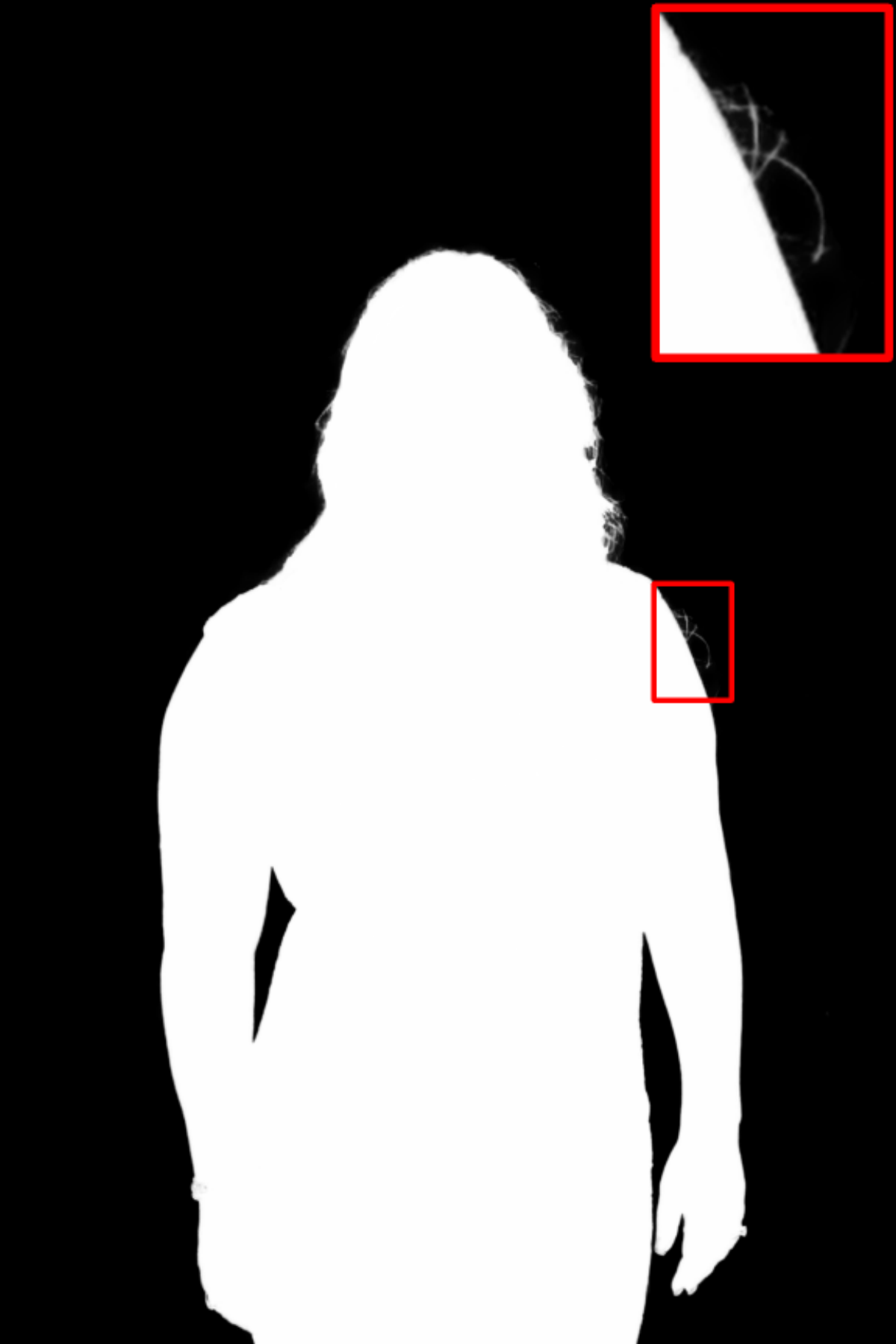}}\vspace{2pt}   
     \centerline{\includegraphics[scale=.1]{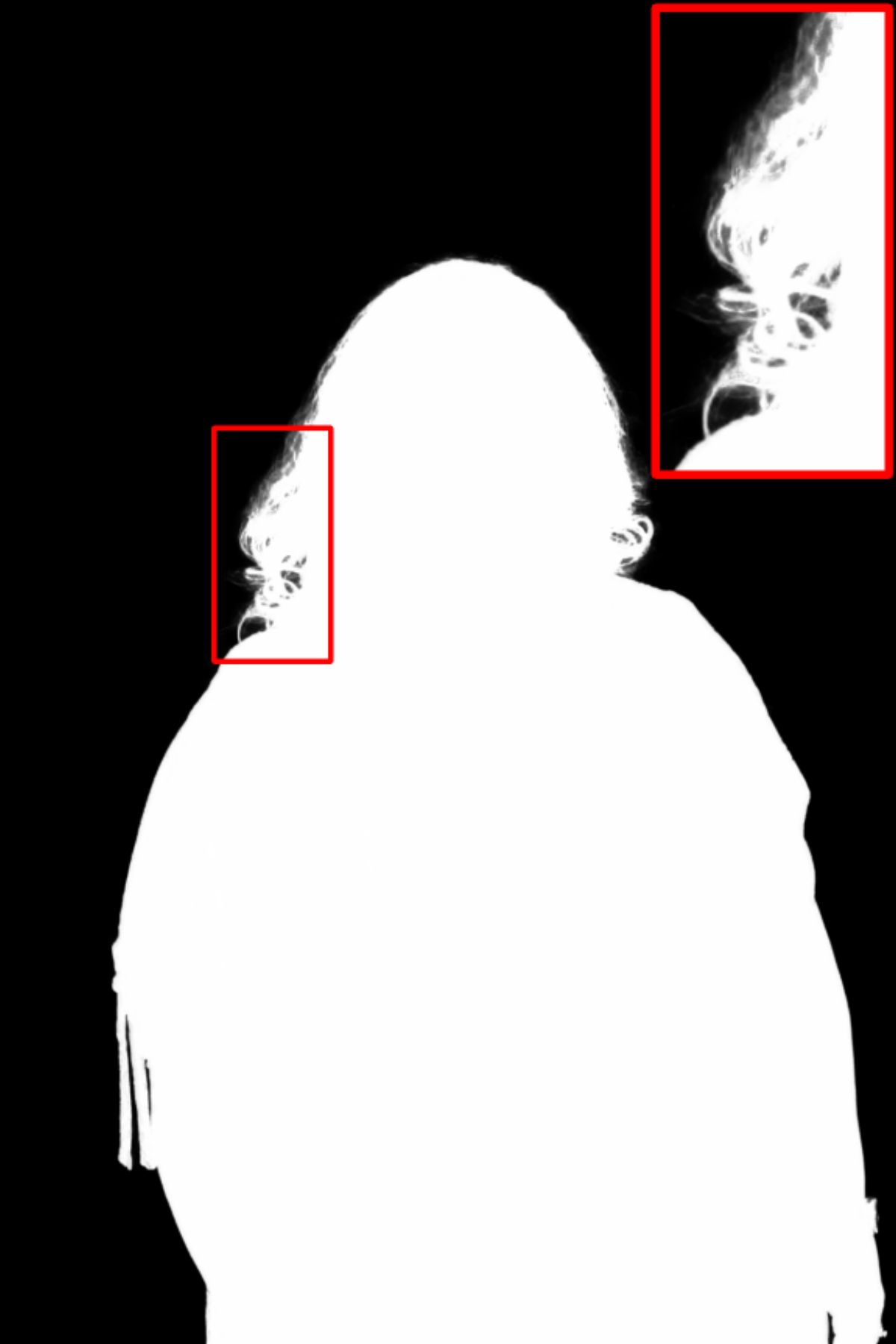}}\vspace{2pt} 
     \centerline{\includegraphics[scale=.1]{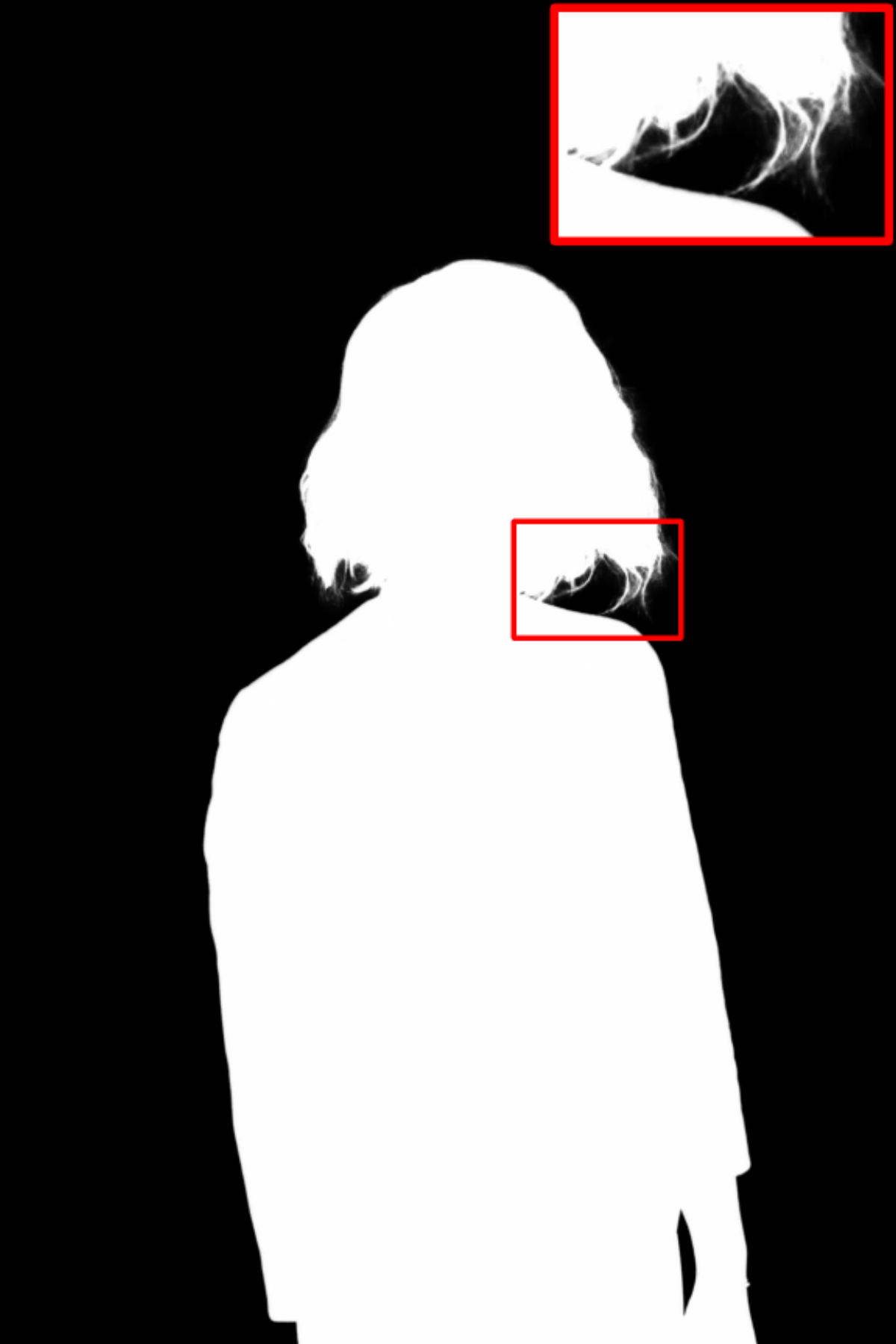}}\vspace{2pt}   
     % \centerline{\includegraphics[scale=.1]{figures/experiment/PH85-4/4_LD_M_20k.pdf}}\vspace{2pt}  
     \centerline{\includegraphics[scale=.1125]{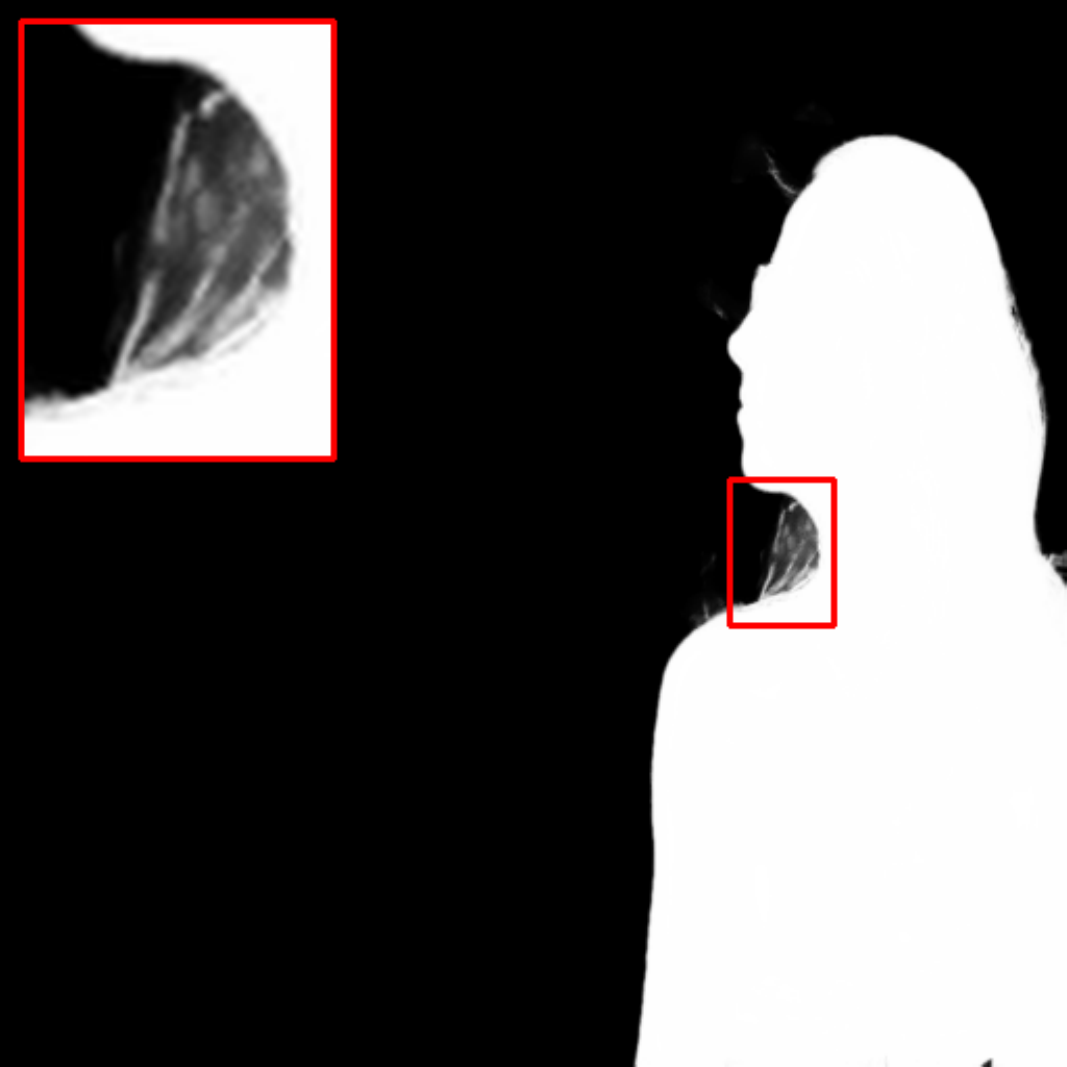}}\vspace{2pt}   
     \centerline{\includegraphics[scale=.1125]{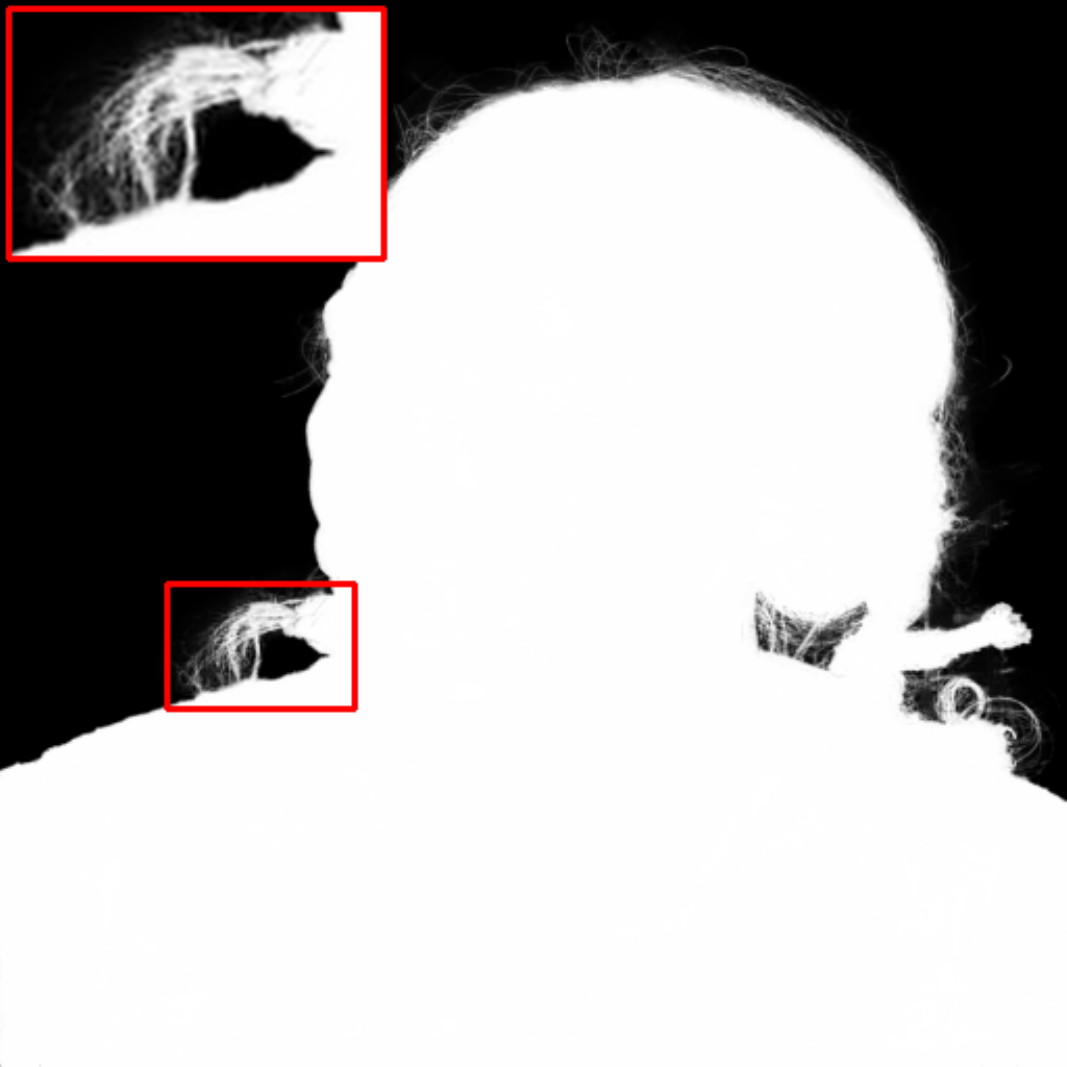}}\vspace{2pt} 
     \centerline{\includegraphics[scale=.1125]{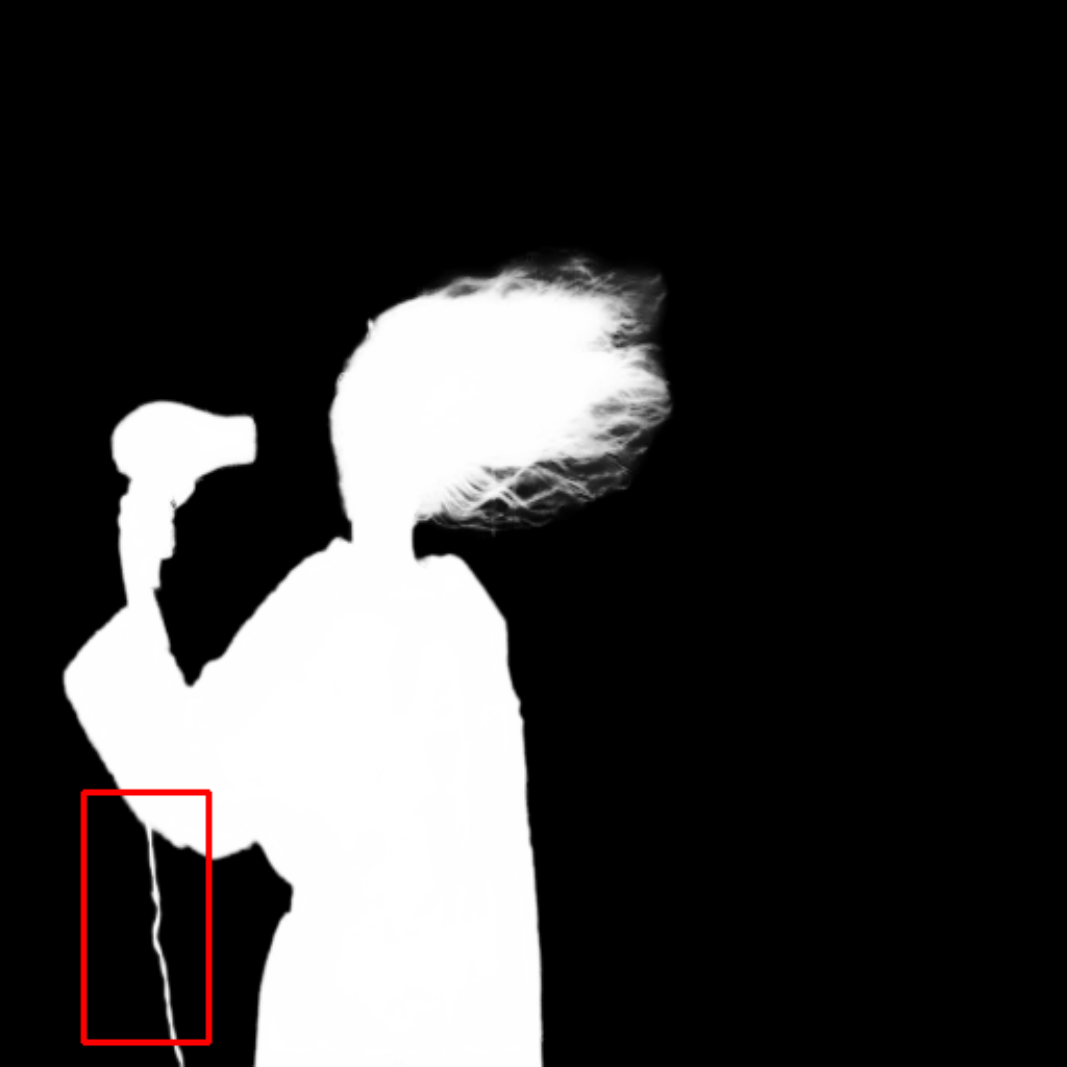}}\vspace{2pt}   
     \centerline{\includegraphics[scale=.1689]{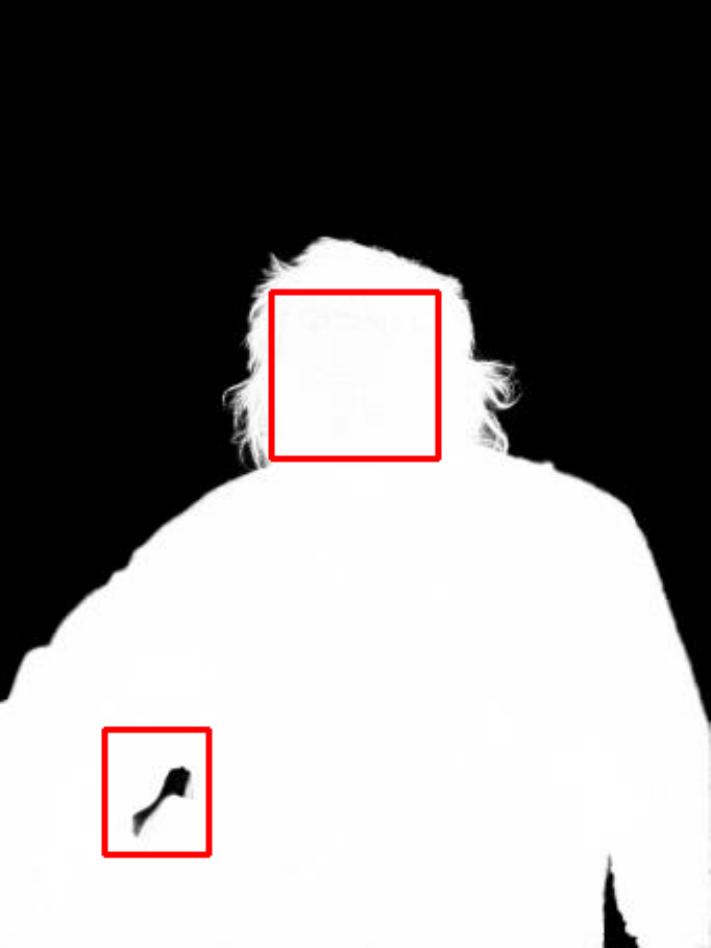}}\vspace{2pt}
     \centerline{\includegraphics[scale=.0473684]{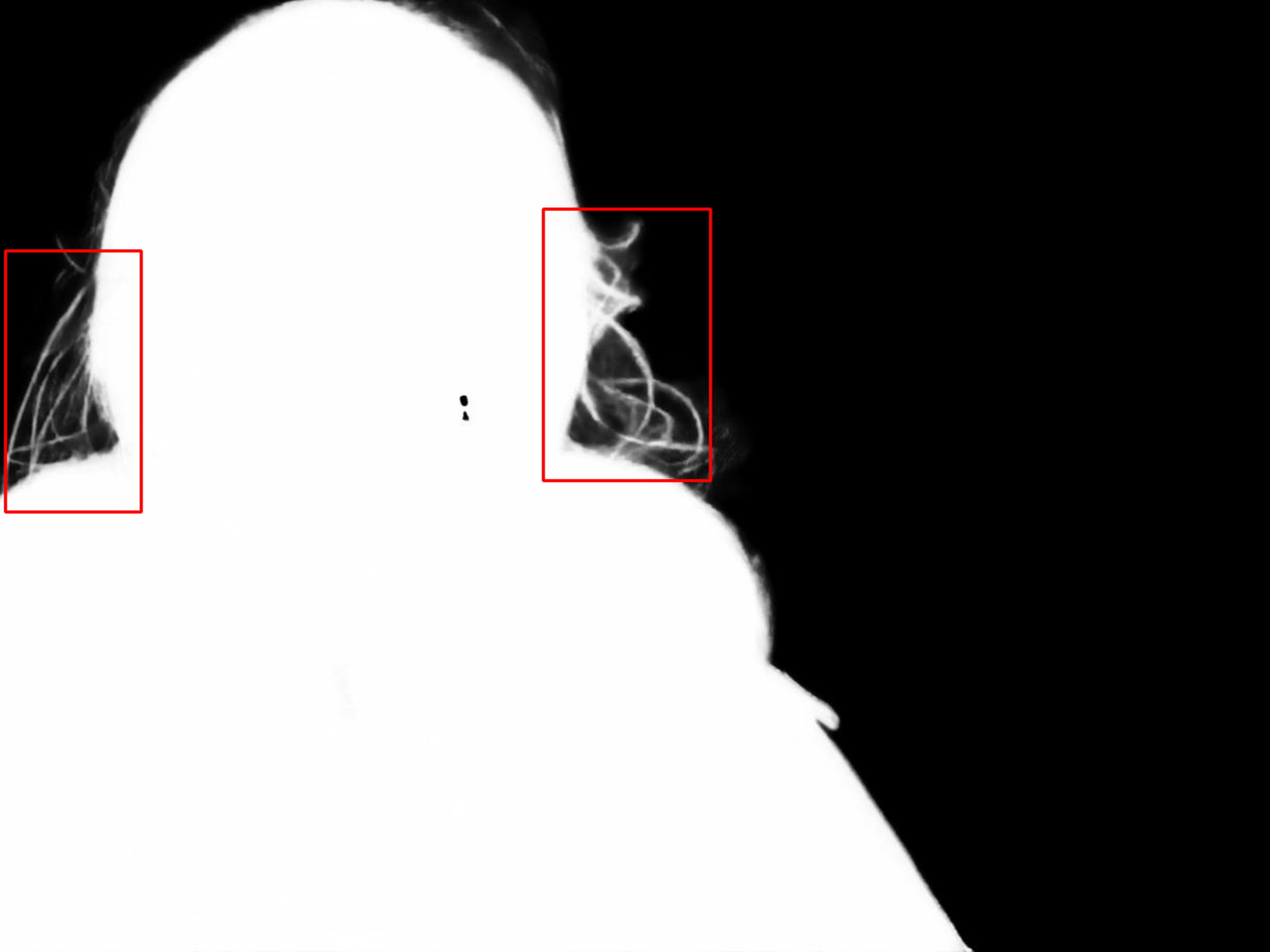}}\vspace{2pt}
     \centerline{\includegraphics[scale=.1056880734]{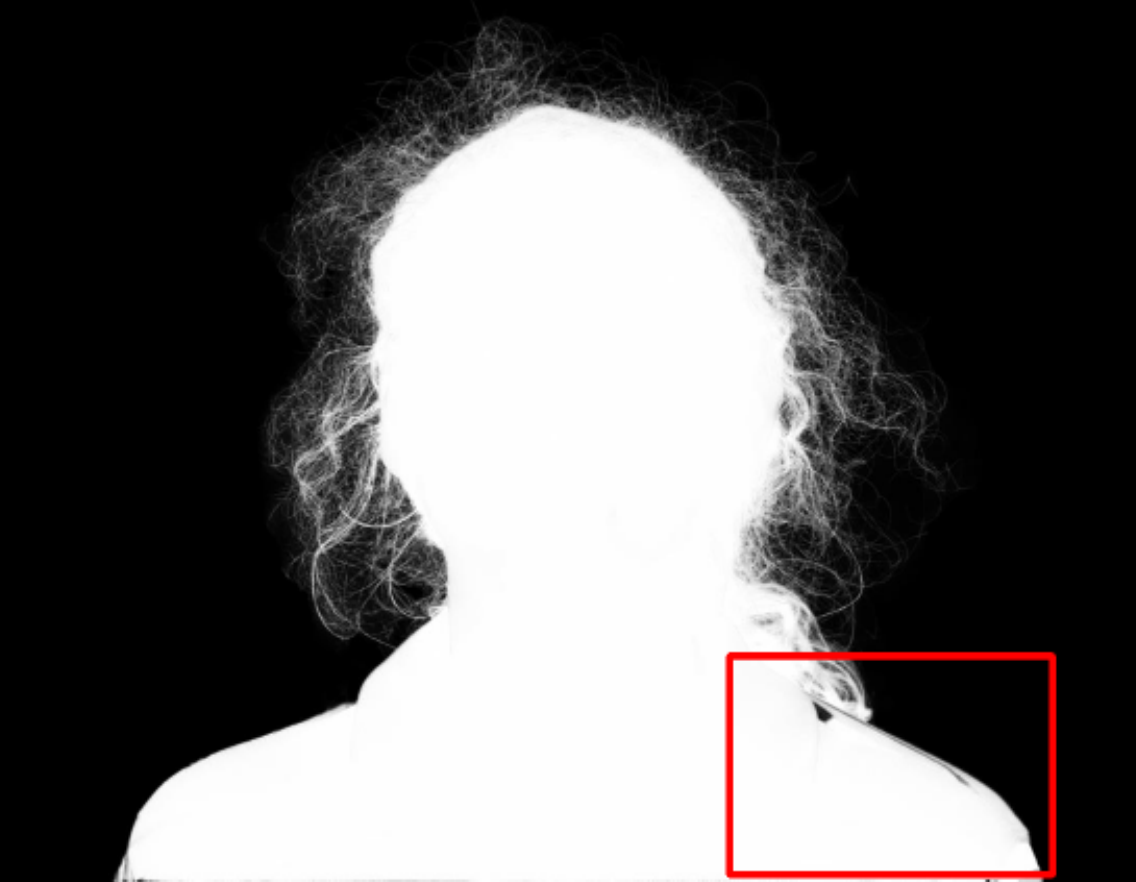}}  
    \centerline{LD-P-20K}
    \end{minipage}
    \hfill
    \begin{minipage}[t]{0.11\textwidth}
    \centerline{\includegraphics[scale=.1]{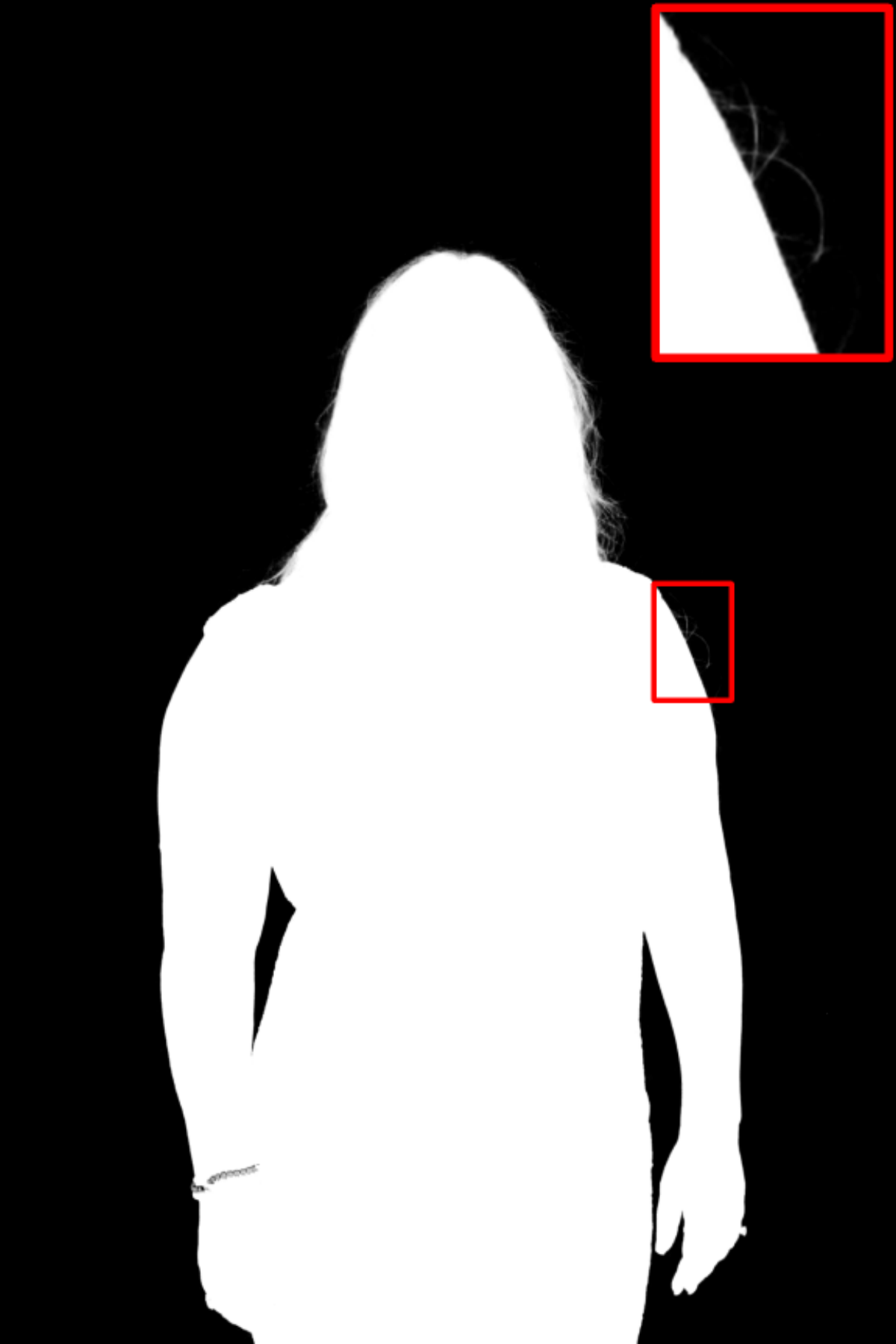}}\vspace{2pt}   
     \centerline{\includegraphics[scale=.1]{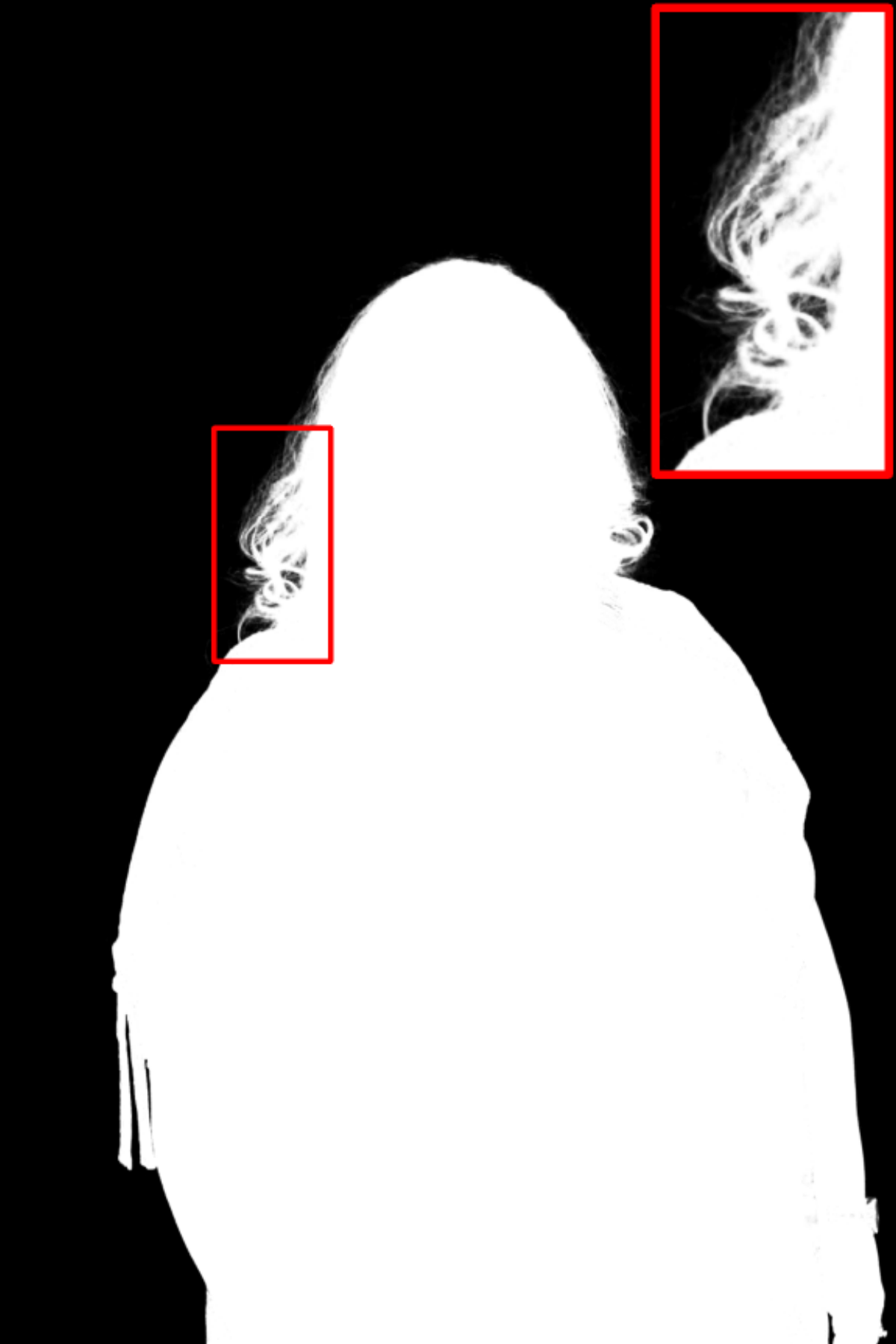}}\vspace{2pt} 
     \centerline{\includegraphics[scale=.1]{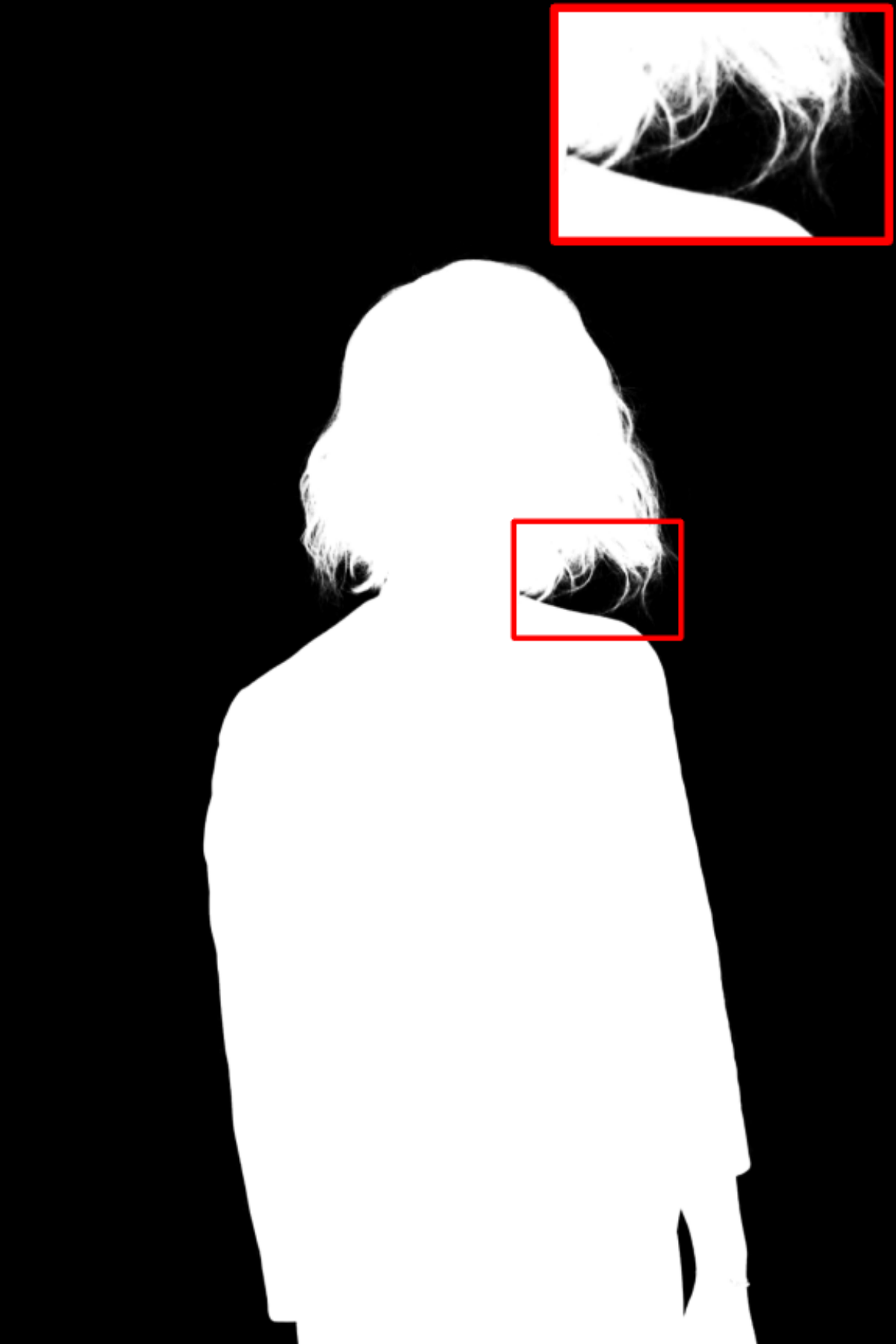}}\vspace{2pt}   
     % \centerline{\includegraphics[scale=.1]{figures/experiment/PH85-4/4_gt.pdf}}\vspace{2pt}  
     \centerline{\includegraphics[scale=.1125]{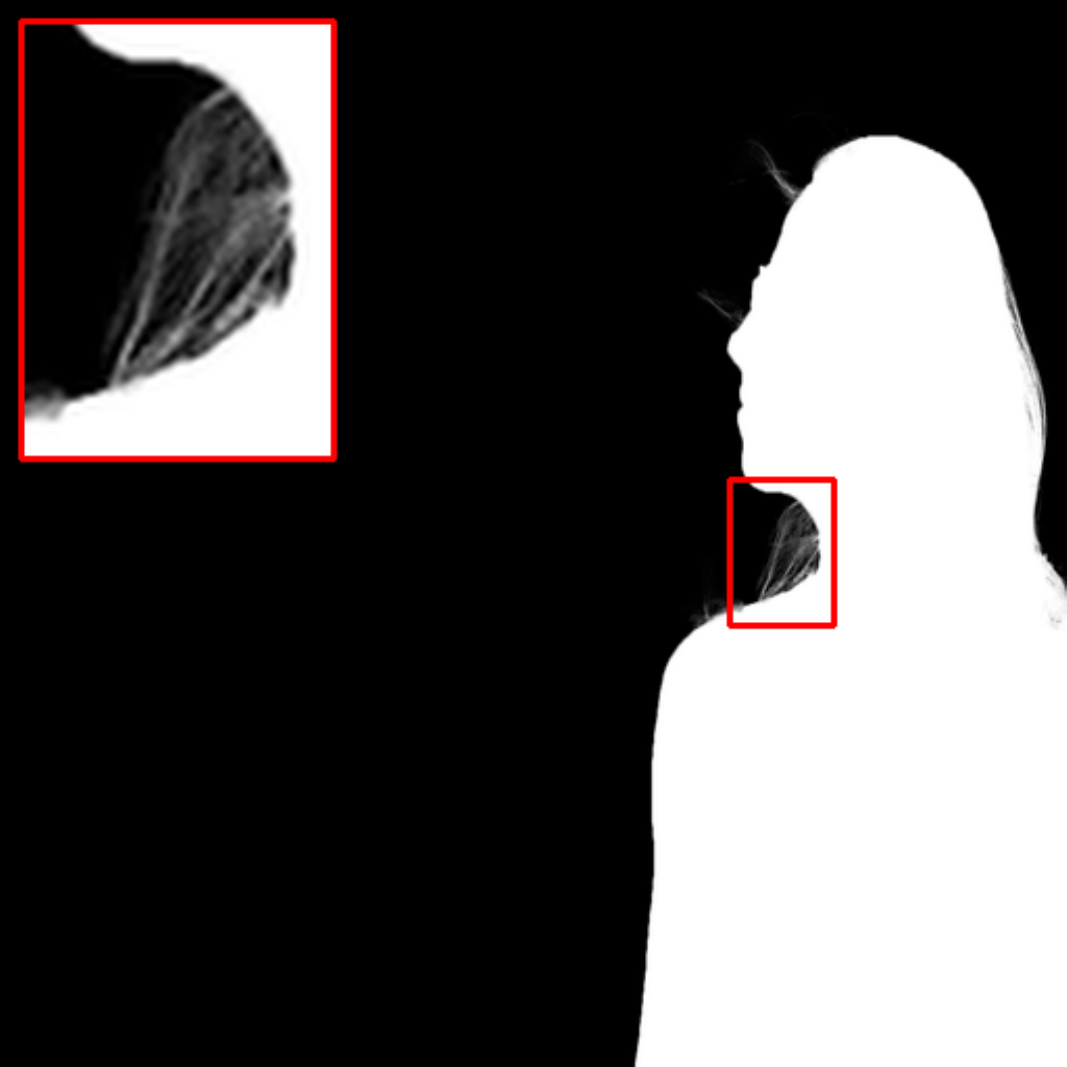}}\vspace{2pt}   
     \centerline{\includegraphics[scale=.1125]{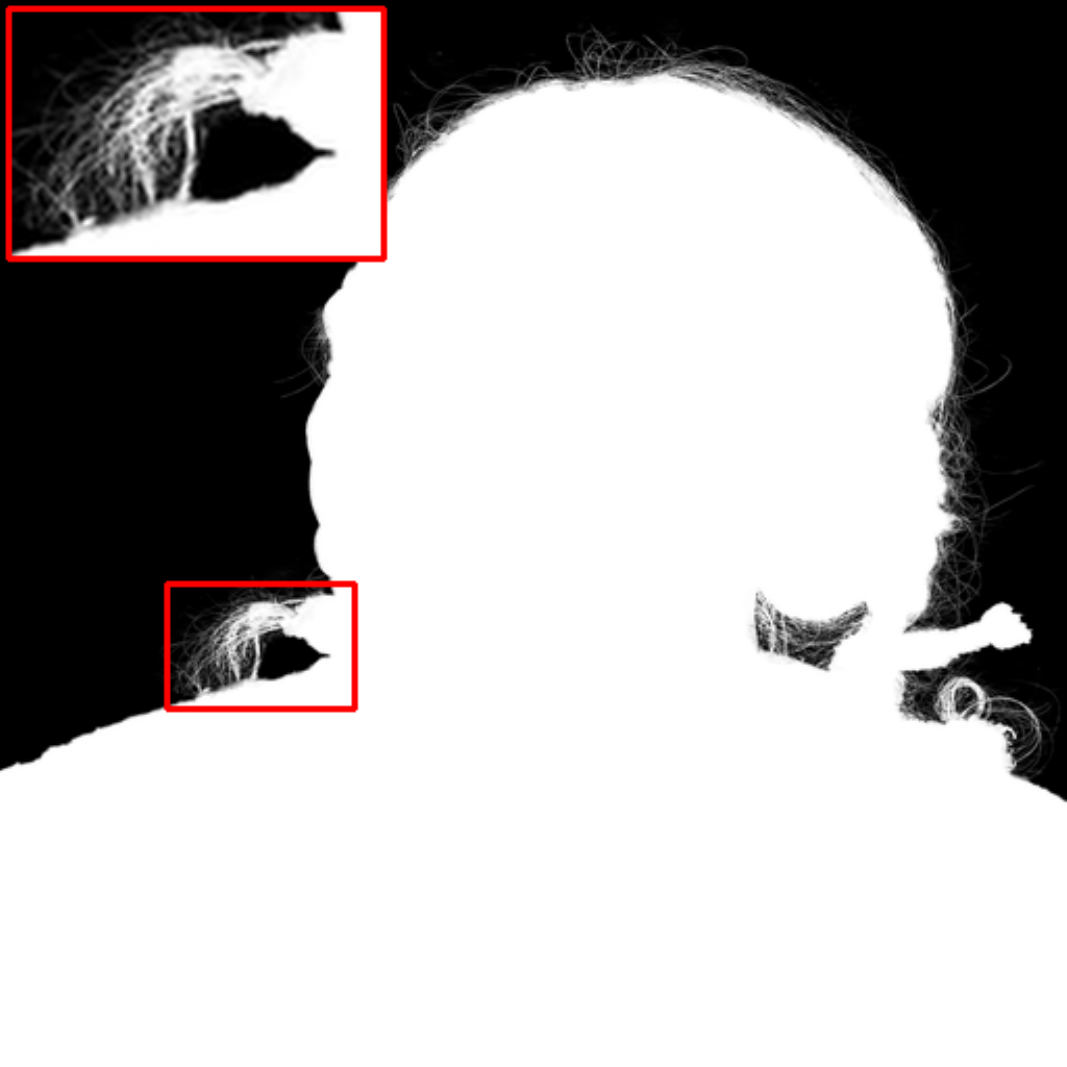}}\vspace{2pt} 
     \centerline{\includegraphics[scale=.1125]{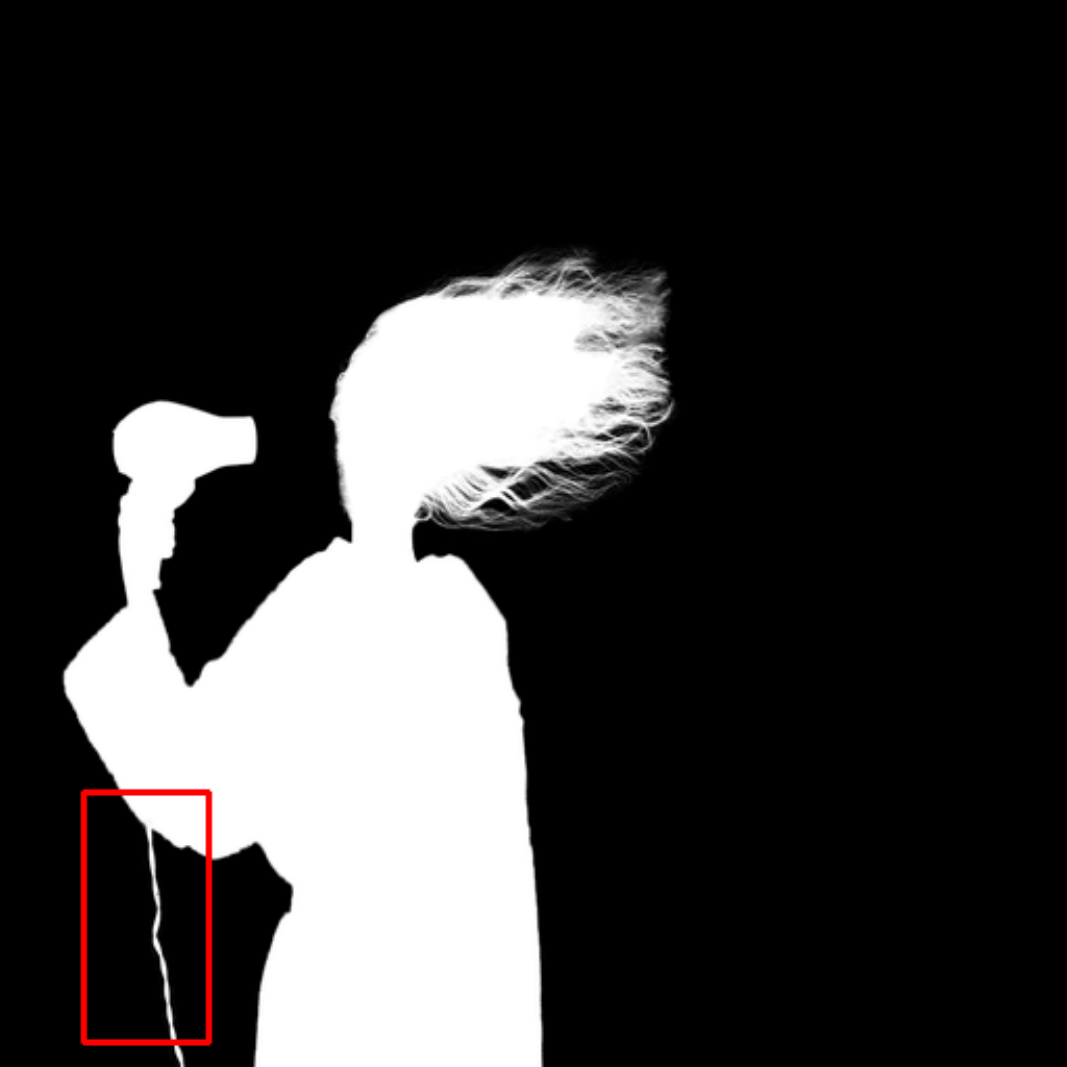}}\vspace{2pt}   
     \centerline{\includegraphics[scale=.1689]{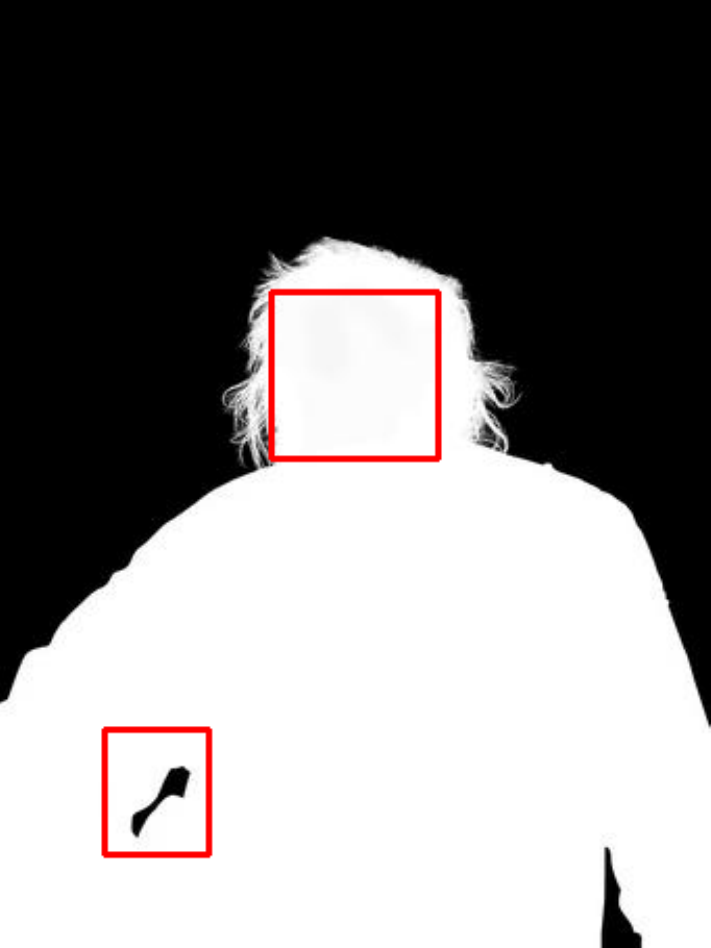}}\vspace{2pt}
     \centerline{\includegraphics[scale=.0473684]{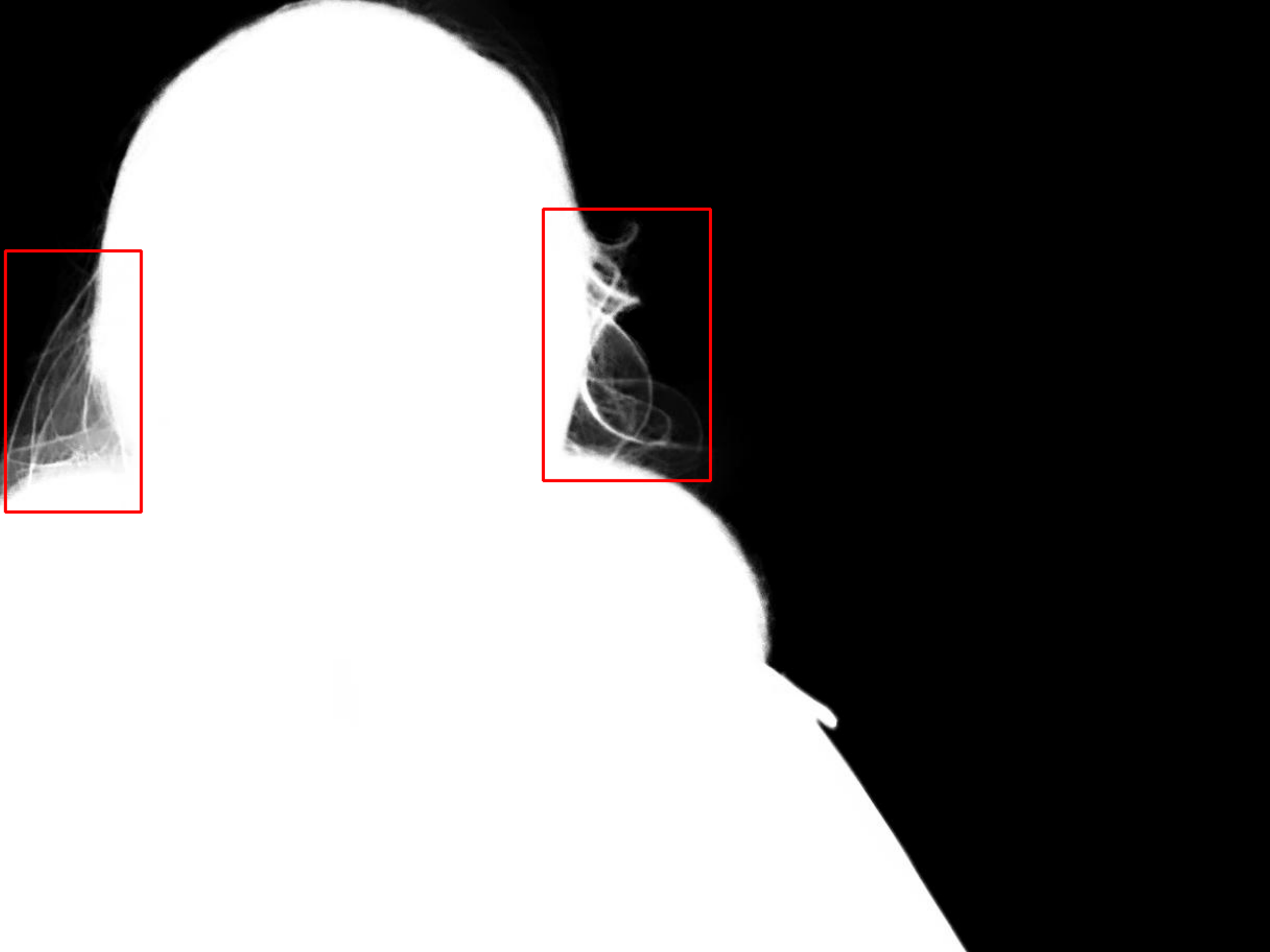}}\vspace{2pt}
     \centerline{\includegraphics[scale=.1056880734]{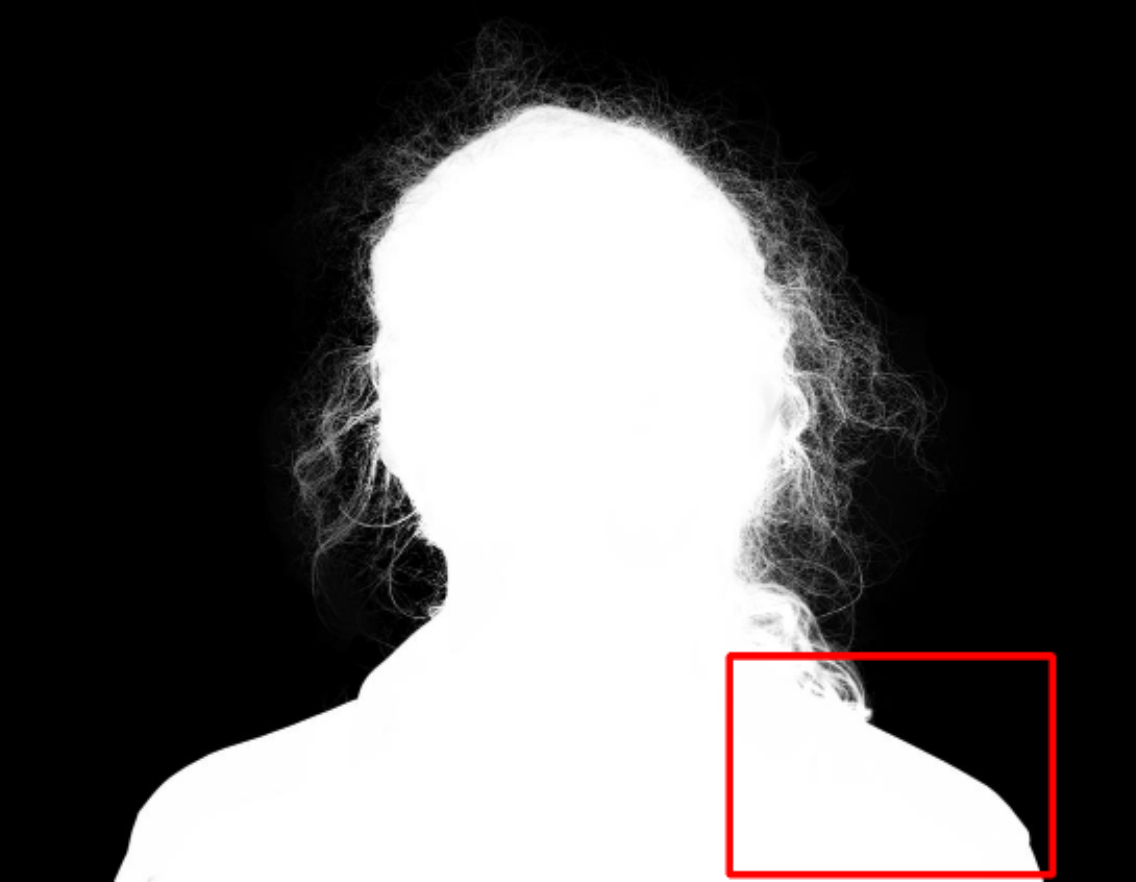}}  
    \centerline{GT}
    \end{minipage}
    }
    \caption{\textbf{Visual comparison results on PhotoMatte85, AIM and PPM-100}. The names of the datasets at the bottom represent the ViTMatte-B models trained by these datasets. LD-P-CK, LD-P-10K, and LD-P-20K refer to LD-Portrait-CK, LD-Portrait-10K, and LD-Portrait-20K, respectively.}
    \label{figure-exp-1}
\end{figure*}

Figure~\ref{figure-exp-1} illustrates the qualitative comparison results. Given that handling edge details, such as hair, is the most challenging aspect of portrait matting, examples focusing on human edge regions were selected. The results show that the model trained on LD-Portrait-20K delivers superior alpha extraction along edge regions, preserving intricate details with exceptional robustness. Compared to models trained on other datasets, LD-Portrait-20K-trained models exhibit a stronger ability to handle complex edge transitions, effectively reducing under-matting and over-matting errors.

Overall, these experimental results strongly validate the effectiveness of the LD-Portrait-20K dataset and emphasize its significant contribution to the field of portrait matting. Both quantitative metrics and qualitative comparisons highlight LD-Portrait-20K’s advantages in advancing portrait matting technology, offering a powerful resource for future research.

\begin{table}[!t]
% \large
\caption{Training of Different Datasets on ViTMatte-B and GFM and Comparison on AIM Dataset.\label{table2}}
\centering
% \Large
% \normalsize
\begin{threeparttable}
\begin{tabular}{@{}lccccc@{}}
\toprule
% Method & Training Dataset & MAD($\downarrow$) & MSE($\downarrow$) & Grad($\downarrow$) & Conn($\downarrow$)  \\
Method & Training Dataset & MAD & MSE & Grad & Conn  \\
\midrule
DeepLabv3~\cite{chen2017rethinking}  &\multirow{5}{*}{\XSolidBrush} &29.64 &23.78 &20.17 &7.71 \\
MODnet~\cite{ke2022modnet}  & &21.66 &14.27 &5.37 &5.23 \\
BGMv2~\cite{lin2021real} & &44.61 &39.08 &5.54 &11.60 \\
RVM~\cite{lin2022robust} & &\underline{14.84} &\underline{8.93} &\underline{4.35} &\underline{3.83}  \\
EFormer~\cite{wang2024eformer} & &\textbf{7.47} &\textbf{2.13} &\textbf{2.83} &\textbf{1.90}\\
\midrule
\multirow{6}{*}{ViTMatte-B} &HHM50K &6.82	&1.25	&1.33	&1.56 \\
 & P3M-10k &5.84	&1.12	&1.19	&1.30 \\
 &ImageMatte&\underline{5.65}	&1.08	&1.24	&\underline{1.27} \\
 &LD-Portrait-CK &6.23 &1.15 &1.13 &1.45 \\
 &LD-Portrait-10K &5.76 &\underline{1.03} &\underline{0.99} &1.29 \\
 &LD-Portrait-20K & \textbf{5.64} & \textbf{1.01} & \textbf{0.96} & \textbf{1.25} \\

\midrule
\multirow{7}{*}{GFM} &HHM50K &105.02 &95.11 &15.55 &27.33 \\
 & P3M-10k &68.30 &59.14 &13.62 &17.80 \\
 &ImageMatte &\underline{55.33} &\underline{45.06} &12.24 &16.92 \\
 &LD-Portrait-CK &60.69 &51.79 &12.60 &15.77 \\
 &LD-Portrait-Ori &88.93 &78.26 &13.35 &22.86 \\
 &LD-Portrait-10K &57.77 &47.66 &\underline{11.18} &\underline{14.89}\\
 &LD-Portrait-20K &\textbf{54.67}	&\textbf{44.60}	&\textbf{10.90}	&\textbf{14.05} \\
\bottomrule
\end{tabular}
\end{threeparttable} 
\end{table}

\begin{table*}[!t]
% \large
\caption{Comparison between the Training Results of the S+V Framework on Diverse Datasets and Other Video Matting Models.\label{table3}}
\centering
% \Large
% \normalsize
\begin{threeparttable}
% \begin{tabular}{c|c|ccccc|ccccc}
\begin{tabular}{@{}lcccccccccccc@{}}
\toprule
\multirow{2}{*}{Method} &\multirow{2}{*}{Training Dataset} & \multicolumn{5}{c}{videomatte$512\times288$}    & \multicolumn{5}{c}{vmformer$512\times288$public} \\
% \cmidrule(lr){3-7}\cmidrule(lr){8-12} & & MAD($\downarrow$)  & MSE($\downarrow$)  & Grad($\downarrow$)  & Conn($\downarrow$) &dtSSD($\downarrow$) & MAD($\downarrow$)  & MSE($\downarrow$)  & Grad($\downarrow$)  & Conn($\downarrow$) &dtSSD($\downarrow$) \\ 
\cmidrule(lr){3-7}\cmidrule(lr){8-12} & & MAD  & MSE  & Grad  & Conn &dtSSD & MAD  & MSE  & Grad  & Conn &dtSSD \\ 
\midrule
RVM &\multirow{4}{*}{\XSolidBrush} &6.48 &1.28 &1.24 &0.48 &1.74 &6.63 &1.48 &1.24 &0.51 &1.61  \\
FTP-VM & &7.35 &2.19 &2.04 &0.6 &2.41 &6.67 &1.74 &1.73 &0.51 &2.12\\ 
SparseMat & &8.43	&2.72 &1.86 &0.75 &2.19 &8.55 &3.10	&1.72 &0.77	&2.20 \\
VMFormer & &6.06 &1.03 &0.76 &0.37 &1.73 &\underline{4.92} &0.56 &\underline{0.41} &0.26 &1.48 \\
\midrule
\multirow{4}{*}{S+V} &HHM50K &5.59	&0.69 &0.72 &0.26 &1.34 &5.39 &0.60	&0.58 &0.23	&1.21\\
 & P3M-10k &5.52	&0.61 &0.57	&\underline{0.22} &\underline{1.25} &5.42 &0.60	&0.54 &\underline{0.21} &\underline{1.15} \\
 &ImageMatte&\underline{5.37} &\underline{0.51} &\underline{0.50} &\underline{0.22} &1.31 &5.30 &\underline{0.48} &0.47 &\underline{0.21} &1.20 \\
 &LD-Portrait-20K & \textbf{4.60} & \textbf{0.20} & \textbf{0.19} & \textbf{0.10} &\textbf{0.83} & \textbf{4.51} & \textbf{0.17} & \textbf{0.14} & \textbf{0.09} &\textbf{0.76}\\

\bottomrule
\end{tabular}
\end{threeparttable} 
\end{table*}

\begin{figure*}
 \centering
    \subfloat{
    \begin{minipage}{0.11\textwidth}
        \includegraphics[scale=.13]{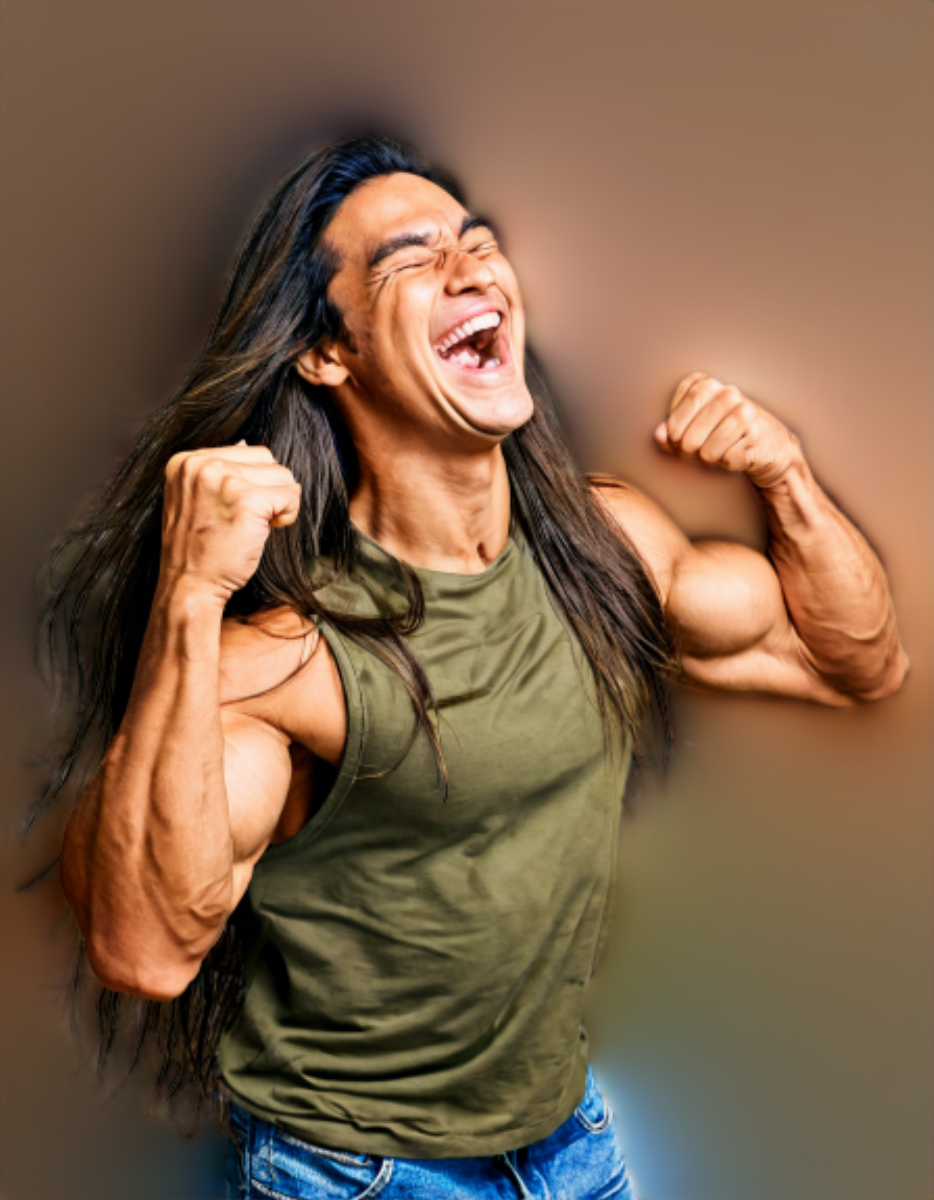}
    \end{minipage}
        \begin{minipage}{0.12\textwidth}
        \includegraphics[scale=.13]{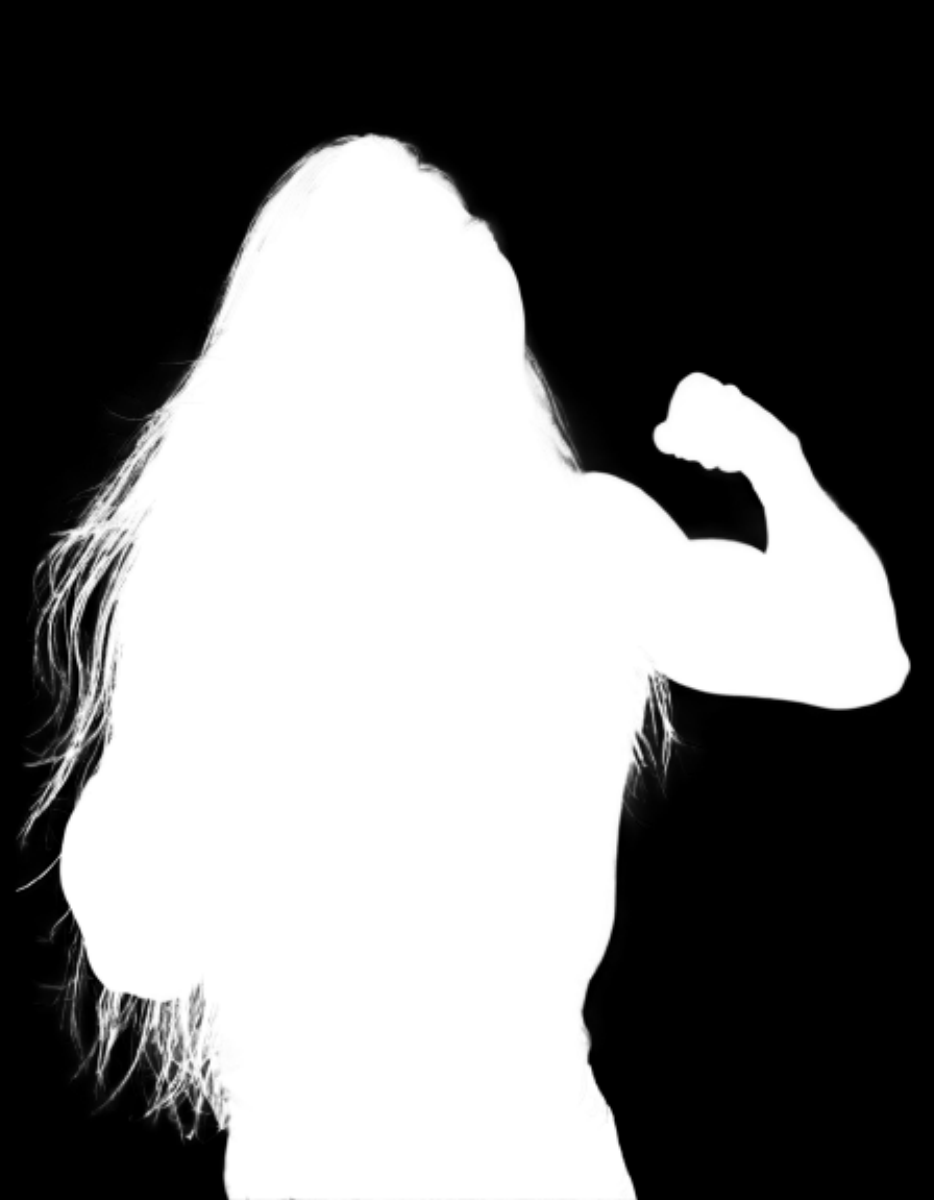}
    \end{minipage}
    \begin{minipage}{0.11\textwidth}
        \includegraphics[scale=.13]{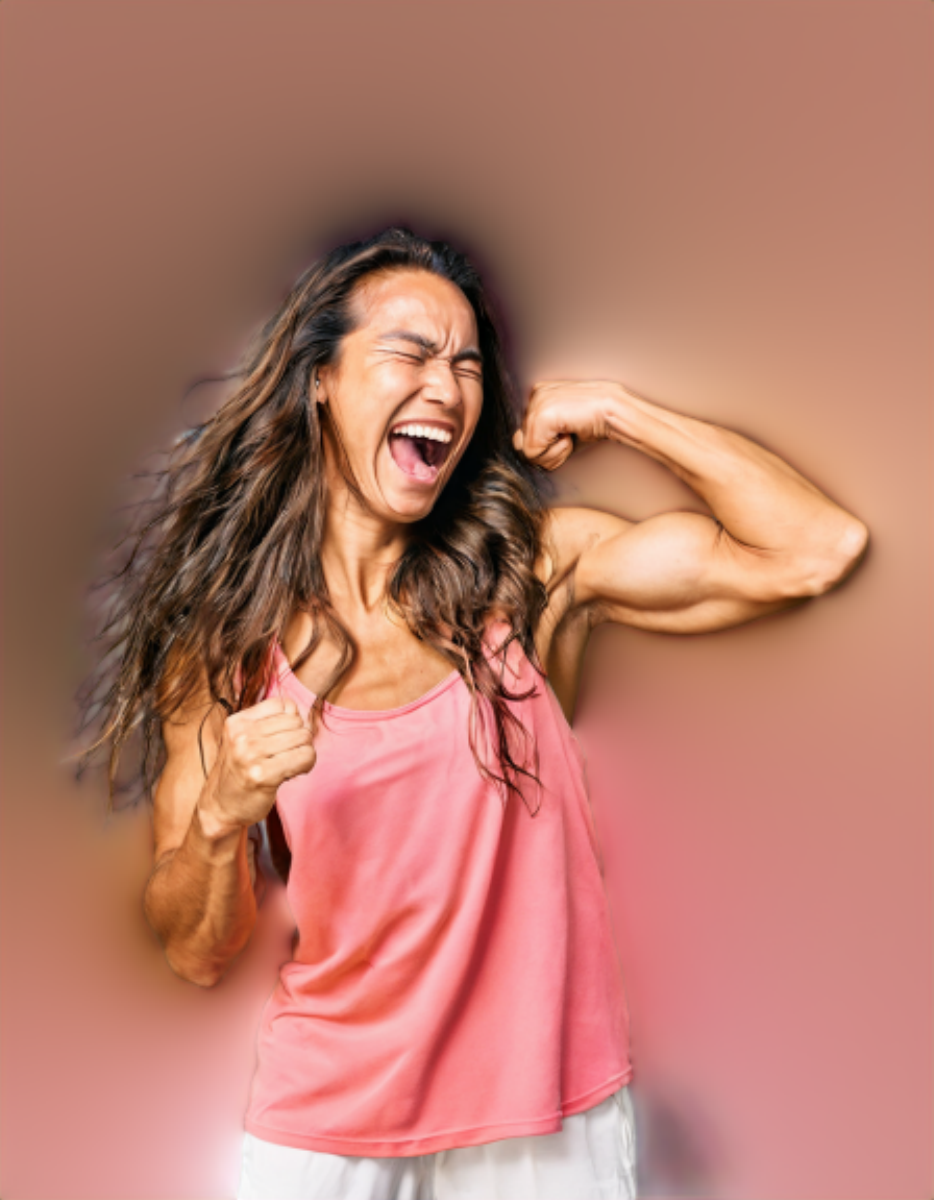}
    \end{minipage}
    \begin{minipage}{0.12\textwidth}
        \includegraphics[scale=.13]{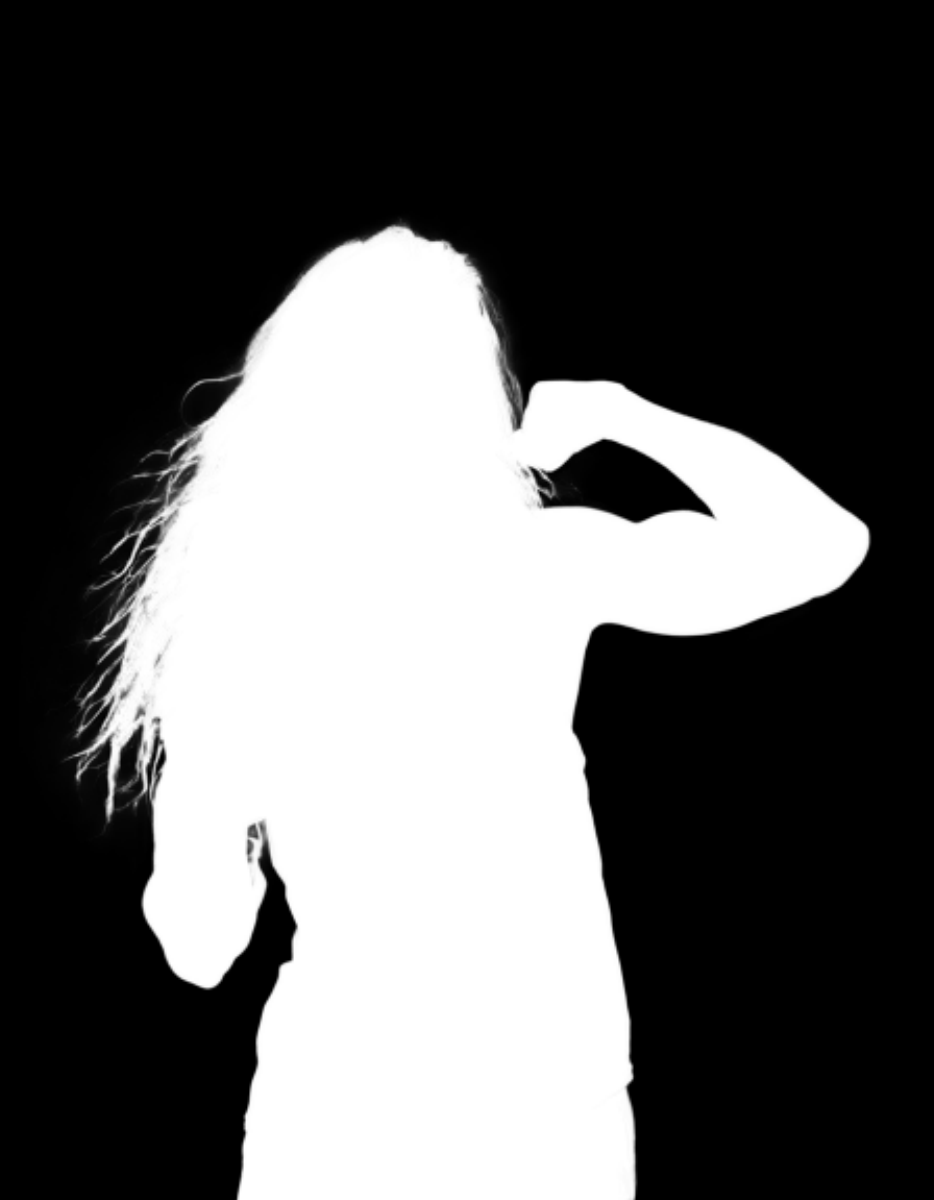}
    \end{minipage}
    \begin{minipage}{0.11\textwidth}
        \includegraphics[scale=.13]{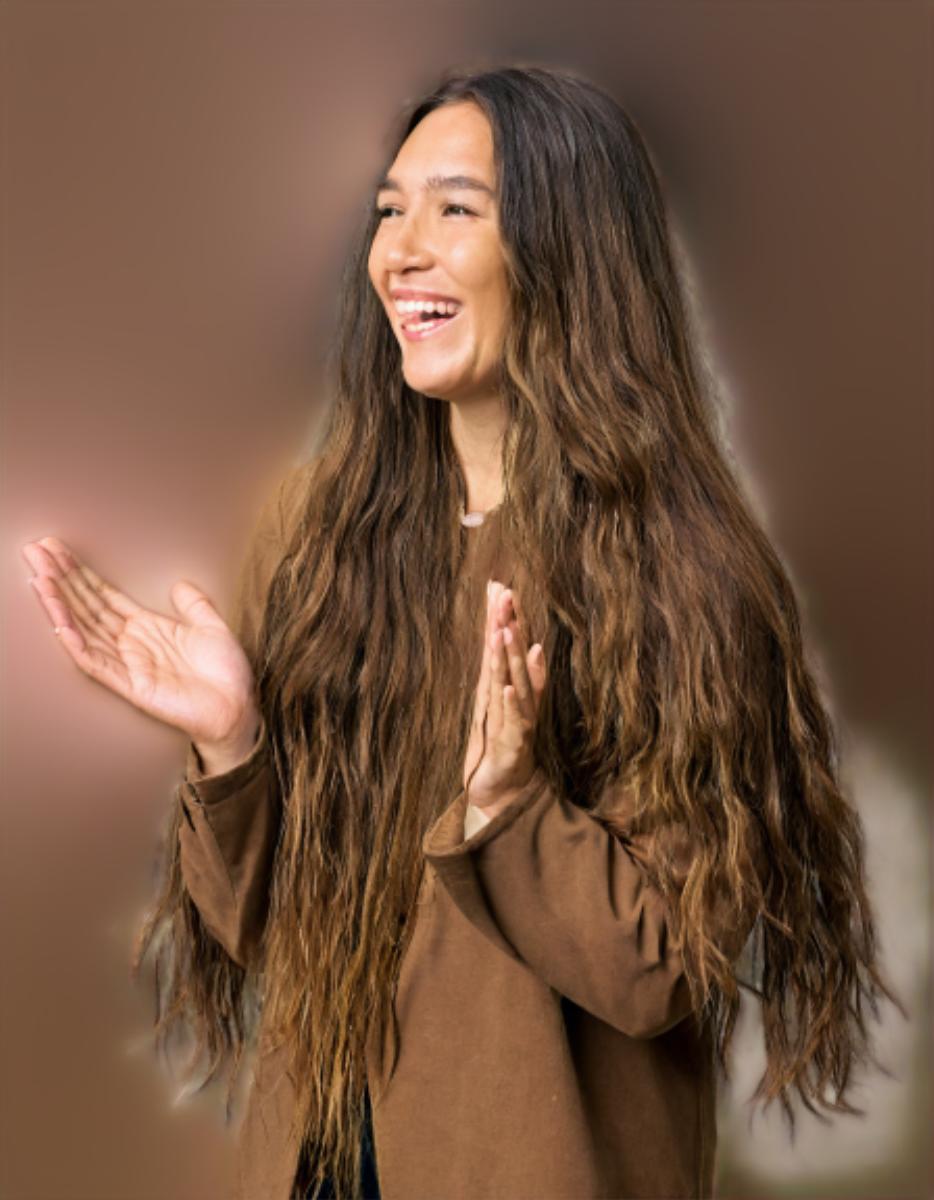}
    \end{minipage}
    \begin{minipage}{0.12\textwidth}
        \includegraphics[scale=.13]{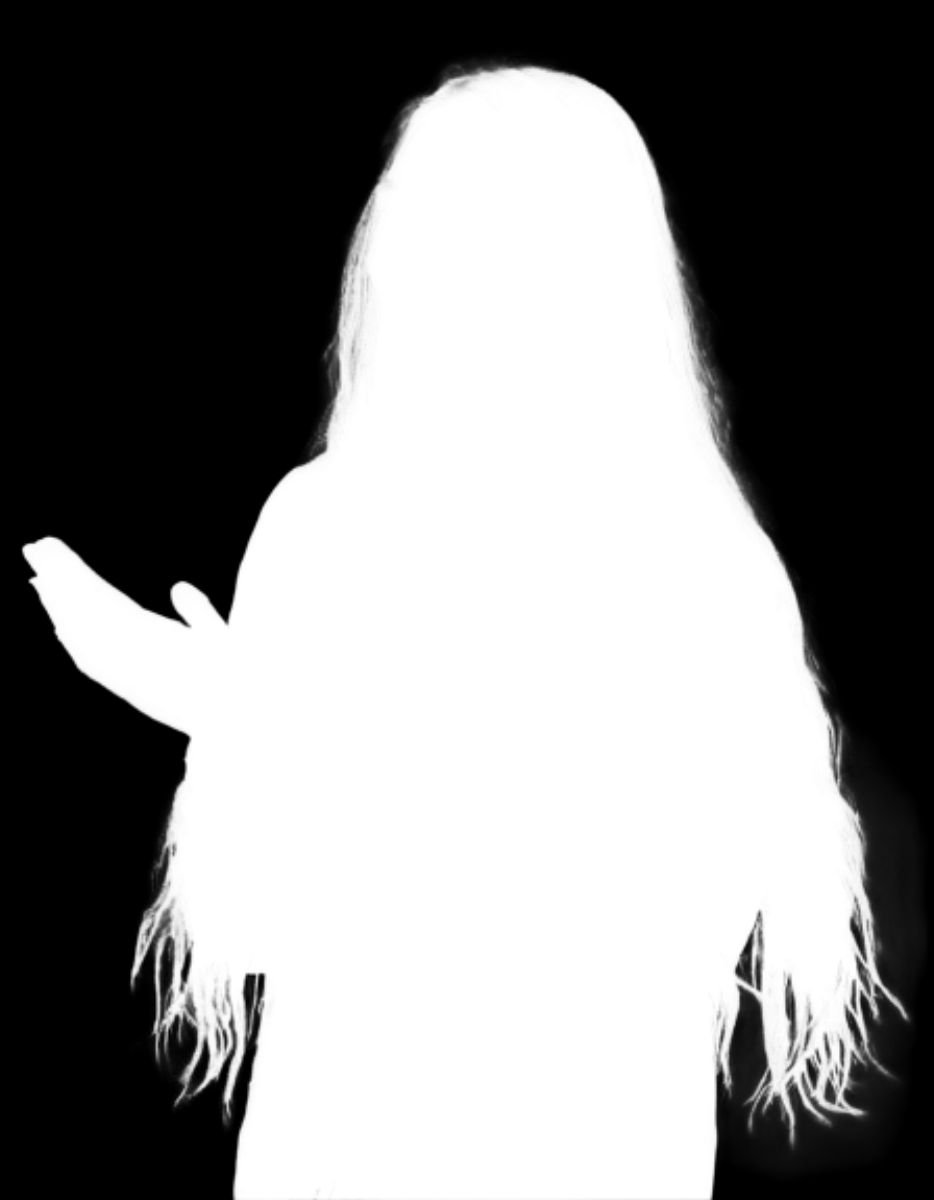}
    \end{minipage}
    \begin{minipage}{0.11\textwidth}
        \includegraphics[scale=.13]{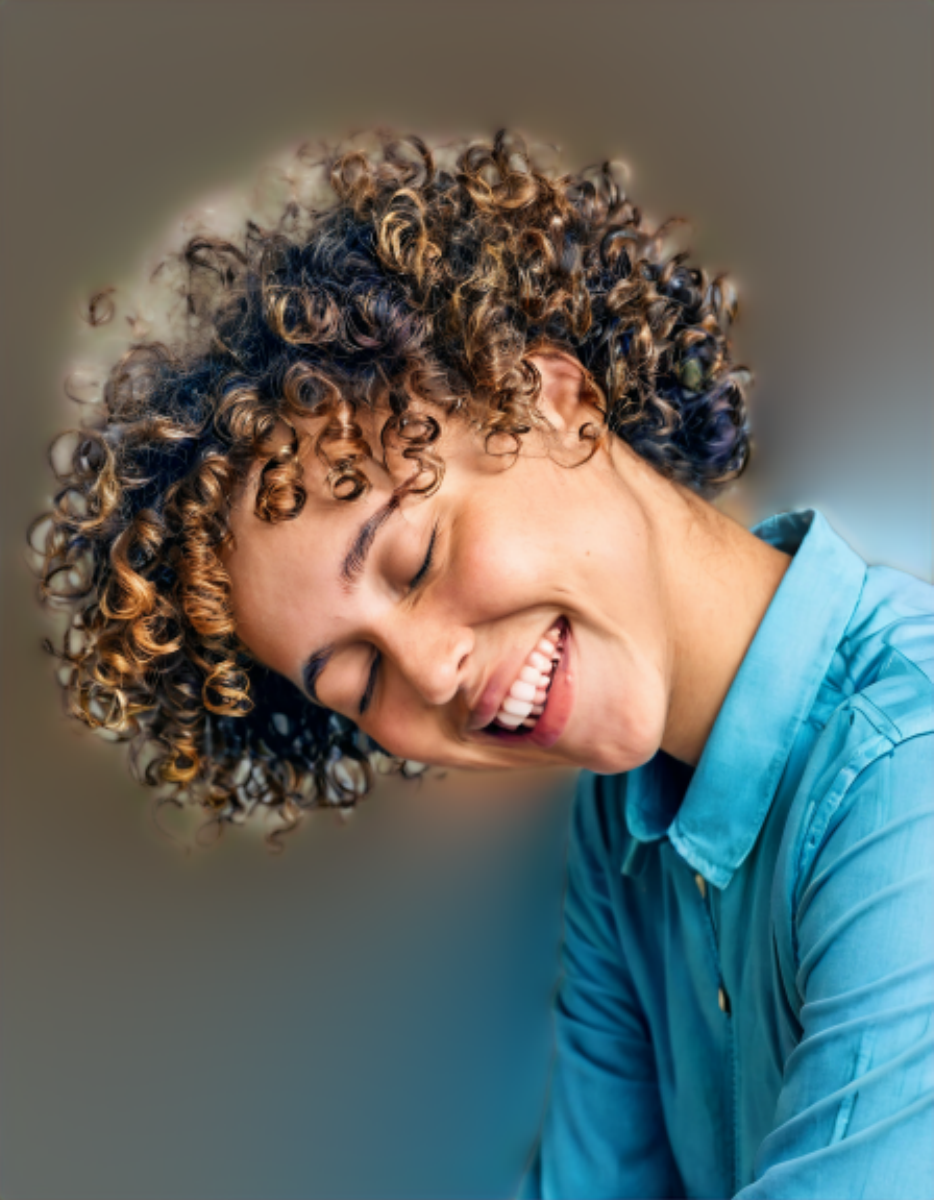}
    \end{minipage}
    \begin{minipage}{0.12\textwidth}
        \includegraphics[scale=.13]{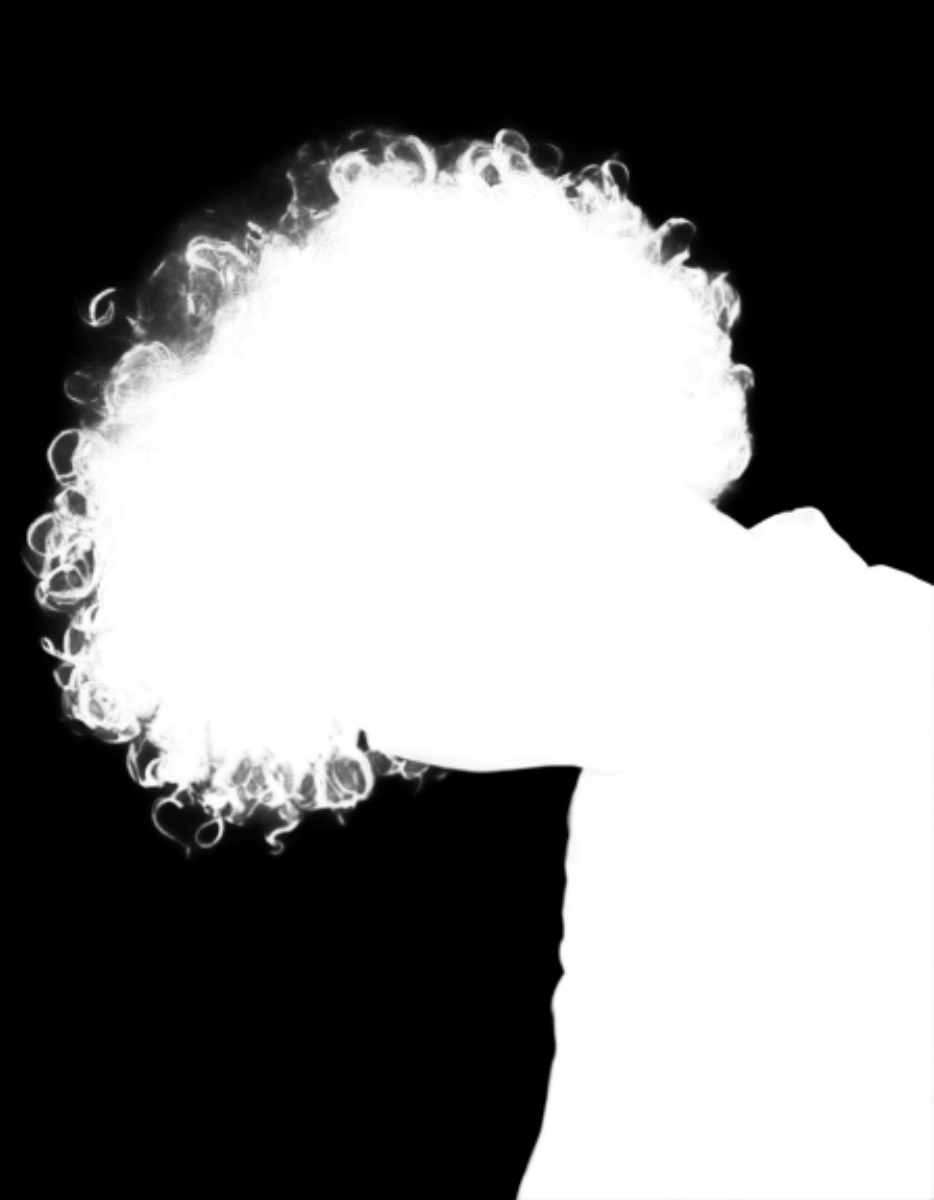}
    \end{minipage}
    }
    \\
    \subfloat{
    \begin{minipage}{0.11\textwidth}
        \includegraphics[scale=.13]{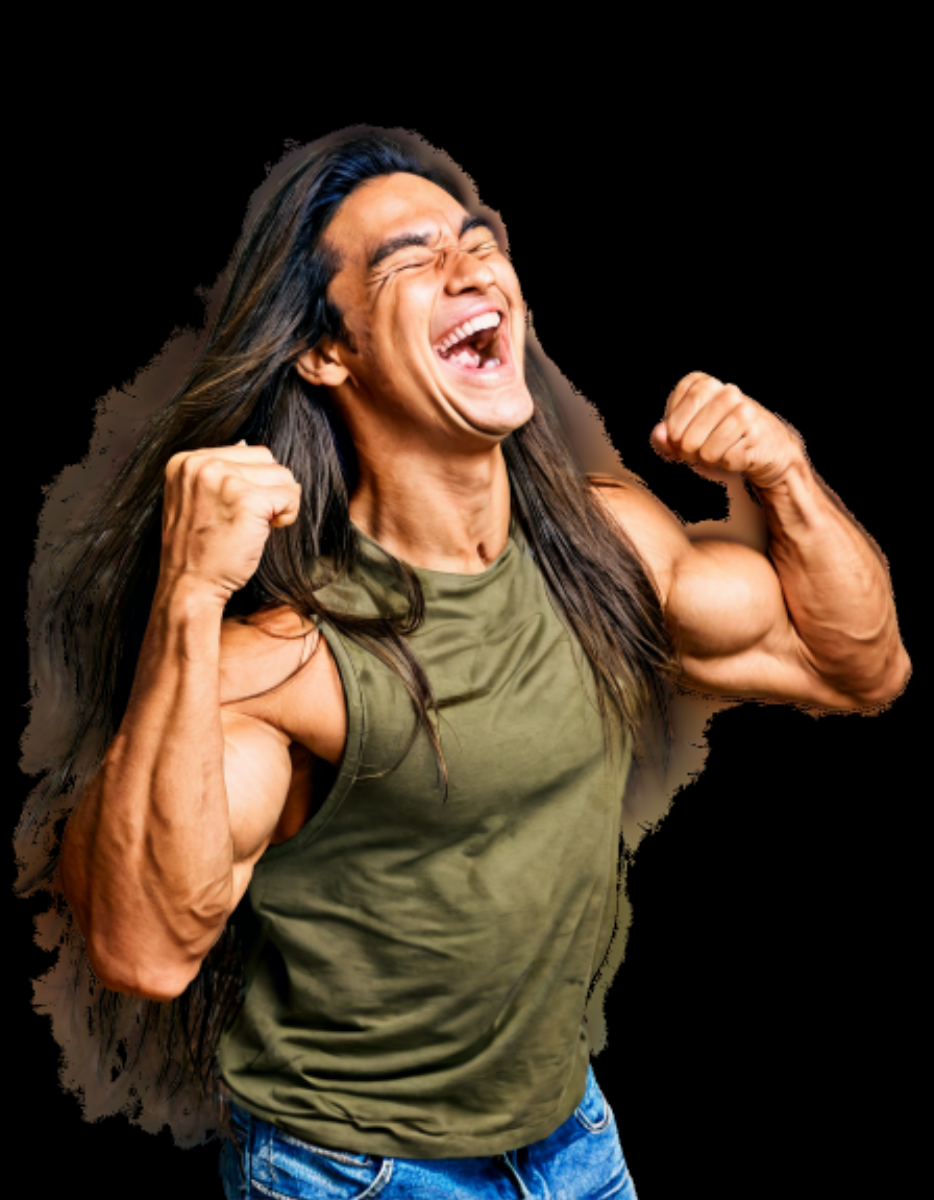}
    \end{minipage}
        \begin{minipage}{0.12\textwidth}
        \includegraphics[scale=.13]{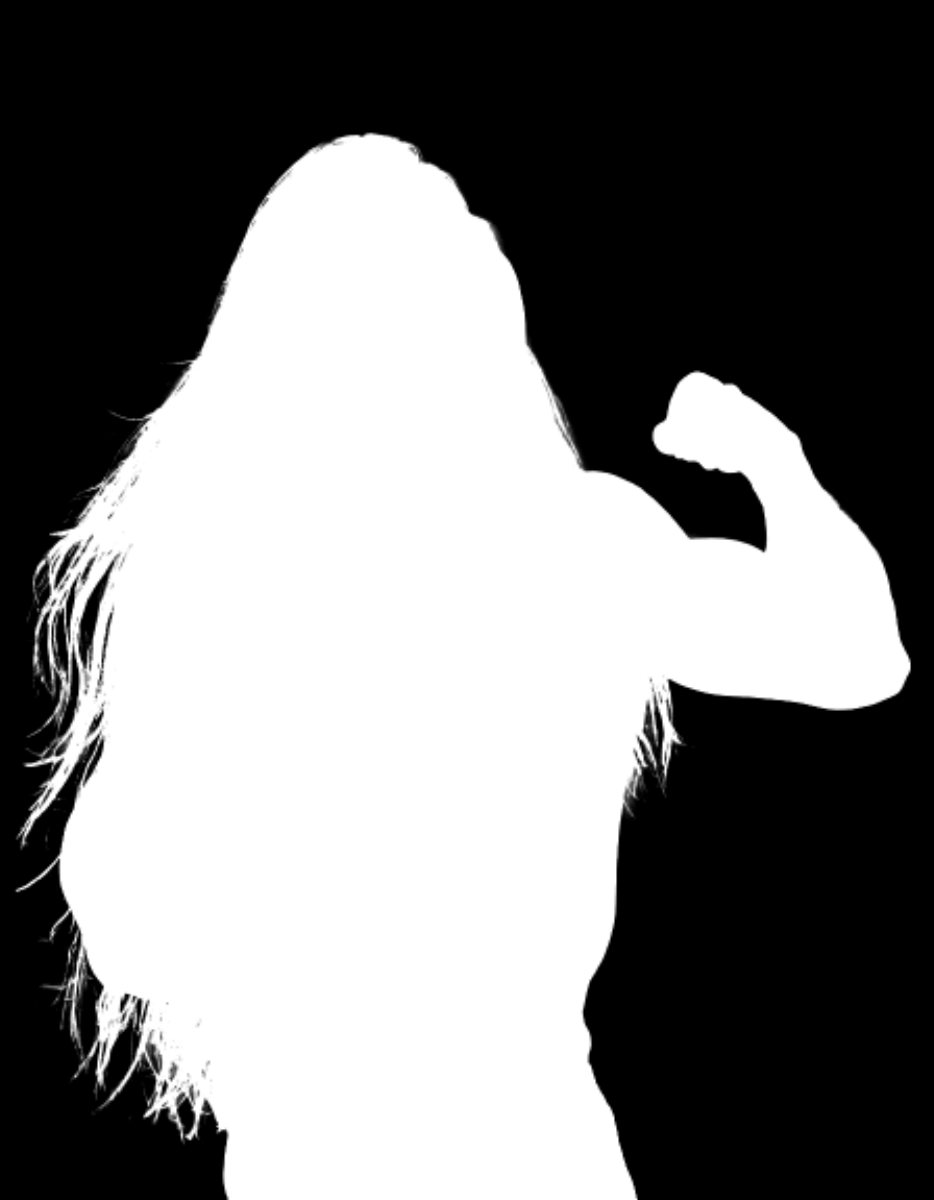}
    \end{minipage}
    \begin{minipage}{0.11\textwidth}
        \includegraphics[scale=.13]{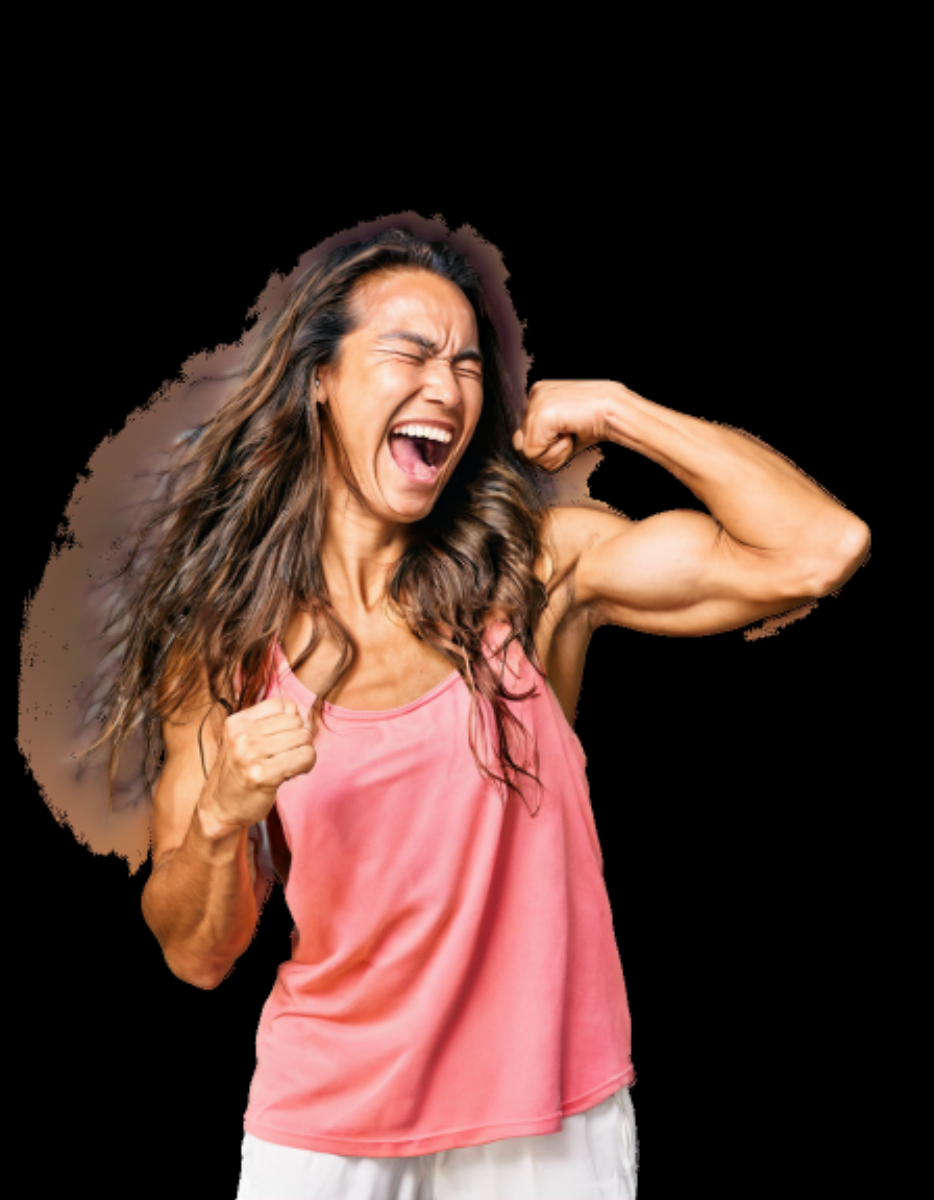}
    \end{minipage}
    \begin{minipage}{0.12\textwidth}
        \includegraphics[scale=.13]{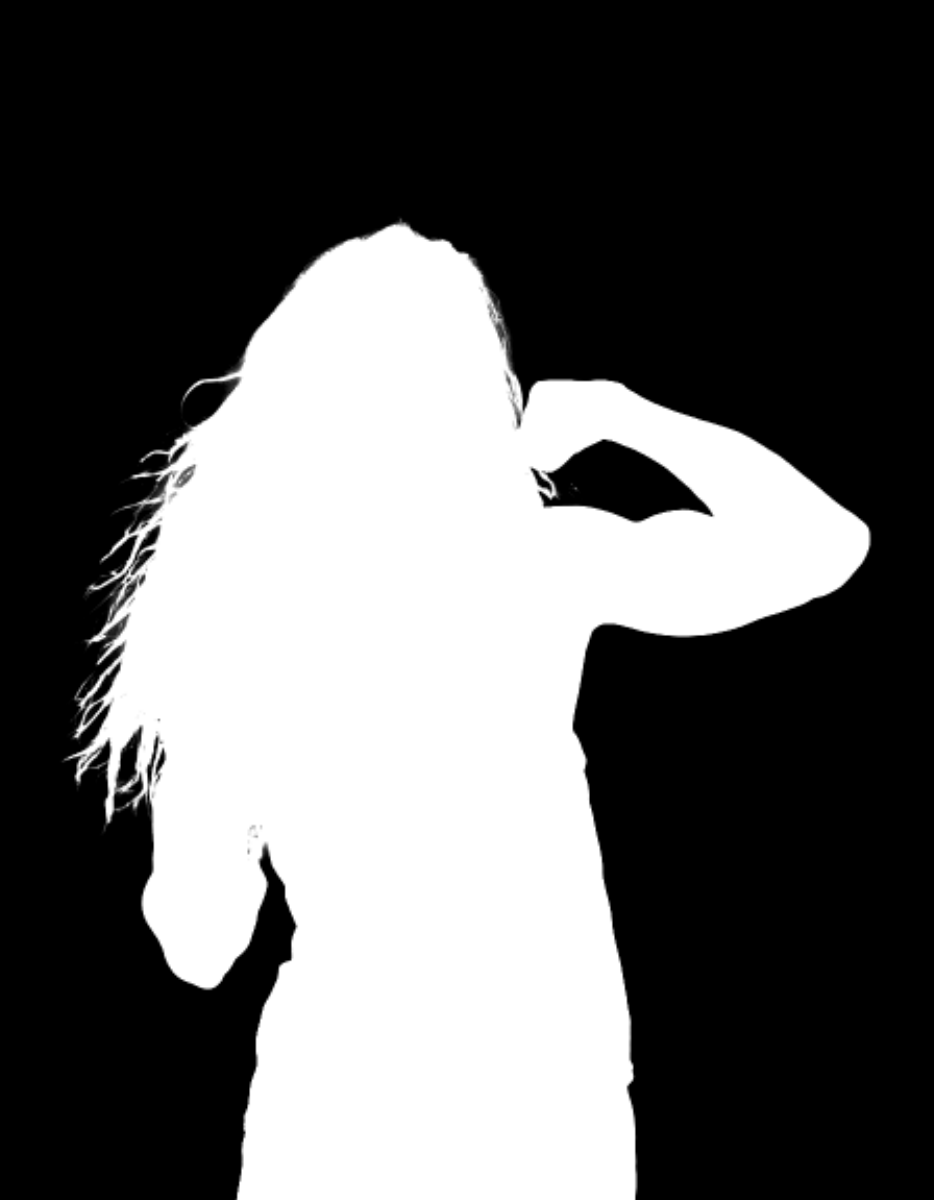}
    \end{minipage}
    \begin{minipage}{0.11\textwidth}
        \includegraphics[scale=.13]{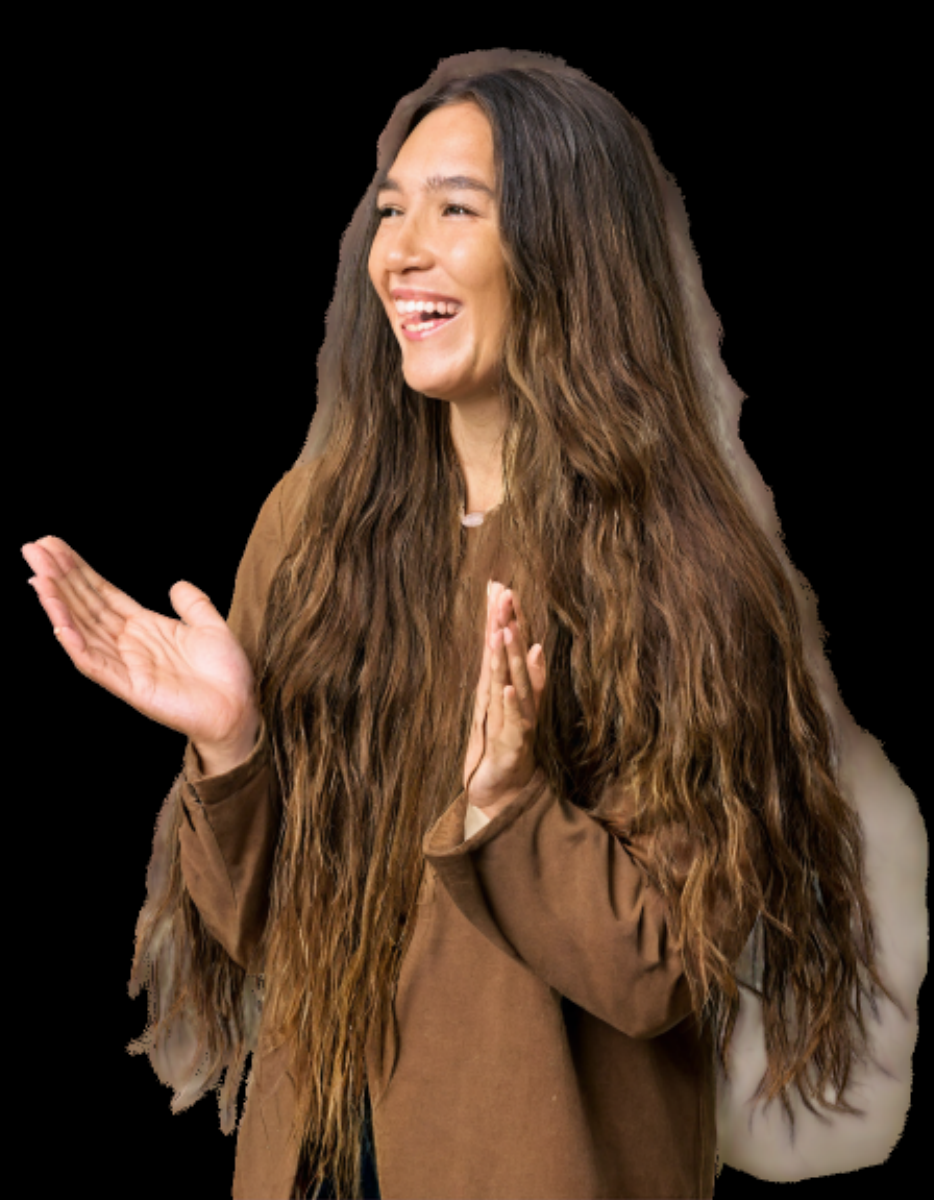}
    \end{minipage}
    \begin{minipage}{0.12\textwidth}
        \includegraphics[scale=.13]{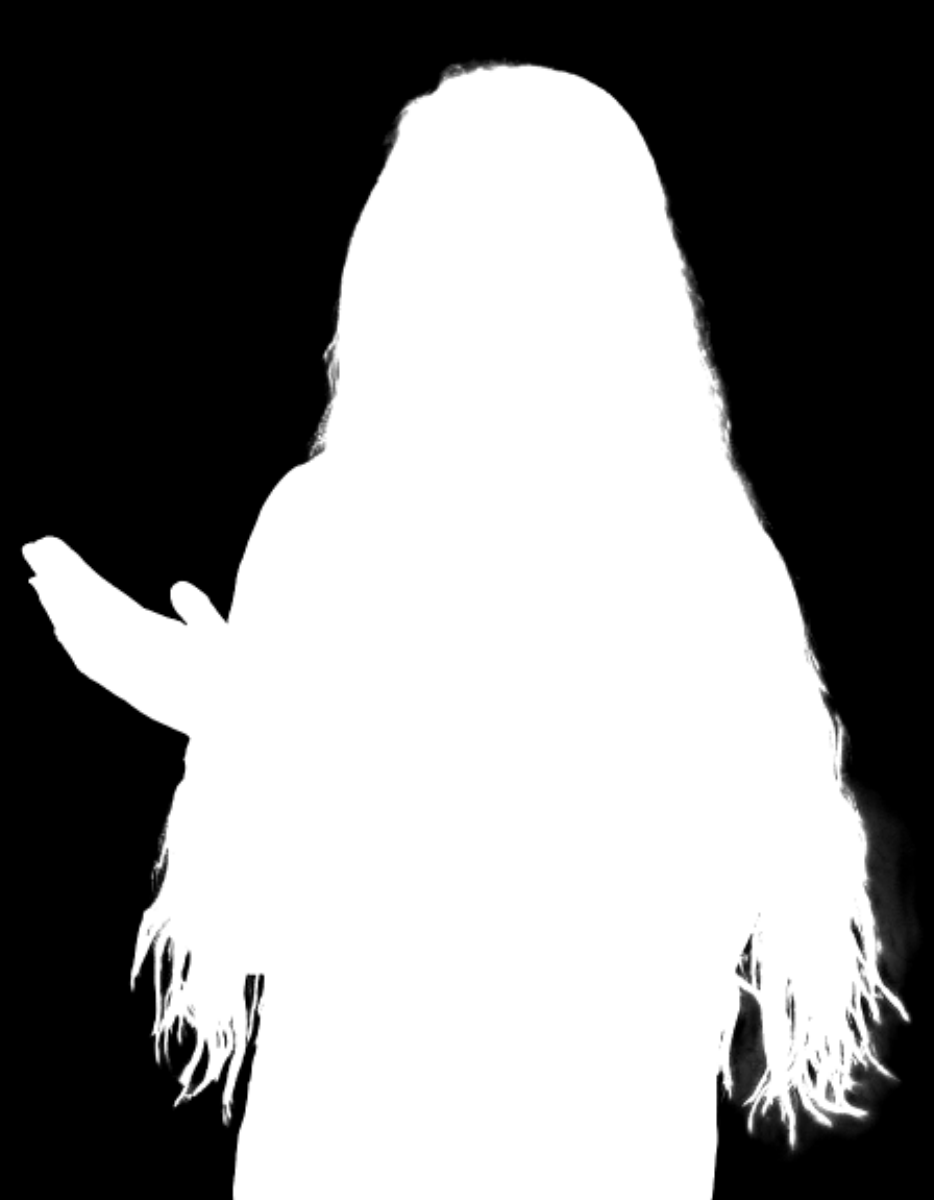}
    \end{minipage}
    \begin{minipage}{0.11\textwidth}
        \includegraphics[scale=.13]{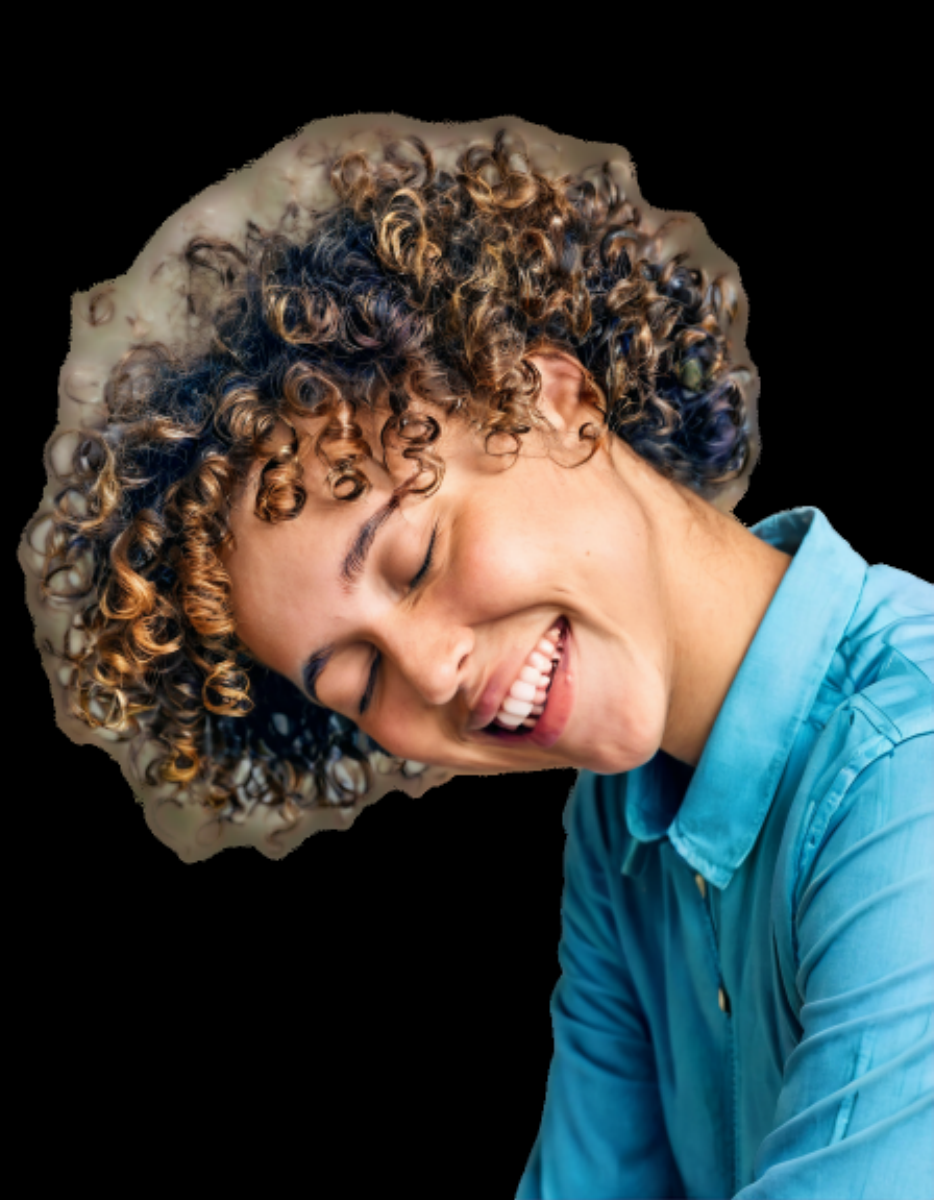}
    \end{minipage}
    \begin{minipage}{0.12\textwidth}
        \includegraphics[scale=.13]{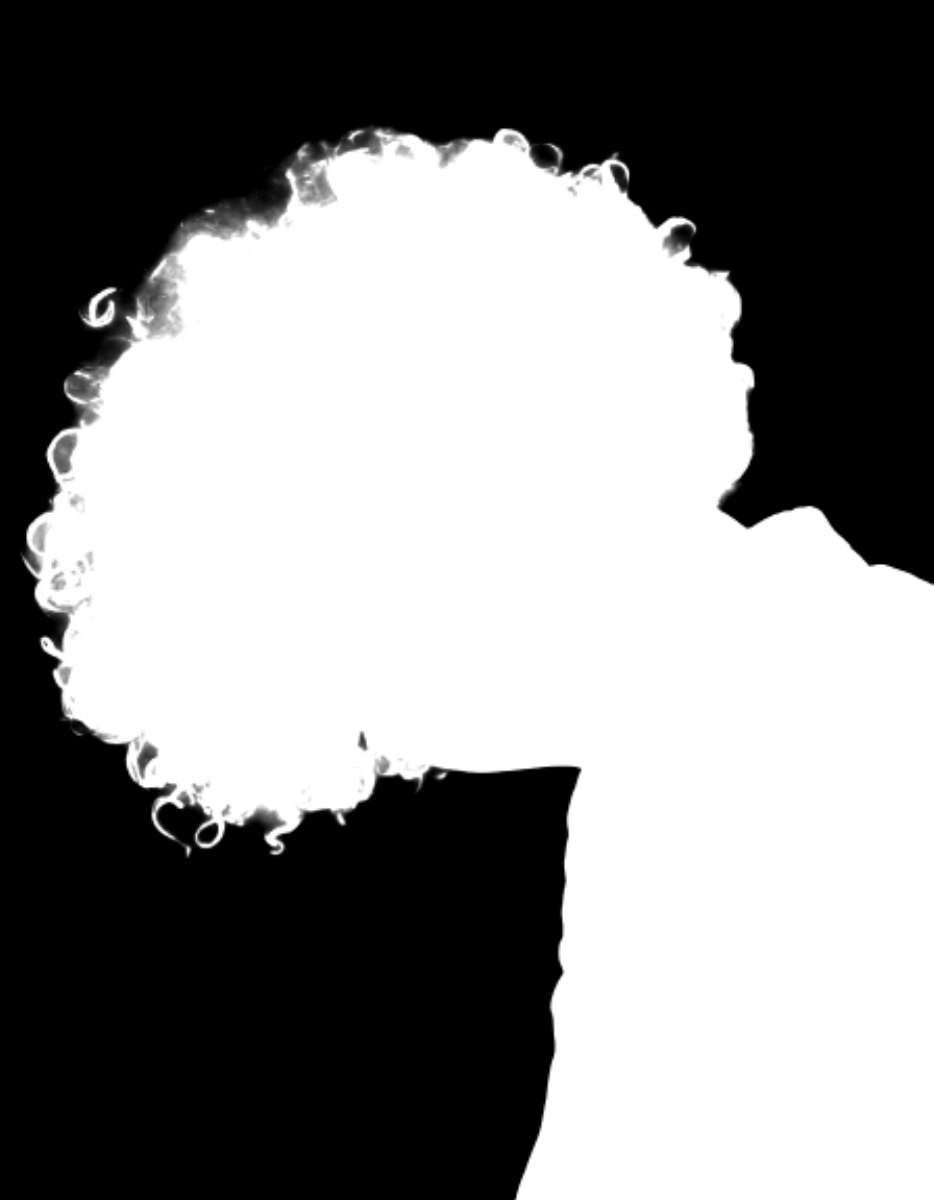}
    \end{minipage}
    }
    \caption{\textbf{Comparison of images and alpha mattes from LD-Portrait-20K and LD-Portrait-CK}. The first row displays images and alpha mattes from the LD-Portrait-20K dataset, while the second row shows the foreground and alpha mattes generated using the chroma keying algorithm.}
    \label{figure-exp-CK}
\end{figure*}

\subsection{Evaluation of Method Effectiveness}

To validate the effectiveness of the proposed dataset construction method, the data from LD-Portrait-20K was merged with a solid color background, and the chroma keying algorithm~\cite{wright2010digital} was applied to compute the new foreground RGB images and alpha mattes. This process is represented by Eq (\ref{eq2}), where $\alpha_1$ and $F_1$ originate from the LD-Portrait-20K dataset, $\alpha_2$ and $F_2$ are the results obtained using the chroma keying algorithm, and $B$ represents the solid color background, with the color selection method following the approach outlined in~\cite{burgert2024magick}. This method generates foreground and alpha mattes for the same image using both the proposed approach and the chroma keying method, facilitating a fair comparison. The foreground and alpha mattes obtained through chroma keying are shown in Figure~\ref{figure-exp-CK}. As demonstrated, the alpha matte generated by the proposed method retains more detail.

\begin{equation}
    \alpha_1 F_1+(1-\alpha_1)B=I=\alpha_2 F_2+(1-\alpha_2)B
    \label{eq2}
\end{equation}

Furthermore, a training comparison was conducted to evaluate the effectiveness of the LD-Portrait-20K dataset against two alternative datasets: LD-Portrait-CK, which consists of chroma keying results, and LD-Portrait-Ori, containing the original alpha mattes generated by Layer Diffusion. Due to the substantial noise in the alpha values of LD-Portrait-Ori, the trimap generation method in ViTMatte-B could not be employed during training. Consequently, LD-Portrait-Ori was only trained on GFM, whereas LD-Portrait-CK was trained on both ViTMatte-B and GFM, with the training configurations remaining consistent with those described earlier. The quantitative comparison results, presented in Table~\ref{table1} and Table~\ref{table2}, demonstrate that the models trained on these two datasets underperformed compared to those trained on LD-Portrait-20K. Notably, the performance of LD-Portrait-Ori is markedly worse, highlighting the effectiveness of the proposed method.

Figure~\ref{figure-exp-1} showcases the testing results of LD-Portrait-CK. The results demonstrate that the LD-Portrait-CK dataset inherits the diversity of the LD-Portrait-20K dataset, allowing the trained models to achieve a certain level of generalization. These models can roughly detect human edges across various complex boundary scenarios. However, their ability to handle finer details remains limited, resulting in weaker performance in extracting precise alpha values along edge regions. This finding further highlights the importance of preserving edge alpha values in the proposed method, which enhances the model’s ability to capture fine-grained details more effectively.

\subsection{Ablation Study of Dataset Capacity}
To further assess the impact of dataset capacity on matting model performance, a subset, LD-Portrait-10K, was created by randomly selecting $10,184$ images from the LD-Portrait-20K dataset. LD-Portrait-10K was then trained for 20 epochs on both ViTMatte-B and GFM. As shown in Table~\ref{table1} and Table~\ref{table2}, the results clearly indicate that increasing the dataset size significantly enhances the performance of the trained models.

Building on this, the qualitative comparison results in Figure~\ref{figure-exp-1} reveal that the model trained on LD-Portrait-10K shows certain advantages over models trained on other datasets. It performs particularly well in extracting edge details and handling complex scenarios, demonstrating acceptable generalization abilities. However, when compared to the model trained on LD-Portrait-20K, the LD-Portrait-10K model falls slightly short in terms of detail preservation and edge accuracy. This highlights that while LD-Portrait-10K offers some performance improvements, the larger LD-Portrait-20K dataset proves significantly more effective in capturing diversity and fine-grained features, reinforcing the effectiveness of the proposed method and dataset.

\subsection{Impact on Video Human Matting Performance}
With the rapid advancements in video segmentation models, these models have become effective priors for generating trimaps in matting tasks. Building on this foundation, a method is proposed that utilizes masks generated by video segmentation, applies morphological operations to construct trimaps, and integrates trimap-based matting approaches to achieve video matting. Following this approach, ViTMatte-B models trained on different datasets were combined with SAM2~\cite{ravi2024sam} to construct a video matting framework, referred to as S+V. To evaluate the effectiveness of this framework, comparisons were made against several state-of-the-art video matting models, including RVM~\cite{lin2022robust}, FTP-VM~\cite{huang2023end}, SparseMat~\cite{sun2023ultrahigh}, and VMFormer~\cite{li2024vmformer}.

In the comparative experiments, RVM, FTP-VM, and VMFormer were trained on video datasets; therefore, their official pre-trained weights were directly used for evaluation. For SparseMat, although it was originally trained on the HHM50K dataset, reproducing its experiments using the publicly available code produced results that differed significantly from those reported in its paper. Consequently, its official pre-trained weights were relied upon for testing.

For validation, the $videoMatte512\times288$ dataset provided by RVM and the $vmformer512\times288public$ dataset provided by VMFormer were utilized. Both validation datasets are derived from VideoMatte240K and share the same resolution of $512\times288$. To comprehensively evaluate video matting performance, in addition to traditional matting metrics, a temporal consistency metric, dtSSD~\cite{erofeev2015perceptually}, scaled by $1e^2$ for clarity, was introduced. Table~\ref{table3} presents the quantitative evaluation results for all models across various metrics.

As shown in Table~\ref{table3}, the S+V framework, which integrates ViTMatte-B models trained on LD-Portrait-20K with SAM2, achieves the best performance across all evaluation metrics, including a significant advantage in temporal consistency (dtSSD). This demonstrates its exceptional capability in preserving edge details and achieving robust temporal consistency. Particularly, the S+V framework significantly outperforms other models in dtSSD, benefiting from the reliable temporal alignment provided by SAM2. These results highlight the adaptability and robustness of the S+V framework in video portrait matting tasks.

\begin{figure}
    \centering
    \subfloat{
    \begin{minipage}{0.01\textwidth}
    \rotatebox{90}{Image}
    \end{minipage}
    \hfill
    \begin{minipage}{0.108\textwidth}
        \includegraphics[scale=.114550781]{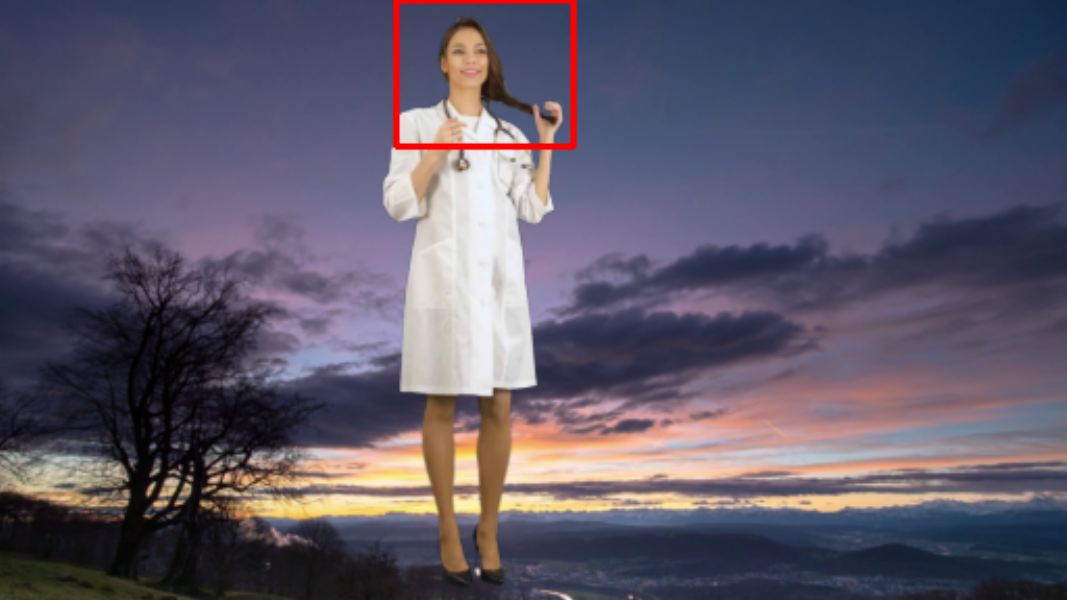}
    \end{minipage}
    \hfill
    \begin{minipage}{0.108\textwidth}
        \includegraphics[scale=.114550781]{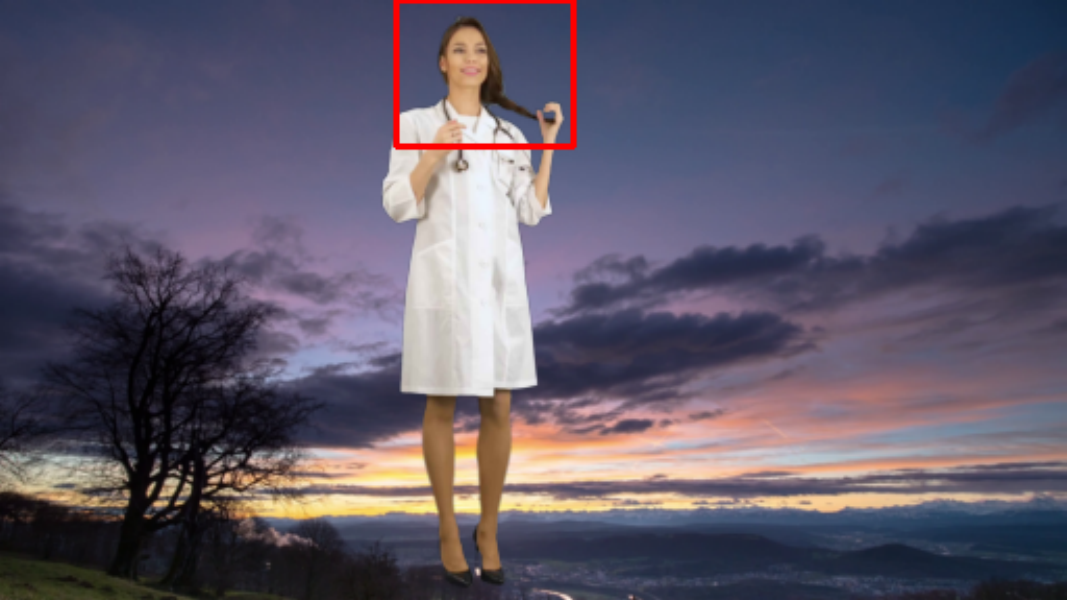}
    \end{minipage}
    \hfill
    \begin{minipage}{0.108\textwidth}
      \includegraphics[scale=.114550781]{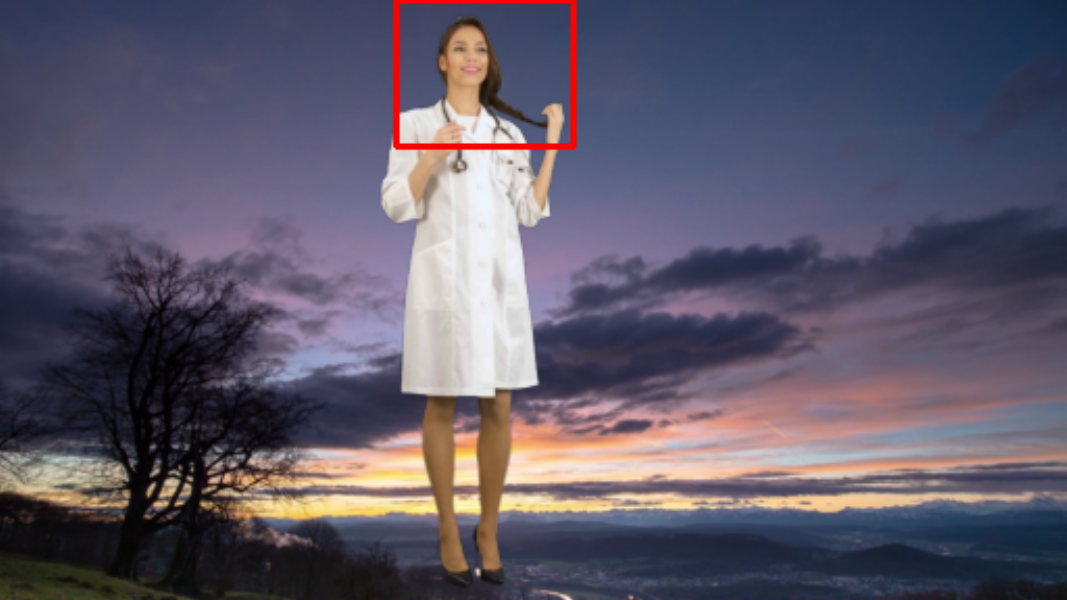}
    \end{minipage}
    \hfill
    \begin{minipage}{0.108\textwidth}
         \includegraphics[scale=.114550781]{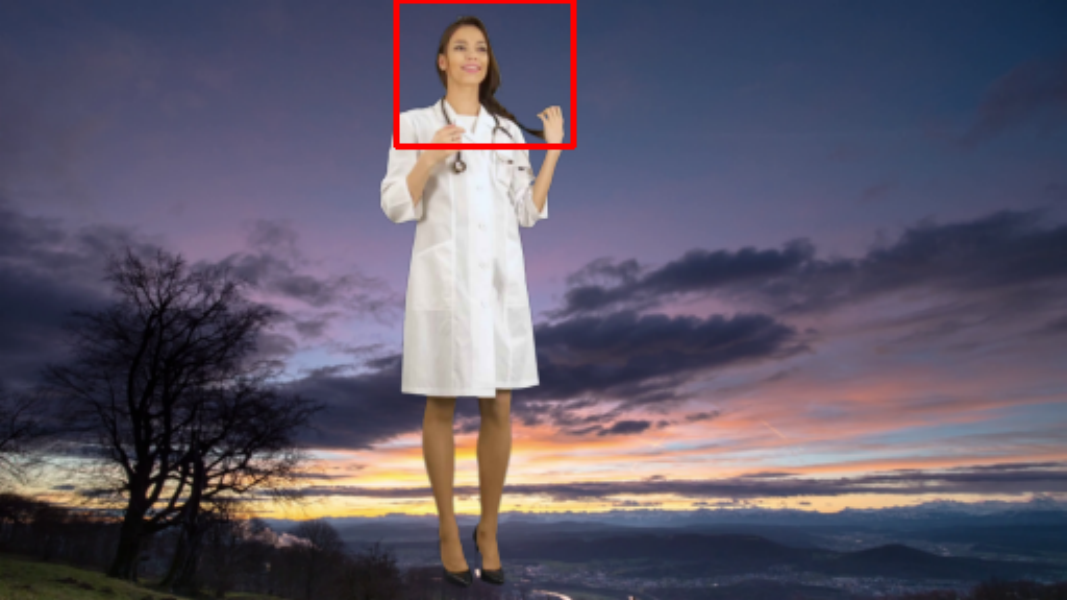}
    \end{minipage}
    }
    \\
    \vspace{-3mm}
    \subfloat{
    \begin{minipage}{0.01\textwidth}
     \rotatebox{90}{RVM}
    \end{minipage}
        \hfill
    \begin{minipage}{0.108\textwidth}
        \includegraphics[scale=.69]{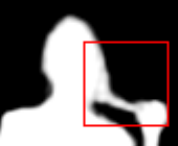}
    \end{minipage}
    \hfill
    \begin{minipage}{0.108\textwidth}
        \includegraphics[scale=.69]{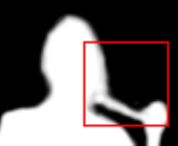}
    \end{minipage}
    \hfill
    \begin{minipage}{0.108\textwidth}
      \includegraphics[scale=.69]{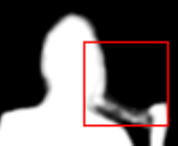}
    \end{minipage}
    \hfill
    \begin{minipage}{0.108\textwidth}
         \includegraphics[scale=.69]{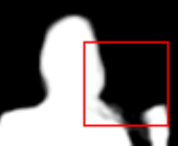}
    \end{minipage}
    }
     \\
     \vspace{-3mm}
    \subfloat{
     \begin{minipage}{0.01\textwidth}
     \rotatebox{90}{FTP-VM}
    \end{minipage}
    \hfill
    \begin{minipage}{0.108\textwidth}
        \includegraphics[scale=.69]{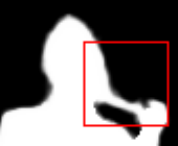}
    \end{minipage}
    \hfill
    \begin{minipage}{0.108\textwidth}
        \includegraphics[scale=.69]{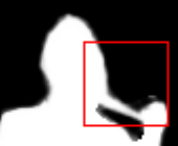}
    \end{minipage}
    \hfill
    \begin{minipage}{0.108\textwidth}
      \includegraphics[scale=.69]{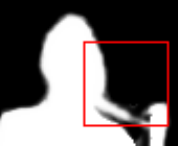}
    \end{minipage}
    \hfill
    \begin{minipage}{0.108\textwidth}
         \includegraphics[scale=.69]{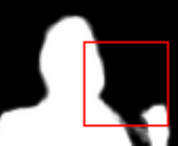}
    \end{minipage}
    }
     \\
     \vspace{-3mm}
    \subfloat{
     \begin{minipage}{0.01\textwidth}
     \rotatebox{90}{SparseMat}
    \end{minipage}
    \hfill
    \begin{minipage}{0.108\textwidth}
        \includegraphics[scale=.69]{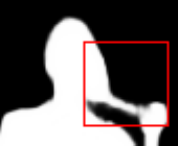}
    \end{minipage}
    \hfill
    \begin{minipage}{0.108\textwidth}
        \includegraphics[scale=.69]{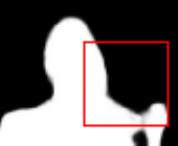}
    \end{minipage}
    \hfill
    \begin{minipage}{0.108\textwidth}
      \includegraphics[scale=.69]{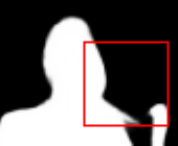}
    \end{minipage}
    \hfill
    \begin{minipage}{0.108\textwidth}
         \includegraphics[scale=.69]{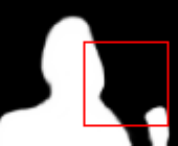}
    \end{minipage}
    }
     \\
     \vspace{-3mm}
    \subfloat{
     \begin{minipage}{0.01\textwidth}
     \rotatebox{90}{VMFormer}
    \end{minipage}
    \hfill
    \begin{minipage}{0.108\textwidth}
        \includegraphics[scale=.69]{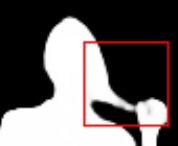}
    \end{minipage}
    \hfill
    \begin{minipage}{0.108\textwidth}
        \includegraphics[scale=.69]{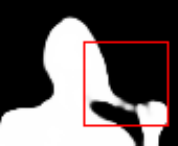}
    \end{minipage}
    \hfill
    \begin{minipage}{0.108\textwidth}
      \includegraphics[scale=.69]{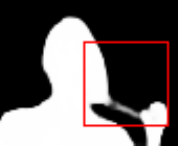}
    \end{minipage}
    \hfill
    \begin{minipage}{0.108\textwidth}
         \includegraphics[scale=.69]{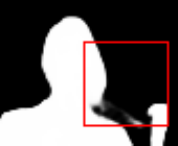}
    \end{minipage}
    }
     \\
     \vspace{-3mm}
    \subfloat{
     \begin{minipage}{0.01\textwidth}
     \rotatebox{90}{S+V}
    \end{minipage}
    \hfill
    \begin{minipage}{0.108\textwidth}
        \includegraphics[scale=.69]{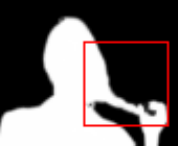}
    \end{minipage}
    \hfill
    \begin{minipage}{0.108\textwidth}
        \includegraphics[scale=.69]{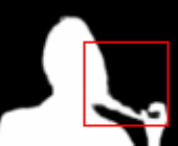}
    \end{minipage}
    \hfill
    \begin{minipage}{0.108\textwidth}
      \includegraphics[scale=.69]{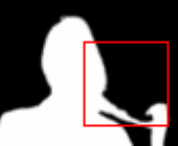}
    \end{minipage}
    \hfill
    \begin{minipage}{0.108\textwidth}
         \includegraphics[scale=.69]{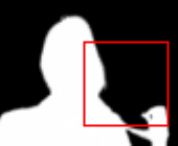}
    \end{minipage}
    }
    \\
    \vspace{-3mm}
    \subfloat{
     \begin{minipage}{0.01\textwidth}
     \rotatebox{90}{GT}
    \end{minipage}
    \hfill
    \begin{minipage}{0.108\textwidth}
        \includegraphics[scale=.69]{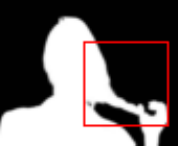}
    \end{minipage}
    \hfill
    \begin{minipage}{0.108\textwidth}
        \includegraphics[scale=.69]{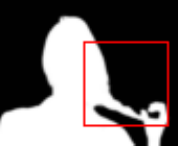}
    \end{minipage}
    \hfill
    \begin{minipage}{0.108\textwidth}
      \includegraphics[scale=.69]{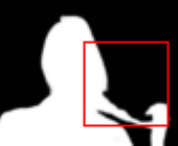}
    \end{minipage}
    \hfill
    \begin{minipage}{0.108\textwidth}
         \includegraphics[scale=.69]{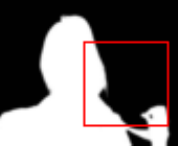}
    \end{minipage}
    }
    \caption{
\textbf{Qualitative Comparison of Different Video Matting Models on Consecutive Frames}. The results of the S+V framework shown in the figure were obtained using the ViTMatte-B model trained on LD-Portrait-20K.}
    \label{figure-video}
\end{figure}

Figure~\ref{figure-video} illustrates the qualitative comparison results on consecutive video frames. The temporally consistent trimaps generated by SAM2 enable the S+V framework to maintain seamless transitions across frames. Combined with the excellent edge-handling capability of ViTMatte-B trained on LD-Portrait-20K, the framework achieves outstanding performance in processing hair regions of moving subjects. It effectively captures fine details in dynamic scenes while maintaining temporal consistency. In contrast, other methods fail to accurately recognize edge regions under such scenarios, resulting in noticeable detail loss or incorrect matting. These findings further demonstrate the superiority of the S+V framework in addressing the challenges of video portrait matting, particularly in handling complex edges and dynamic variations.

\section{Conclusion}
This paper introduces an observed prior that the semi-transparent regions of an alpha matte are connected and, based on this insight, proposes a low-cost, high-quality method for generating portrait matting datasets using Layer Diffusion. Specifically, Layer Diffusion is utilized to generate portrait images with alpha channels. Initial screening is performed to filter out samples with significant deviations in their alpha mattes. Following this, Connectivity-Aware Alpha Refinement is applied, leveraging the connectivity of the inverse alpha to correct erroneous alpha values, ultimately producing high-quality alpha mattes. Using this approach, the LD-Portrait-20K dataset, comprising $20,051$ high-quality images, is constructed.

Through comprehensive comparative experiments with other datasets, this paper demonstrates the superior quality of LD-Portrait-20K and its effectiveness in improving the performance of trained models. Additionally, the effectiveness of the proposed dataset generation method is validated through comparisons with chroma keying and ablation studies on dataset size. Furthermore, segmentation models are integrated with trimap-based matting models to develop a video portrait matting framework. Experimental results reveal that this framework significantly outperforms existing video matting models, further underscoring the contribution of the dataset to video portrait matting tasks. This paper aims to improve the efficiency of creating portrait matting datasets and foster further advancements in portrait matting techniques.

\bibliography{main}
\bibliographystyle{IEEEtran}

\newpage

\vfill

\end{document}